\documentclass{article}

\usepackage{epsfig}
\usepackage{algorithm,algpseudocode}
 \usepackage{amsfonts,tabularx}
 \usepackage{multirow}
\usepackage{amsmath}
 \usepackage{arydshln}
 \usepackage{booktabs}
  \usepackage{multirow}
  \usepackage{setspace}
  \usepackage{color}
    \usepackage[margin=1in]{geometry}
  \usepackage[font={small},labelfont={small}]{caption}
  \usepackage{graphicx}
\usepackage{tikz}
\usepackage{adjustbox}
\usepackage[square,sort,comma,numbers]{natbib}
\usepackage{graphicx}
\usepackage{subcaption}
\usepackage{hyperref}
\newcommand{\compresslist}{%
\setlength{\itemsep}{1pt}%
\setlength{\parskip}{0pt}%
\setlength{\parsep}{0pt}%
}

\newcommand{\bth}{\boldsymbol\theta}
\newcommand{\Bth}{\boldsymbol\Theta}
\newcommand{\bga}{\boldsymbol\gamma}
\newcommand{\bw}{\mathbf{w}}
\newcommand{\Br}{\mathbf{R}}
\newcommand{\br}{\mathbf{r}}

\newcommand{\bell}{\boldsymbol{\ell}}

\title{Surrogate Optimization of Deep Neural Networks for Groundwater Predictions}

\author{
  Juliane M\"uller, Jangho Park, Reetik Sahu, Deborah Agarwal\\
  Computational Research Division \\
  Lawrence Berkeley National Laboratory\\
  1 Cyclotron Rd.\\
  Berkeley, CA, 94720 \\
  \texttt{\{JulianeMueller, JanghoPark, ReetikSahu, daagarwal\}@lbl.gov} \\
\and
 Charuleka  Varadharajan, Bhavna Arora, Boris Faybishenko \\
Earth and Environmental Sciences Area \\
  Lawrence Berkeley National Laboratory\\
  1 Cyclotron Rd.\\
  Berkeley, CA, 94720 \\
    \texttt{\{cvaradharajan, barora, bafaybishenko\}@lbl.gov} \\
}

\begin{document}
\maketitle

\begin{abstract}
Sustainable management of groundwater resources under changing climatic conditions require an application of reliable and accurate predictions of groundwater levels.  Mechanistic multi-scale, multi-physics  simulation models are often too hard to use for this purpose, especially for groundwater managers who do not have access to the complex compute resources and data. Therefore, we analyzed the applicability and performance of four modern deep learning computational models for predictions of groundwater levels.  We compare three  methods for optimizing the models' hyperparameters, including two surrogate model-based algorithms and a random sampling method.  The models were tested using predictions of the groundwater level in  Butte County, California, USA, taking into account the temporal variability of streamflow, precipitation, and ambient temperature. Our numerical  study shows that the optimization of the hyperparameters can lead to reasonably accurate performance of all models (root mean squared errors of groundwater predictions of 2 meters or less), but the ``simplest'' network, namely a multilayer perceptron (MLP) performs overall  better for learning and predicting groundwater data than the more advanced long short-term memory  or convolutional neural networks in terms of prediction accuracy and time-to-solution, making the MLP a suitable candidate for groundwater prediction. 

\end{abstract}

\section{Introduction}

The massive amount of data obtained from field observations and computer simulations has resulted in heightening an interest in the application of machine learning (ML) models and artificial intelligence to interpret observational data, detect patterns, and derive operational decisions related to management of water resources.  Although scientists have been exploiting ML tools for image recognition~\cite{LeCun2015}, development of autonomous vehicles~\cite{Kuderer2015}, inferring physical and chemical phenomena~\cite{Rudy2019}, and many more, their application for hydrological predictions is still limited. 

Selecting an appropriate ML model is a challenging problem and the model architecture needs to be designed with an optimal set of hyperparameters such as the number of hidden layers and nodes per layer in deep learning models.   Once these hyperparameters have been set, a very large-scale  global optimization problem has to be solved in which we fit the learning model to the data by optimizing the network's weights. 
Stochastic optimization methods such as stochastic gradient descent~\cite{Bottou2010, Robbins1951} or the ADAM optimizer~\cite{Kingma2015} have been successfully used for this task. In practice, the hyperparameters are often tuned ``by hand''~\cite{Najah2013,Zhang2018}, i.e.,  hyperparameters are chosen, perhaps, at random or based on previous experience, and then the ML model's weights are optimized.  One of the drawbacks of this approach is that training the model can be computationally expensive and, depending on the dataset,  may require  from a few minutes to several hours of compute time,  and thus, the final hyperparameters  obtained with this approach are usually not optimal.

To date, only a few systematic approaches have been developed to automate HPO in learning models. This includes the widely-used random and grid search approaches~\cite{Bergstra2012}. However, the grid search does not scale well with  the number of hyperparameters in the model.  Ilievski et al.~\cite{Ilievski2017} use a deterministic surrogate model algorithm based on radial basis functions to tune the hyperparameters of deep neural networks. The authors  show the performance of the method on image data only, and it is unclear if the results can be generalized to time series data.  Snoek et al.~\cite{Snoek2012} developed a Gaussian process (GP) based algorithm for HPO, and the authors investigate how the choice of the GP kernel and the GP's own hyperparameters can influence the performance of the ML model. Bergstra~\cite{Bergstra2013}  used Bayesian methods for HPO. Bayesian optimization has also been used by~\cite{Klein2017} to train support vector machines and convolutional neural networks (CNNs) for large data sets. The authors train their learning model on a subset of the  data in order to decrease the amount of required training time.  The developers of Auto-Keras~\cite{Jin2019} also used Bayesian optimization with a new type of kernel for the GP  model, and a new acquisition function that enables the optimization of network morphism. The disadvantages of Bayesian optimization and GPs, in general, are the underlying assumptions that are made regarding the distributions of the errors and the covariance structure, which is usually not  available a priori. Moreover, GPs do not scale very well with the number of observations because optimizing the GP's own hyperparameters becomes computationally expensive. Another approach to HPO is the use of genetic algorithms. For example, the Multinode Evolutionary Neural Networks for Deep Learning~\cite{MENNDL} tool uses a genetic algorithm and support vector machines for tuning the hyperparameters of CNNs. Evolutionary algorithms are known to require hundreds to thousands of function evaluations to find a solution, which is not always feasible to do.  The authors of HyperNOMAD~\cite{hypernomad} extended  the NOMAD algorithm~\cite{LeDIgabel2011} to automatically optimize the hyperparameters and the learning process of deep neural networks using the mesh adaptive direct search algorithm. The authors show the performance of HyperNOMAD on image data.   Finally, surrogate models have been used for the automated HPO of deep learning models by~\cite{Balaprakash2018}. Their strategy is based on exploiting high performance computers and asynchronous computations to evaluate a large number of model architectures.\\
\indent In this article, we focus on the problem of ML  model selection and   hyperparameter optimization (HPO) for  a pressing application problem of  predicting daily groundwater levels in California (CA), United States, for groundwater wells that are affected by streamflow and climatic changes. Due to the non-stationary nature of the groundwater levels,  traditional time series \textit{forecasting} like ARIMA and \textit{regression} methods are not suitable.
We analyze  the performance of a long short-term memory recurrent neural network (LSTM), a multilayer perceptron (MLP), a simple recurrent neural network (RNN), and a convolutional neural net (CNN) because they represent the most widely used deep learning models and their suitability for hydrological timeseries data has not been studied thoroughly.
We compare the models based on  their prediction  accuracy and  the time it takes to find the best hyperparameters for each model. \\
\indent In order to find the best hyperparameters, we pose the HPO problem  as a bilevel  black-box  optimization problem.  The upper level problem objective function, the model performance, is not given in closed form as it requires solving a computationally expensive optimization lower level problem, the model fitting problem. The goal is to find the best hyperparameters (upper-level variables) with as few calls to the lower level problem as possible.
We solve this problem  by using two types of adaptive, derivative-free  surrogate model  algorithms, namely one based on radial basis functions (RBFs) and one based on  GPs. We compare the surrogate model approaches for HPO to the widely used method of randomly sampling the hyperparameter space.
We demonstrate the application of the surrogate model  approaches for a groundwater prediction  application. Here, we consider two cases: one, in which we use a single  groundwater observation well, and the second, in which we use multiple  groundwater observation wells for model training and predictions. \\
\indent The remainder of this paper is organized as follows. 
In Section~\ref{sec:problem}, we introduce the mathematical problem description of HPO. We describe the details of the surrogate model based optimization algorithms in Section~\ref{sec:surrogates}. In Section~\ref{sec:ML}, we briefly review the learning models we use in our  numerical experiments.  In Section~\ref{sec:numerical}, we provide the details of the groundwater prediction problem, and we describe the setup and the results of our numerical experiments. Finally, Section~\ref{sec:concl} concludes the paper and outlines future research directions.  Supplemental materials contain additional data and results of numerical experiments.


\section{Mathematical Problem Description of Hyperparameter Optimization}\label{sec:problem}
The problem of finding the best parameters  of a learning model can be stated as a bilevel optimization problem:
\begin{eqnarray}
&\min_{\bth, \bw^*} \ell(\bth, \bw^*;\mathcal{D}_\text{test}) \label{eq:up}\\
& \text{s.t. } \bth\in\Omega \label{eq:theta}\\
& \bw^*\in\arg\min_{\bw\in\mathcal{W}}L(\bw; \bth, \mathcal{D}_\text{train}).\label{eq:low}
\end{eqnarray}
Here, we divide  the dataset $\mathcal{D}$  into training data $\mathcal{D}_\text{train}$ and testing data $\mathcal{D}_\text{test}$. At the upper level, we optimize the $d$ hyperparameters (model architecture), $\bth=[\theta^{(1)}, \ldots, \theta^{(d)}]^T$ using the test data  $\mathcal{D}_\text{test}$ (equations~(\ref{eq:up}) and~(\ref{eq:theta})). At the lower level, we fit the model to the training  data  $\mathcal{D}_\text{train}$ by determining the optimal weights $\bw^*$ given the current network architecture $\bth$ (equation~(\ref{eq:low})). Note that there may be  more than one vector $\bw^*$ that solve problem~(\ref{eq:low}) for a given network architecture, however the python package we use to solve this lower level problem in our implementation  returns only  a single solution.

The set $\Omega$ is modeled as a discrete  finite set
\begin{equation}
\Omega =\prod_{i=1}^d \mathcal{I}_i
\end{equation}
where  $\mathcal{I}_i$ denotes a discrete set of values that can be assumed by  parameter $\theta^{(i)}$. The domain $\mathcal{W}$ over which the weights are optimized is continuous. We use the notation  $L(\mathbf{w}; \bth,  \mathcal{D}_\text{train})$ in~(\ref{eq:low}) to indicate that  we optimize over the parameters $\bw$ \textit{given} a set of parameters $\bth$ and the training data set $\mathcal{D}_\text{train}$.

Our assumption   that all hyperparameters can only assume  a limited number of values is motivated by practical considerations. Some parameters, such as the number of network layers  can only be integers. Other parameters, such as the dropout rate can range between 0 and 1. However,   in practice,  only  a finite number of options is considered for all hyperparameters due to limited resources and an assumption that very small changes in these hyperparameters are not likely to affect the performance of the learning model significantly.  

The loss functions $\ell$ and $L$ in equations~(\ref{eq:up}) and~(\ref{eq:low}) may be the same, for example, a root mean squared error between the data (actual value) and the fitted model  (estimated values), but they may also be different depending on the goal of the analysis. For example,  the upper level function could also include a regularization term to encourage model simplicity or low training and prediction times. 
In order to obtain the objective function value $\ell(\bth, \bw^*; \mathcal{D}_\text{test}) $ for a certain architecture $\bth$ at the upper level, we have to solve the fitting problem~(\ref{eq:low}) at the lower level. The lower level problem is usually a difficult   large-scale global optimization problem and stochastic optimizers  are commonly used to solve it. Due to the stochasticity of the optimization method,  solving problem~(\ref{eq:low}) twice for the same hyperparameters but with different random number seeds  will in general lead to two different solutions. Therefore, the upper  level problem can be considered as a stochastic optimization problem. In order to obtain a good estimate of $\ell$ for a given $\bth$,  the lower level problem should be solved several times,  and thus instead of minimizing $\ell$ at the upper level, we minimize its expected value $\mathbb{E}_\zeta[\ell(\bth, \bw^*(\zeta);\mathcal{D}_{\text{test}})]$ where $\zeta$ is a random variable that encodes the randomness due to the lower level  optimizer. Depending on the application, it might also be appropriate to minimize the worst case loss.  Thus, the  optimal solution $\bw^*$ as well as the best hyperparameter choice $\bth^*$ depend on $\zeta$. 

Solving the lower level optimization problem is in general time-consuming and, depending on the network architecture (defined by $\bth$), may require many minutes to hours (a large number of layers, nodes, and epochs  lead to long compute times at the lower level). This compute time requirement restricts how many times the lower level problem can be solved for each hyperparameter set in order to obtain statistically significant point estimates of the expectation value.  

Due to the problem structure described above, there is  no algebraic description of the upper level $\ell$ available (black-box, see, e.g.,~\cite{Audet2016} for a discussion of black-box optimization) and neither is gradient information. The loss function is generally multimodal with several local and global optima. Therefore,  HPO requires an efficient, gradient-free,  optimization method that can search both locally and globally in order to be able to find good solutions, yet be able to escape from local optima.

\section{Surrogate Models for  Efficient Hyperparameter Optimization (HPO)}\label{sec:surrogates}
Surrogate model-based derivative-free optimization algorithms have been developed to efficiently solve   computationally expensive black-box  problems. A surrogate model $m$ approximates  the costly objective function~\cite{Booker1999}: $\ell(\bth)=m(\bth)+e(\bth)$, where $e(\cdot)$ denotes the difference between the true objective and the surrogate approximation. Here we omit the dependence of $\ell$ on $\bw^*(\zeta)$ for ease of notation and because we are only interested in approximating the relationship between $\bth$ and the corresponding  value of $\ell$, in which  the optimization over the $\bw$'s is implied.  

Most surrogate model optimization algorithms have been developed for problems with continuous parameters, which are constrained only by lower and upper bounds, see, e.g.,~\cite{Gutmann2001, Regis2007b, Holmstrom2008b, Jones1998}. However,  there are some algorithms  for  problems with other  characteristics such as computationally expensive-to-compute constraints~\cite{Audet2006, Mueller2017, Nunez2018, Regis2011}, integer constraints on all parameters~\cite{Muller2013a}, mixed-integer parameters~\cite{Abramson2009, Davis2009, Holmstrom2008a,Muller2015, Muller2013},  multiple conflicting objectives~\cite{Datta2016, Mueller2017a}, hidden constraints~\cite{Gramacy2011, Lee2011, Mueller2019}, and multiple levels of objective function fidelity~\cite{Audet2010, Mueller2019b, Forrester2007}. For an overview of derivative-free and black-box optimization topics, we refer the reader to~\cite{Audet2017},

Many difficult optimization problems have been tackled successfully using surrogate model methods, e.g.,  in the structure optimization of nanomaterials~\cite{Osman2019}, in cloud simulations~\cite{Langhans2019}, in climate modeling~\cite{Muller2015a}; in  watershed water quality management~\cite{Mueller2017}, and in aerodynamic design~\cite{Toal2011}.   

In this article, we will use a surrogate model approach  for optimizing  the integer-constrained  hyperparameters of deep learning models.  We implement and compare two different  approaches: (a) a radial basis function (RBF) as surrogate and a weighted score to iteratively select sample points  (similar to SO-I~\cite{Muller2013a}); (b) a Gaussian process (GP) model and an expected  improvement~\cite{Jones1998} acquisition function, which we optimized over an integer lattice, using a genetic algorithm to iteratively generate a new sample point (a new set of hyperparameters $\bth$). 

Global surrogate model algorithms generally follow the same structure: (1) Generate an initial  experimental design  and evaluate the expensive objective function at the selected points; (2) Use all  input-output data pairs to compute the parameters of the surrogate model; (3) Optimize a computationally cheap auxiliary function on the surrogate model to determine the next point in the parameter space where the expensive objective will be evaluated. We iterate between steps (2) and (3) until a stopping criterion has been met. Natural stopping criteria include a maximum number of allowed function evaluations, a maximum CPU time, or a maximum number of failed consecutive improvement trials. In step (2), different types of surrogate models can be used, such as GP models~\cite{Matheron1963}, RBFs~\cite{Powell1992, Powell1999}, or polynomial regression models~\cite{Myers1995}. Different methods have been developed for iteratively selecting new sample points. For example,~\cite{Jones1998} maximized  an expected improvement criterion,~\cite{Regis2007b} used a stochastic sampling approach, and~\cite{Gutmann2001} minimized a ``bumpiness" measure. In this article, we compare the performance of  the expected improvement criterion and the stochastic sampling when all parameters are assumed to be integer values. The pseudocode of the hyperparameter optimization method is shown in Algorithm~\ref{alg:pseudo}, and the individual steps are described in more detail in the following subsections. 
\begin{algorithm}[htbp]
\caption{Algorithm for computationally expensive HPO  of learning models}
\label{alg:pseudo}
\begin{algorithmic}[1]\compresslist
\State{\textbf{Prepare hyperparameters for optimization:} Map the finite set of possible values using sequences of consecutive integers, denote this domain by $\Omega$.}\label{alg:prep}
\State{\textbf{Initial design:} Create an initial experimental design with $n_0$  points $\Bth=\{\bth_1, \ldots,\bth_{n_0}\}$ by randomly selecting integer-valued vectors from $\Omega$. For each $\bth_j\in\Bth$,  compute the upper level objective function value  ($\ell(\bth_j)$)  by solving the lower level problem~(\ref{eq:low}). }\label{alg:bev1}
\State{Set $n\leftarrow n_0$.}
\State{\textbf{Adaptive sampling:}}\label{alg:bev2}
\State{Use all input-output pairs $\{(\bth_j, \ell(\bth_j))\}_{j=1}^n$ to compute the parameters of the surrogate model. }\label{alg:rbf}
\State{Optimize an acquisition function to select the next evaluation point $\bth^{\text{new}}$.} \label{alg:bev3}
\State{Solve the lower level problem (\ref{eq:low}) to obtain the objective  function value  at $\bth^{\text{new}}$.}
\State{Set $n\leftarrow n+1$ and go to Step~\ref{alg:rbf}.}
\State{\textbf{Stop when the termination condition is satisfied.}}\label{alg:bev5}
\end{algorithmic}
\end{algorithm}

\subsection{Step 1: Preparing the Hyperparameters for Optimization}
When preparing the hyperparameters in Step~\ref{alg:prep}, we have to take into account that they  may live on very different ranges  and therefore we map them to sequences of consecutive integers. For example, if the first parameter can assume the values   $\{0, 0.1, \ldots, 0.5\}$ and the second parameter can assume the values $\{50, 100, 150\}$, then the search domain is  $\Omega=\{0, 1, \ldots, 5\}\times\{1, 2, 3\}$.
 This  decreases the difference in parameter ranges\footnote{Note that we usually use a similar approach in continuous optimization where we scale the parameters to the unit hypercube, which improves the surrogate models and eliminates difficulties when  sampling by perturbation.}. To evaluate the loss function, we map the parameters back to their original values, e.g., by multiplying with 0.1 and 50, respectively, for the two examples above.

\subsection{Step 2: Initial Experimental Design}
In Step~\ref{alg:bev1}, we generate $n_0$ randomly selected initial points from $\Omega$ at which we  evaluate the upper level objective function. These points satisfy the integer constraints by definition. For each point, we solve the lower level problem $N=5$ times and compute the upper level objective function as the average of $N$ evaluations.  These $N$ evaluations can be  done in parallel to reduce  the required  evaluation time. 

\subsection{Steps 4-9: The Adaptive Sampling Loop}
During the adaptive sampling loop, we compute the parameters of the surrogate model using \textit{all} previously evaluated points (which is different from trust region type methods that may  use a subset of the evaluated points).
Generally,  any surrogate model can be used in the algorithm. We consider RBFs and GPs in this article. In the following, we describe the details of these surrogate models and the associated adaptive sampling methods. 

\paragraph{Radial Basis Functions and Stochastic Sampling}
Our first approach (``RBF approach'') to the HPO of learning models is based on RBF models with a cubic kernel. These models have previously been shown to perform well for various  optimization problems including problems with integer constraints, see e.g.,~\cite{Osman2019, Mueller2017,Wild2013}. RBFs have the general form 
\begin{equation}
m_{\text{RBF}}(\boldsymbol\theta) = \sum_{j=1}^n \lambda_j \varphi(\|\boldsymbol\theta-\boldsymbol\theta_j \|_2)+p(\boldsymbol\theta),
\end{equation}
where $\boldsymbol\theta_j, j = 1,\ldots,n$, are already evaluated points, $\varphi(r)=r^3$ is the  RBF function with a cubic kernel (other kernels are possible, see e.g.,~\cite{Gutmann2001}), $\|\cdot\|_2$ denotes the Euclidean norm, and $p(\boldsymbol\theta) = \beta_0+\boldsymbol\beta^T\boldsymbol\theta$ is a linear polynomial tail. The parameters $\lambda_1, \ldots, \lambda_n, \beta_0$, and $\boldsymbol\beta=[\beta_1, \ldots, \beta_d]^T$ are determined by solving a linear system of equations: 
\begin{equation}
\begin{bmatrix}
\boldsymbol{\Phi} & \mathbf{P} \\
\mathbf{P}^T & \mathbf{0}
\end{bmatrix}
\begin{bmatrix}
\boldsymbol{\lambda}\\
\boldsymbol{\tilde{\beta}}
\end{bmatrix}=
\begin{bmatrix}
\bell\\
\mathbf{0}
\end{bmatrix},\label{eq: RBFsystem} 
\end{equation}
where the elements of the  matrix $\boldsymbol{\Phi}$ are  $\Phi_{{k,l}}=\varphi(\|\boldsymbol\theta_k-\boldsymbol\theta_l\|_2)$, $k,l=1,\ldots, n$, $\mathbf{0}$ is a matrix with all entries  0 of appropriate dimension, and
\begin{equation}
\mathbf{P} = \begin{bmatrix}
\boldsymbol\theta_{1}^T & 1\\
\vdots &\vdots\\
\boldsymbol\theta_{{n}}^T & 1\\
\end{bmatrix}
\quad
\boldsymbol{\lambda}=
\begin{bmatrix}
\lambda_1\\
\lambda_2\\
\vdots\\
\lambda_{n}
\end{bmatrix}
\quad
\boldsymbol{\tilde\beta}=
\begin{bmatrix}
\beta_1\\
\beta_2\\
\vdots\\
\beta_d\\
\beta_0
\end{bmatrix},
\quad
\bell=
\begin{bmatrix}
\mathbb{E}_\zeta[\ell(\boldsymbol\theta_1, \mathbf{w}_1^*(\zeta); \mathcal{D}_{\text{test}})]\\
\mathbb{E}_\zeta[\ell(\boldsymbol\theta_2, \mathbf{w}_2^*(\zeta); \mathcal{D}_{\text{test}})]\\
\vdots\\
\mathbb{E}_\zeta[\ell(\boldsymbol\theta_n, \mathbf{w}_n^*(\zeta); \mathcal{D}_{\text{test}})]
\end{bmatrix}.
\label{eq: RBFmatrices}
\end{equation}
 The matrix   in~(\ref{eq: RBFsystem}) is invertible if and only if  rank$(\mathbf{P})=d+1$~\cite{Powell1992}, where $d$ denotes the problem dimension. Here, $\bw_j^*$ is an optimal weight vector (lower level solution obtained by solving problem~(\ref{eq:low})) corresponding to the $j$th evaluated parameter vector $\bth_j$. 
 
In order to identify the next sample point, we find the point  $\bth^{\text{best}}\in\arg\min\{\mathbb{E}_\zeta[\ell(\bth_j,\bw_j^*(\zeta); \mathcal{D}_{\text{test}})],  j=1, \ldots,n\}$ that has the lowest objective function value (if there are multiple, we select one at random). Then,  we generate a large number $M=500$ of candidate points by perturbing each value of the best point found so far by either adding or subtracting a value randomly chosen from $\{1,2\}$. If the perturbed point falls outside of $\Omega$, we reflect it over the corresponding boundary to the inside of $\Omega$. The points  created by perturbation enable a local search around $\bth^{\text{best}}$. 
In addition, we generate a second set of candidate points by randomly selecting another $M$ points from $\Omega$. These points represent our global search. Thus, we have $2M$ candidate points in total.

In order to select the next evaluation point $\boldsymbol\theta^{\text{new}}$ from the candidate points, we follow the ideas of~\cite{Regis2007b} in which two scores are computed for each point, namely the distance to the  already evaluated points (distance score) and the  objective function value as predicted by the RBF surrogate (RBF score). We compute a weighted sum of both scores and the candidate point that achieves the best weighted sum will become the next evaluation point (for details, see the online supplement Section A).   By placing a large weight on the distance score, we  preferentially sample  at points that  are far away from already sampled points (global search). If the weight for the RBF score is large, we preferentially sample points with low predicted objective function values. This represents a local search, because points with low predicted objective function values are usually in the vicinity of the best point found so far. Note that candidates that have already been evaluated previously are discarded. Thus, if we did not have a stopping criterion and sampling continued, the globally optimal solution would eventually be found, following a  simple counting argument.

\paragraph{Gaussian Process Models and Expected Improvement}
Our second approach (GP approach) to the HPO of learning models is based on  GP models and  expected improvement sampling~\cite{Jones1998} over an integer lattice.
According to a literature review, GP (or kriging) models have been widely used as surrogate models to approximate expensive-to-compute objective functions. By definition, GP models not only provide a prediction of the function values at unsampled points, but also an estimate of the associated uncertainty. In kriging, we treat the function we want to approximate like the realization of a stochastic process, and write the surrogate as
\begin{equation}
    m_{\text{GP}}(\bth) = \mu + Z(\bth).
\end{equation}
Here, $\mu$  represents the mean of the stochastic process and $Z(\bth)$ is distributed as $\mathcal{N}(0,\sigma^2)$. Thus, the term $Z(\bth)$ represents a deviation from the underlying global model (the mean). We assume  a correlation between  the errors, which is based on the distance to the other points. At an unsampled point $\bth^{\text{new}}$, the kriging prediction is treated like a realization of a random variable $M_{\text{GP}}(\bth^{\text{new}})$ that is normally distributed with mean $\mu$ and variance $\sigma^2$. The correlation between two random variables $Z(\bth_k)$
 and $Z(\bth_l)$ is defined as
 \begin{equation}
     Corr\left(Z(\bth_k),Z(\bth_l)\right)=\exp\left( -\sum_{i=1}^{d} \bga_i |\theta_k^{(i)}-\theta_l^{(i)}|^{q_i} \right),\label{eq:corr}
 \end{equation}
 where 
 $\gamma_i$ determines how fast the correlation decreases in the $i$th dimension, and $q_i$ reflects the smoothness of the function in the $i$th dimension (we use $q_i=2$ for all $i$). We denote by $\Br$ the $n\times n$ matrix whose $(k,l)$th element is given by (\ref{eq:corr}). The kriging parameters $\mu, \sigma^2$, and $\gamma_i$ are estimated by maximum likelihood.
 Then, the prediction at a new point $\bth^{\text{new}}$ is defined as
\begin{equation}
    m_{\text{GP}}(\bth^{\text{new}})=\hat{\mu}+\br^T\Br^{-1}(\bell-\mathbf{1}\hat\mu),
\end{equation}
where $\mathbf{1}$ is a vector of ones of appropriate dimension and $\bell$ as in~(\ref{eq: RBFmatrices}) and
\begin{equation}
    \br=
    \begin{bmatrix}
    Corr\left(Z(\bth^{\text{new}}), Z(\bth_1)\right)\\
    \vdots\\
    Corr\left(Z(\bth^{\text{new}}
    ), Z(\bth_n)\right)
    \end{bmatrix}.
\end{equation}
The  corresponding  mean squared error is 
\begin{equation}
    s^2(\bth^{\text{new}})=\hat{\sigma}^2\left(   1-\br^T\Br^{-1}\br +\frac{(1-\boldsymbol{1}^T\Br^{-1}\br)^2}{\mathbf{1}^T\Br^{-1}\mathbf{1}}\right)
\end{equation}
with 
 \begin{equation}
     \hat{\mu} = \frac{\mathbf{1}^T\Br^{-1}\bell}{\mathbf{1}^T\Br^{-1}\mathbf{1}}
 \end{equation}
 and
 \begin{equation}
     \hat{\sigma}^2=\frac{(\bell-\mathbf{1}\hat{\mu})^T\Br^{-1}(\bell-\mathbf{1}\hat{\mu})}{n}.
 \end{equation}
From the properties of the GP model, we can now define an expected improvement function which we use to iteratively select a new sample point $\bth^{\text{new}}$. We denote by $\mathcal{L}$ a random variable that represents our uncertainty about the function value.  Its mean and variance at a point $\bth$ are defined by the kriging predictor, namely $m_{\text{GP}}(\bth)$ and $s^2(\bth)$. We denote by $\ell^{\text{best}} = \ell(\bth^{\text{best}})$ the best function value we have found so far during the upper level optimization. If the value  $\mathcal{L}$ is less than $\ell^{\text{best}} $, we have an improvement $I=\ell^{\text{best}}-\mathcal{L}$. The expected value of this improvement is computed as 
\begin{equation}
    \mathbb{E}(I) = s(\bth)\left(v\Phi(v)+\phi(v)   \right),
\end{equation}
where 
\begin{equation}
    v=\frac{\ell^{\text{best}}-m_{\text{GP}}(\bth)}{s(\bth)}
\end{equation}
and $\Phi$ and $\phi$ are the normal cumulative distribution and density functions, respectively, and $s(\bth)=\sqrt{s^2(\bth)}$ is the squared root of the mean squared error.  

In order to select the next evaluation point $\bth^{\text{new}}$, we have to maximize the expected improvement. Since the maximum argument will not necessarily fall onto a point that satisfies the  integer constraints, we have to maximize the function over an integer lattice. Therefore, we use a genetic algorithm (see~\cite{Mitchell1996} for introducing genetic algorithms), which returns an integer-feasible point $\bth^{\text{new}}\in\Omega$ at which we do the next expensive  function evaluation.

\subsection{Algorithm Parameters}
Our algorithm has several parameters that can be adjusted and may impact the results of our simulations. In our numerical experiments,  we use $n_0= d+1$ points as the  number of initial experimental design points. Generally, we would like to use a large number, but the more initial points we use, the fewer points can be acquired iteratively (as the bottleneck is usually the number of allowed evaluations). For each hyperparameter vector, we solve the lower level problem $N=5$ times to obtain an estimate of the expected value of the outer  objective function. The value of $N$ should be adjusted based on the available compute resources.  

 In each iteration, we generate $2M$ candidate points, with $M=500$. If the number of hyperparameters  and the number of values each hyperparameter can assume are small, we can alternatively generate candidate points by  completely enumerating  all possible solutions for which we would then compute  the weighted scores  or  the  expected  improvement values, and choose the best candidate, but in our HPO setting, this is not applicable due to the huge number of possible solutions (see also Table~\ref{tab:hyp} in Section~\ref{sec:setup}).

The weights for the  selection criteria in the RBF approach cycle through the pattern $\{0, 0.25, 0.5, 0.75, 1\}$.  In the first iteration, the weight is 0; in the second iteration, the weight is 0.25, and so on, and after the weight 1 was used, it will be set back to 0 in the next iteration. This cycling enables the sampling to transition from searching locally to globally. 

Another important parameter is the set of randomly chosen perturbations $\{1,2\}$ that are added or subtracted to the current best solution. The goal of the perturbation is to enable a local search and given that our parameter ranges are relatively small, using the values  $\{1, 2\}$ is appropriate and a maximum perturbation of 2 is reasonable to prevent creating too many candidate points that are outside the upper and lower bounds. If larger parameter ranges were present, larger perturbation values could be chosen. 

In general, the amount of available compute time and the number of hyperparameters to be optimized  dictate  the algorithm parameter values for $N, M$, and the maximum number of allowed upper level evaluations. We use a maximum number of 50 upper level  function evaluations as stopping criterion.

\section{Review of Learning Models}\label{sec:ML}

In this section, we provide a brief review of the different deep learning models that we are using in our numerical study and their associated hyperparameters. These hyperparameters  determine the model architecture, and thus  how well the model performs in terms of minimizing the loss. 

The number of layers, the number of nodes in each layer, the type of activation functions, and the type of optimizer are examples of hyperparameters.
Each node in a network layer may have a different activation function. The Rectified Linear Unit (\textsf{ReLU}) activation function, $f(\cdot)=\max(0,\cdot)$, is  commonly used.
The learning model corresponding to  a given architecture ($\bth$)  is trained  to find the weight vector ($\bw^*$) that minimizes the loss $L$.
The  optimizers used for this task share the idea of backpropagation \cite{rumelhart1988learning}  which is an iterative method based on gradient descent and the chain rule.
Therefore, the number of iterations (=the number of epochs) and the batch size (=the number of samples in the gradient update) are additional hyperparameters that must be determined.
Among many existing optimizers, ADAM~\cite{Kingma2015}, Adamax, and RMSprop~\cite{hinton2012neural} are widely used.
Each of these optimizers has their own hyperparameters such as the learning rate and the decay rate. 
Finally, a dropout rate ($\alpha \in (0,1)$) can be set to help prevent overfitting at each layer except for the output layer. The dropout rate means that randomly selected $\alpha\times$(number of nodes in the layer) nodes are ignored during training. 

\subsection{Multilayer Perceptron Model (MLP)}
\label{ssec:MLP}
The MLP~\cite{bishop1995neural} is a feed-forward neural network
that consists of an input layer, at least one hidden layer, and an output layer.
It can model nonlinear patterns by introducing multiple layers and nonlinear activation functions that transform input signals. 
It has been primarily used for classification and function estimation problems
\cite{trenn2008multilayer}. MLPs have been used for, e.g., forecasting drought \cite{ali2017forecasting},  estimating evapotranspiration \cite{tabari2013multilayer}, and predicting water flow \cite{araujo2011multilayer}. 
Unlike other statistical models, the MLP does not make any assumptions regarding the prior probability density distribution of the training data \cite{gardner1998artificial}.

\subsection{Convolutional Neural Networks (CNN)}

The CNN \cite{lecun1998gradient} is also a feed-forward neural network that applies a sliding window filter on the input dataset, and has three main characteristics, namely local connectivity, shared weight, and spatial or temporal sub-sampling. 
The local connectivity means that the weighted linear combination 
is restricted to within the filter (\textit{not fully connected}). 
All local fields 
share the same weights. 
The weighted values are transformed with the activation function $f(\cdot)$, and the function returns a feature map. 
The sub-sampling is an optional step to extract other features in the feature map by reducing its size. An example method of sub-sampling is to pool the data to replace the input by a summary statistic such as the maximum (max-pooling) or mean (avg-pooling) value. The filters can also be used to extract   features. The filter can be applied to both the input and the feature map to make deeper neural networks.
Finally, the extracted features are fully connected to the output layer. Note that additional hidden layers  can be constructed between the final feature layer and the output layer. 
Additional CNN specific hyperparameters that must be optimized include the number of filters, the filter size, whether or not to apply sub-sampling, and the method and the size of the sub-sampling.

In practice, CNN's have been used for a variety of application areas ranging from image recognition (e.g., ResNet \cite{he2016deep}) to time series 
classification \cite{cui2016multi}, and time series forecasting \cite{borovykh2017conditional}.

\subsection{Simple Recurrent Neural Networks (RNN)}
Recurrent Neural Networks (RNNs)  can represent dynamic temporal behavior by using a directed graph structure between nodes to store sequence representations. The decision an RNN sees at time $t-1$ affects the decision at time $t$. Unlike feed-forward neural networks, which  take  the entire sequence input as is, the RNN first takes in  the output generated from the previous sequence steps (memory) before taking the input step. 
 In the simplest RNN, the memory is carried forward through the hidden state, which is a function 
of  the product of the input 
at time $t$  and a weight matrix 
that is added to a product of the hidden state at time $t-1$  
and another hidden weight matrix. 
The output 
is a function 
of the product of the last  hidden state 
and a weight matrix.  
The activation function in the RNN 
is a hyperbolic tangent function that allows an increase or decrease in the value of the state. The weight matrices 
are adjusted  by error backpropagation. 
A drawback of RNNs is that often, despite choosing the optimal set of hyperparameters, errors flowing ``backward in time'' in a fully connected RNN may  either vanish or blow up, which leads to oscillating weights or long training times.
RNNs have been  used in applications like speech recognition \cite{mikolov2010recurrent}, text generation \cite{sutskever2011generating}, and time series forecasting in hydrology \cite{chiang2004comparison, coulibaly2001artificial}, where system memory and context are critical to prediction.

\subsection{Long Short-Term Memory Recurrent Neural Networks (LSTM)}

Long short-term memory networks~\cite{Hochreiter1997}  are an improvement of RNNs and are able to learn long-term dependencies. LSTMs have a  structure that allows them to represent temporal behavior and remember previous behavior through a special arrangement of   modules.   Besides  a hidden state, LSTMs also have a cell state that runs through the entire chain and is responsible for storing the long term dependencies. 
The information flow between the cell state, the hidden state, and the input state is controlled through different types of gates. A simple LSTM unit contains  an input gate, an output gate, and a forget gate. The forget gate~\cite{Gers2000}  enables the network to overcome the potentially unbounded growth of cell states when a continuous time series is used.  
The forget gate 
is a sigmoid activation function that decides how much  information from the previous cell state should be passed  to the next.  The output from the forget gate 
is a function 
of the  input,  the previous hidden states with their corresponding weight matrices, 
and  a bias vector. It enables resetting  the  memory blocks once their contents are out of date. 

The input gate  decides how to update the cell state.  The  hyperbolic tangent activation function  generates new candidate solutions using the weighted input and  hidden state 
and a bias vector. 
The new cell state is generated from the weighted sum of the previous cell state and the input state.  The output gate generates the new hidden state from the new cell state and the input carrying the most recent information.

LSTMs have been used in many different applications, e.g,  for language modeling~\cite{Sundermeyer2012}, speech recognition~\cite{Graves2013}, 
traffic speed prediction~\cite{Ma2015}, and  time series modeling~\cite{Hsu2018}.

\section{Numerical Experiments for Timeseries Data}\label{sec:numerical}

 In the following, we describe our groundwater application problem, the data used in the numerical experiments, and the setup and results of the experiments. Additional results and data are provided in the online supplement. 

We designed numerical experiments  to answer  two main questions: (1)~Which learning model is best suited for making accurate predictions of groundwater levels?  (2) Which  HPO method leads to the best results fastest? 
In order to answer these questions, we use our HPO methods to train  and test the learning models on historical observations  and we compare the performance of the four learning models to each other. We  use the optimized model architectures to predict the future groundwater levels and we compare these predictions to  ``future'' data, i.e., groundwater data that was not used for optimizing the  hyperparameters or training the models.

\subsection{Motivation and Description of the Groundwater Prediction Application}

In California (CA), United States, groundwater has always been an important resource for agriculture, drinking, and other beneficial purposes.  An unprecedented multi-year drought endured in the recent past (2012-2016) led to regulations on long-term water conservation and management of groundwater (see the 2014 Sustainable Groundwater Management Act (SGMA)~\cite{SGMA} in CA).

However, this situation is not unique to CA, and drought severity is expected to increase over many regions with changes in climate~\cite{Cook2018} increasing the agricultural dependencies on groundwater \cite{Taylor2010}.
There is therefore an increased urgency for time series forecasting of groundwater depths.

Groundwater depths are typically estimated using mechanistic multi-scale, multi-physics simulation models (see, e.g.,~\cite{langevin2017documentation, steefel2015reactive,xu2011toughreact}).
However, these models require extensive characterization of hydrostratigraphic properties and accurate quantification of boundary conditions, including recharge sources, climate variability as well as changes in pumping~\cite{sahoo2017machine}, which are not always known apriori.
Moreover, these models' computational complexity does  not make them  desirable when   groundwater predictions for multiple future climate scenarios  are needed  to make actionable decisions for  sustainable groundwater management. Thus, it is appealing to use  purely data-driven  machine learning methods that are computationally inexpensive and that  can be continually updated as new observations are obtained. Moreover, in some cases data-informed learning models have shown better performance than process-informed simulations, possibly due to deficiencies in model structure from incomplete process representations~\cite{Karandish2016}.

In the groundwater modeling context, artificial neural networks (ANNs) have previously been  used for both modeling and prediction purposes (see e.g.,~\cite{daliakopoulos2005groundwater}). 
More recently, RNNs have been used to predict groundwater levels  (e.g.,~\cite{coulibaly2001artificial, Zhang2018}). The application of LSTMs is still limited in the field of hydrologic sciences. For example, LSTMs have recently been used to predict changes of river discharge from precipitation across the Continental United States using publicly available datasets \cite{kratzert2018rainfall}.~\cite{Zhang2018} use an LSTM to predict groundwater levels in the Hetao Irrigation District in China. 

In this article, we focus in particular on  the  Butte County watershed  in CA.  This  area belongs to CA's  Central Valley, which is one of the largest agricultural lands in the world and produces over 250 crops a year~\cite{Faunt2009}. 
Butte County is considered a high-priority watershed by SGMA and  it is therefore vital to develop tools for groundwater predictions that will enable actionable and timely decision support for the planning and management of groundwater resources in this region in the future.

\subsection{Observation Data in Butte County, CA, USA}

At the Butte County watershed, we have high-frequency (daily) observations of groundwater levels  available  for the duration of approximately eight years (2010-2018). We train our  learning models on historic daily observations  of temperature, precipitation, streamflow, groundwater levels, as well as  features such as the week and month of the year that  the measurement was taken.  

Since there were gaps in the observations for temperature, precipitation, and groundwater levels (e.g., which may happen due to faulty sensors or bad data quality), we imputed the missing values  using the R package \textsf{imputeTS} ``Time Series Missing Value Imputation''~\cite{Moritz2017}.  This package provides state-of-the-art imputation algorithms along with plotting functions for univariate time series missing data statistics. We  use the \textsf{na.seadec} function with the Kalman Smoothing and State Space Models algorithm. This algorithm removes the seasonal component from the time series, performs imputation on the deseasonalized time series, and then adds the seasonal component again. To create an uninterrupted time series (due to some faulty data entries at the end of the hydrological year, i.e., end of September), we used the \textsf{na.locf} algorithm ("Imputation by Last Observation Carried Forward") of the \textsf{imputeTS} package. 

In the following study, we consider two scenarios, namely a``single well case''  and ``multi-well case''.  In the single well case, we train the model on the  groundwater levels measured only at a single well  and we predict only for this one well. In the multi-well case,  we use the groundwater level  data from  three different wells  to train the learning models, and we make predictions for all three wells. Note that  we do \textit{not} train a separate model for each well, but rather we have one model for all three wells.  
A summary of the data sources we use in our study as well as the  time series for all variables  
are shown in the online supplement in Section~B. The time series for the groundwater levels at all wells reflect  the significant decline in the amount of available groundwater between 2010 and 2016.

\subsection{Setup of the Numerical Experiments}\label{sec:setup}
We conducted our numerical experiments with python ($version~3.7$) on Ubuntu 16.04 with Intel\textsuperscript{\tiny\textregistered} Xeon(R) CPU E3-1245 v6 @ 3.70GHz $\times$ 8, and 31.2 GiB memory. We use the \textsf{Keras} package~\cite{Chollet2015} with the TensorFlow~\cite{abadi2016tensorflow} backend for our deep learning architectures.  For the genetic algorithm in the GP approach we use the \textsf{DEAP} python package~\cite{DEAP}. In the genetic algorithm we use \textit{100 generations} with \textit{100 individuals} each and we use a \textit{crossover probability of 0.75}. For the GP itself, we use the implementation in \textsf{Sklearn}~\cite{sklearn}.

\paragraph{Preparing the Data}
We divide our observation data into two sets -- one for training the models ($\mathcal{D}_{\text{train}}(\bth)$) and one for testing the models ($\mathcal{D}_{\text{test}}(\bth)$), i.e.,  comparing the model's groundwater predictions to the true observations. 
In our setting, there is a dependency of the number of daily observations used for  training and  testing the models    on a hyperparameter that we are optimizing. This dependency is described in more detail in  the following.   The model output  is the groundwater level for the next day and the inputs are the values of all the  variables  at the current day and all values at some previous days.
  More specifically, we define the number  of previous days whose data are used as input  as a \textit{lag number}, $G$. The lag number $G$ means that the variable values of the  $G$ previous days are used.
  This lag thus determines the size of $\mathcal{D}_{\text{train}}(\bth)$ and $\mathcal{D}_{\text{test}}(\bth)$. 
  To illustrate it with an example, suppose that there are a total 5 days of observations and for each day we have measurements of  6 variables.
  We denote  $\delta(t)$,  $t=1,\ldots,5$, as the day of the measurement.
  If the lag $G=1$, each sample consists of two days of observations (the current and the previous day) and therefore has 2$\times$6 =12 variable  values. The total number of samples we obtain is thus four: $(\{\delta(1), \delta(2)\}, \{\delta(2),\delta(3)\}, \{\delta(3),\delta(4)\}, \{\delta(4),\delta(5)\})$.  
  If the lag $G=2$, we use the measurements for each variable at  the two previous days in addition to the data for the current day, and thus  each sample consists of three days and has 3$\times$6=18 variable values. The total number of  samples we have is then  three: $(\{\delta(1),\delta(2),\delta(3)\}, \{\delta(2),\delta(3),\delta(4)\}, \{\delta(3),\delta(4),\delta(5)\})$.    
 Thus, as is, the lag would determine  the total number of samples for our model input. However,  we divide the dataset such that  the  number of training samples is the same  regardless of the lag.   
 In other words, the number of samples in $\mathcal{D}_{\text{train}}(\bth)$ is always the same and 
 the dimension of each sample (= the number of days $\times$ the number of variables in each day) varies depending on the lag.
 The samples that were not used for defining $\mathcal{D}_{\text{train}}(\bth)$ constitute  the test set  $\mathcal{D}_{\text{test}}(\bth)$.
 Therefore,  the lag determines how much historical information will be used in each sample. By setting the lag as a hyperparameter, the optimization aims to  not only find the best network  architecture, but also the optimal amount of  historical information in each sample that leads to  minimizing the outer objective function.
 
  In order to make future groundwater predictions, we use the given  values (here from the dataset $\mathcal{D}_{\text{test}}(\bth)$) of temperature, precipitation, and streamflow discharge and predict the corresponding  groundwater levels with the trained models. 
Because the input groundwater data in $\mathcal{D}_{\text{test}}(\bth)$ are assumed to be unknown, we use the predicted groundwater level value from the model. Therefore, the input groundwater data are dynamically obtained as we make predictions for each additional day.

 Before using the variable values  for training and testing our learning models, we normalize all variables' values to [0,1]. The reason is that  different observed variables affect the groundwater level in  specific ways and their values  are generally on different ranges. Scaling them to [0,1] enables us to better capture  the variables' interactions within the learning models. For the groundwater data,  this is, however, not  straight-froward. The recent drought in CA has caused the  groundwater levels to drop to unprecedented low levels.  Thus, for scaling purposes, it is difficult  to assign  a maximum and a minimum groundwater level as future groundwater levels may go below or above  recorded  values.  We therefore   assigned a minimum groundwater level of 0, which  here refers to the mean sea level and it is a fair assumption that the water level can never drop below 0 even though this value has never been observed (unless  it is located near the sea coast). For the upper bound, we use the maximum observed groundwater level. The reason is that over the past, we have observed a steadily decreasing trend in the amount of groundwater. Although wet years may cause the groundwater levels to go above the maximum recorded values,  it is a good choice as it reflects the reality observed so far.  Another possible upper limit  could be  the ground surface elevation (GSE), as groundwater cannot go above the ground. However, the GSE is also a dynamic parameter, as several places in CA  suffer from land subsidence due to groundwater withdrawal.    For the  other  variables (temperature, precipitation, streamflow discharge) the scaling is trivial as we have full information of the minimum and maximum  values of the whole time series (past observations as well as future predictions).  For the streamflow discharge, we applied a log-transformation before scaling the data to [0,1] because of the huge variations that occur (compare the time series in the online supplement).
 Note that the  scaling  does not mean that the predicted groundwater levels will be  below the chosen maximum, and the future groundwater  level predictions can go beyond their historical maximum levels depending on the input values of the other variables.

 \paragraph{Replications of Inner and Outer Optimizations and  Hyperparameters} 
The training of the  models  in the inner optimization and the search for the optimal hyperparameter  in the outer optimization both have  stochastic elements. For performing the inner large-scale optimization for a given hyperparameter set, we use  stochastic optimizers.   Therefore, given a specific hyperparameter set $\hat{\bth}$, we solve the inner problem five times and we  define the inner optimization's  loss function as the average  over these five trials (as an approximation of $\mathbb{E}_\zeta[\ell(\hat{\bth}, \bw^*(\zeta);\mathcal{D}_{\text{test}}(\hat{\bth}))]$). The outer optimization also has stochastic elements (in the initial experimental design and in the adaptive sampling methods).  Therefore, in order to obtain an average outer optimization performance, we repeat the HPO for each model with each algorithm five times. We allow each HPO method to try  50 different hyperparameter sets.

The hyperparameters we optimize  as well as the possible (unmapped) values for each hyperparameter  are summarized in Table~\ref{tab:hyp}. We keep  the learning and the decay rates at their default values of $0.001$ and $0.0$, respectively. From a practical perspective, allowing a lag of 365 days is reasonable as this would allow the model to use the information from a whole year worth of data, and thus recognize seasonality and correlations between the time of the year and the groundwater levels. However, for RNN and  also for LSTM, a maximum lag of 365 days was too large, causing problems with backpropagation in training and thus problems in the optimization. We were able to run the LSTM and RNN five times successfully by  restricting the maximum lag to 60 days for the single well case. 

Although the structure of the networks and their detailed mechanics are different (see Section \ref{sec:ML}), the number of input/output nodes are formed in the same way.
The number of input nodes are the same as the number of variables in a sample.
For the earlier example with five days and  lag $G=1$, all models would have 12 input nodes. The number of output nodes are the same as the number of wells whose groundwater data we are learning, i.e., one for the single well case and three for the multi-well case.

\begin{table}[htbp]
\caption{Hyperparameters  in the optimization and their possible values. NA means that a model does not have that particular hyperparameter. \# indicates ``number of''. LSTM=long short-term memory network; MLP=multilayer perceptron; RNN=recurrent neural network; CNN=convolutional neural network.}
\label{tab:hyp}
\begin{adjustbox}{width=\textwidth}
\begin{tabular}{p{2cm}llll}
\hline\noalign{\smallskip}
        & LSTM      & MLP       &   RNN     &CNN\\
\hline\noalign{\smallskip}
\#Epochs &  $\{50, 100, \ldots, 450, 500\}$ &      $\{50, 100, \ldots, 450, 500\}$& $\{50, 100, \ldots, 450, 500\}$& $\{50, 100, \ldots, 450, 500\}$    \\
Dropout rate     &    $\{0.0, 0.1, \ldots, 0.5\}$ &    $\{0.0, 0.1, \ldots, 0.5\}$ &    $\{0.0, 0.1, \ldots, 0.5\}$   &    $\{0.0, 0.1, \ldots, 0.5\}$ \\
Batch size      &  $\{50, 100, \ldots, 450, 500 \}$ &    $\{50, 55, \ldots, 195, 200 \}$    &    $\{50, 55, \ldots, 195, 200 \}$  &    $\{50, 55, \ldots, 195, 200 \}$\\
\#Layers        &  $\{1, 2,3, 4\}$  & $\{1, 2, 3, 4, 5, 6\}$ & $\{1, 2, 3, 4, 5, 6\}$ & $\{1, 2, 3, 4, 5, 6\}$       \\
Lag     &  $\{3,6,9, \ldots, 60\}$ & $\{30, 35, \ldots, 360, 365\}$& $\{3,6,9, \ldots, 60\}$&   $\{30, 35, \ldots, 360, 365\}$      \\
\#Nodes per hidden layer            &  $\{3, 6, \ldots, 27, 30 \}$ &   $\{5, 10, \ldots, 45, 50 \}$ &   $\{5, 10, \ldots, 45, 50 \}$ &NA       \\
\#Filters &  NA &NA & NA & $\{5, 10, \ldots, 45, 50\}$  \\
Filter size  &  NA &NA & NA &     $\{2, 3, 4, 5\}$\\
\hline\noalign{\smallskip}
Possible combinations &     480,000  &      7,588,800 &       2,232,000 &  30,355,200       \\
\noalign{\smallskip}\hline
\end{tabular}
\end{adjustbox}
\end{table}

For LSTM, we use the \textsf{ReLu} activation function  in all the deeper layers except the outer layer. For training the LSTM for a given hyperparameter set, we use the \textsf{RMSprop} optimizer. In the MLP, we also use the \textsf{ReLu} activation function for all nodes except those in the output layer. 
The nodes of the output layer have the linear activation function because we are dealing with real-valued outputs.
We use \textsf{ADAM} for  the lower level optimization. During the construction of the CNN, we add a fully connected hidden layer with 50 nodes between the last convolution layer and the output layer.
For simplicity, we do not apply the sub-sampling option here. 
The  activation function and the lower level  optimizer for the CNN and the RNN are the same as for the MLP.
We leave the dropout rate of the recurrent step fixed at 0.   
Even with our simplifying assumptions that all layers in the model  have the same hyperparameter values for    the dropout rate, the number of nodes, the number of  filters, the filter size, and the large step sizes (e.g., steps of 50 for the number of epochs),  we can see from Table~\ref{tab:hyp} (bottom)  that  the  number of possible hyperparameter combinations ranges from hundred thousands to millions. Trying all of these combinations is computationally infeasible, especially in our setting where optimal hyperparameters have to be determined fast to enable timely groundwater management decisions.

\subsection{Numerical Results}
In the following subsections, we compare the  results of our numerical experiments  obtained with LSTM, MLP, CNN, and RNN  for the single well case. For the multi-well case, we compare only  LSTM, MLP, and CNN. 
We did not include the RNN model for the multi-well case as it suffers from the problems with backpropagation mentioned earlier. It  therefore often fails to fit a model to a given  architecture. Also, as we will see for the single well results, the RNN performed worst among all learning models. Therefore, we do not include the RNN in the multi-well analysis.

 We compare the performance of our two optimization approaches (the RBF approach and  the GP approach) with a  random sampling (RS) approach for HPO. 
We illustrate our results in the form of progress plots,  time series plots, wall clock time plots, and  plots of the distributions of the optimal hyperparameters found in each of the five trials. In addition, we illustrate the models' predictive performance for data in the future that was not used during optimizing the hyperparameters or during training the models. This ``out-of-sample prediction'' allows us to gain insights into the models' extrapolation performance.   We  report additional results including tables with the optimal hyperparameter values for each model and each optimization strategy as well as the  corresponding losses (and the resulting groundwater prediction errors) in the online supplement in Section~C.

\subsection{Results for Predicting Groundwater at a Single Well in  Butte County, CA}

\begin{figure}[htbp]
\centering
    \includegraphics[scale=.2]{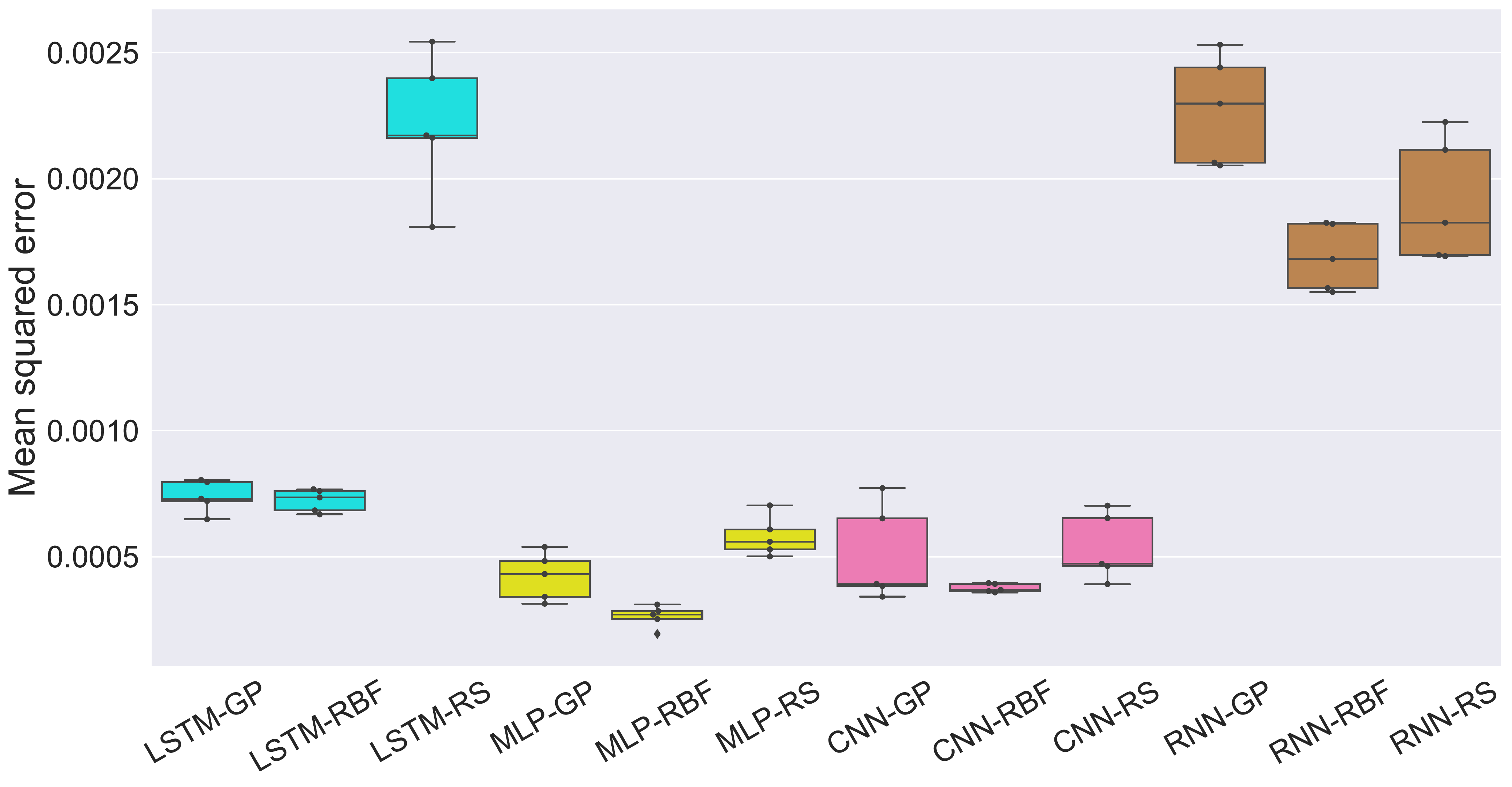}
    \caption{Boxplots of  mean squared errors for the single well case over five HPO trials. Low values are better.  LSTM=long short-term memory network; MLP=multilayer perceptron; CNN=convolutional neural network; RNN=recurrent neural network; GP=Gaussian process; RBF=radial basis function; RS = random sampling.}
    \label{fig:single_well_box_plot}
\end{figure}

\paragraph{Comparison of  Mean Squared Errors}
In Figure~\ref{fig:single_well_box_plot} we show the distribution of the optimal outer objective function values over  five trials  for all models and all optimization methods after the optimization has finished. The final  mean squared errors (MSEs) are fairly low for all learning models and all optimization approaches. In the online supplement in Table~2, we show the numerical MSE values for all trials as well as the corresponding deviation of the predicted groundwater level from the truth. Almost all trials of all models reach a deviation of about 2 meters or less, which is considered highly accurate. Figure~\ref{fig:single_well_box_plot} shows that there is a noticeable difference between the performance of the types of learning models,  with MLP and CNN generally leading to the smallest (best) final MSEs. We also observe a difference in the variation of the results over the five trials with RBF leading to the smallest variation for all HPO methods.  The   RNN performs worst regardless of the  HPO method used. This may be related to the fact that all HPO methods have difficulties in finding optimal solutions or that   the RNN is simply   not able to capture the groundwater data any better.  
Figure~\ref{fig:single_well_box_plot}  also indicates  that  the RBF approach generally  leads  to the lowest errors among all HPO methods. 

\begin{figure}[htbp]
    \centering
    \includegraphics[scale = .2]{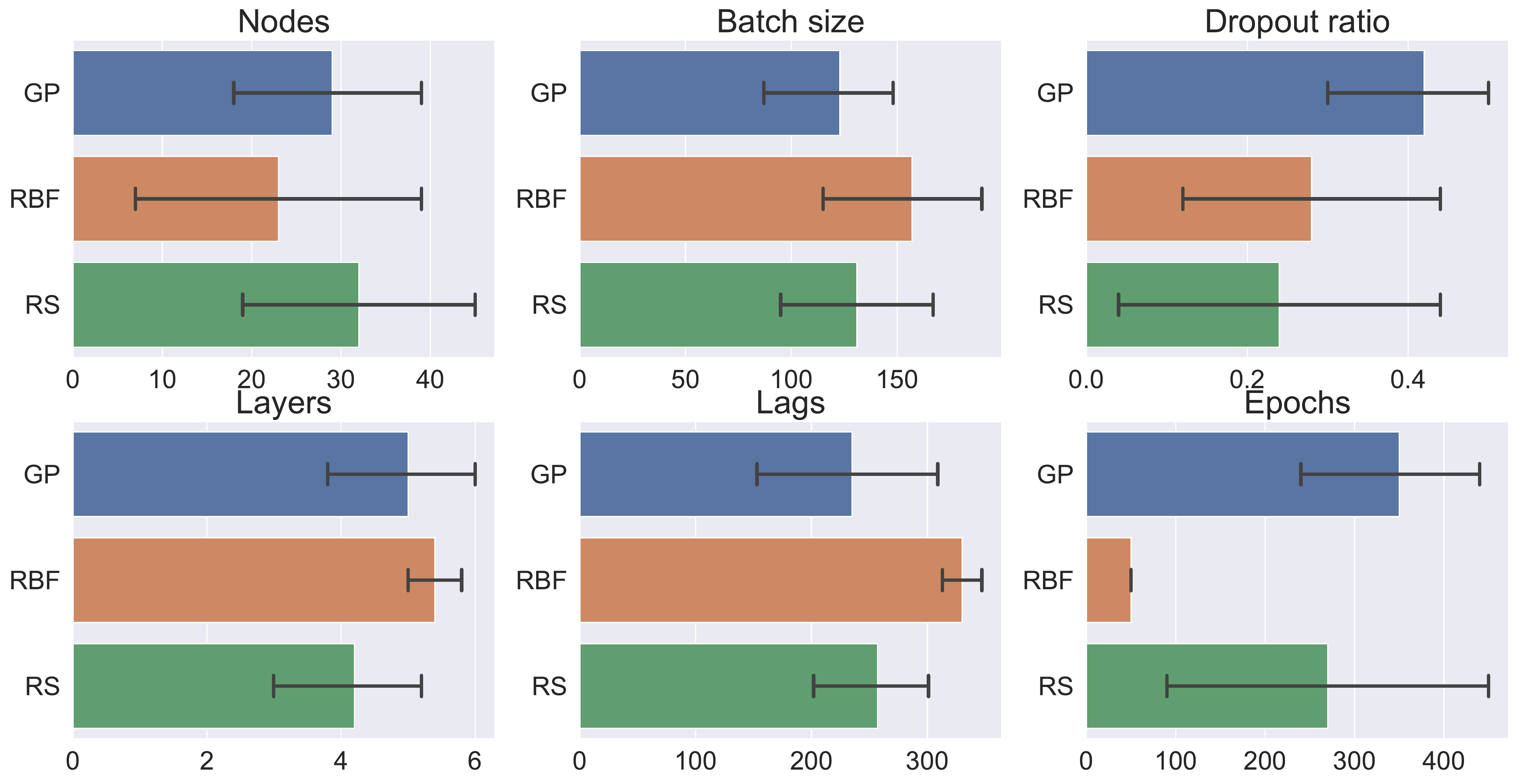}
    \caption{Distribution of the optimized hyperparameter values for MLP for the single well case. The colored bars represent the  mean and the black lines show the standard deviation over   five solutions.    LSTM=long short-term memory network; MLP=multilayer perceptron; CNN=convolutional neural network; RNN=recurrent neural network; GP=Gaussian process; RBF=radial basis function; RS = random sampling.}
    \label{fig:single_well_MLP_hyperparameter}
\end{figure}

\paragraph{Analysis of Optimized Hyperparameters}
We  compare the distributions of the final hyperparameter values over all five trials obtained with  the different HPO methods.   Figure~\ref{fig:single_well_box_plot} showed that the MLP optimized with the RBF approach leads overall to the best results.  Figure~\ref{fig:single_well_MLP_hyperparameter} shows the corresponding hyperparameters. Here, the colored bars represent the  mean and the black lines show the standard deviation over the  five solutions for the MLP. We see that generally all HPO methods lead to a more complex MLP network, the RBF method leading to approximately 5 layers with approximately  25 nodes each.   For the RBF method,  the lag is large, using almost a full year of past data. The dropout rate and the optimal number of epochs for the RBF method are smaller than  those obtained with the GP. Note that larger dropout rates discourage overfitting of the models to the data. We observe the largest differences between the HPO methods for the number of epochs. 

The distributions of the optimal hyperparameters for LSTM, CNN, and RNN are shown  in the online supplement  in Section~C.1.1 in Figure 3. For the CNN, we observe a similar behavior as for the MLP: the optimal RBF solutions have a large number of layers, filters, and lags. For LSTM, all HPO methods prefer smaller, less complex networks, and for the RNN, the HPO methods return  networks of different complexities  with generally larger standard deviations  on the optimal hyperparameter values.   

\begin{figure}[htbp]
    \centering
    \begin{subfigure}[b]{0.48\textwidth}
        \includegraphics[width=\textwidth]{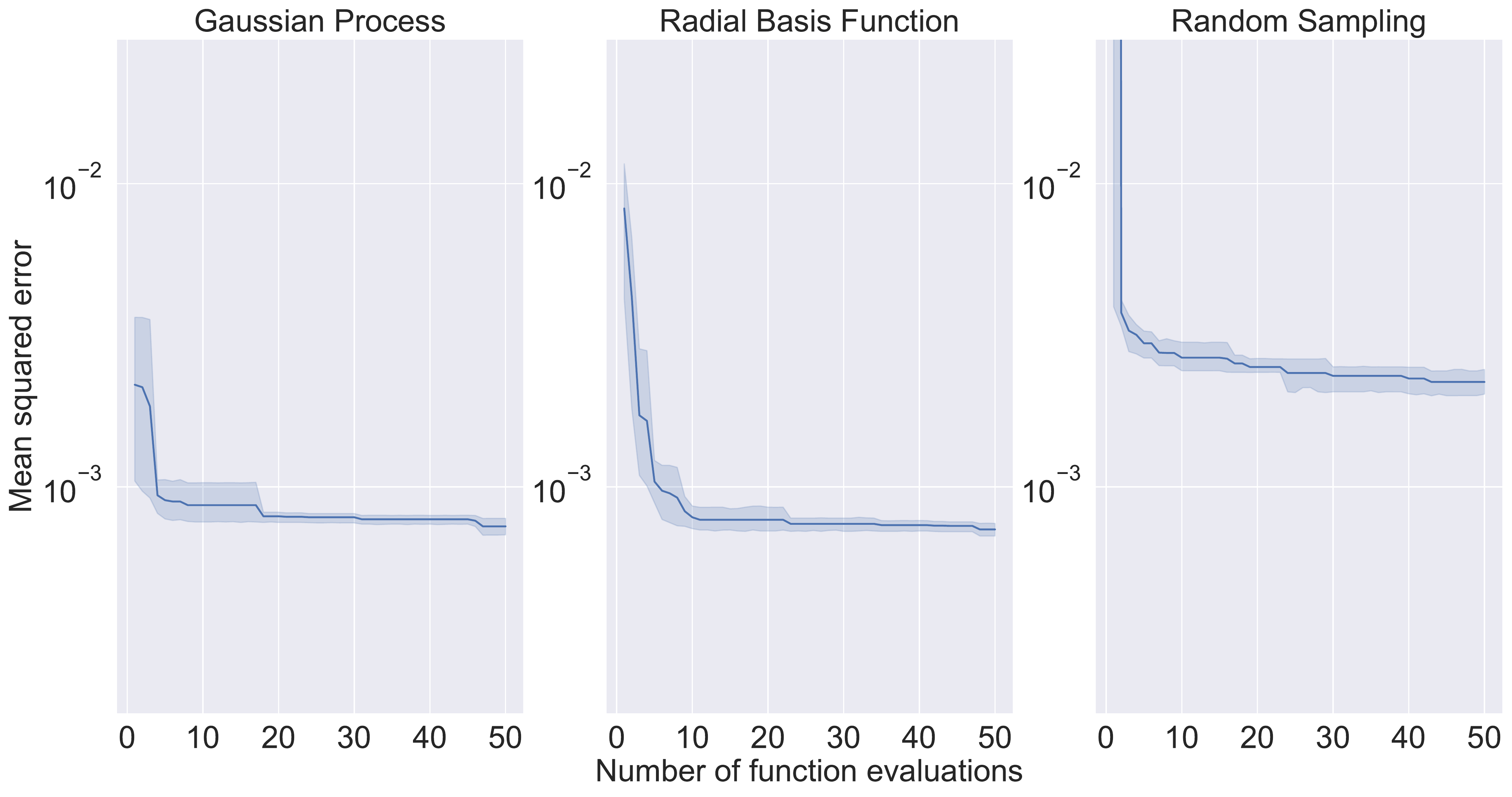}
        \caption{LSTM}
        \label{fig:single_well_error_LSTM}
    \end{subfigure}
    ~ 
    \begin{subfigure}[b]{0.48\textwidth}
        \includegraphics[width=\textwidth]{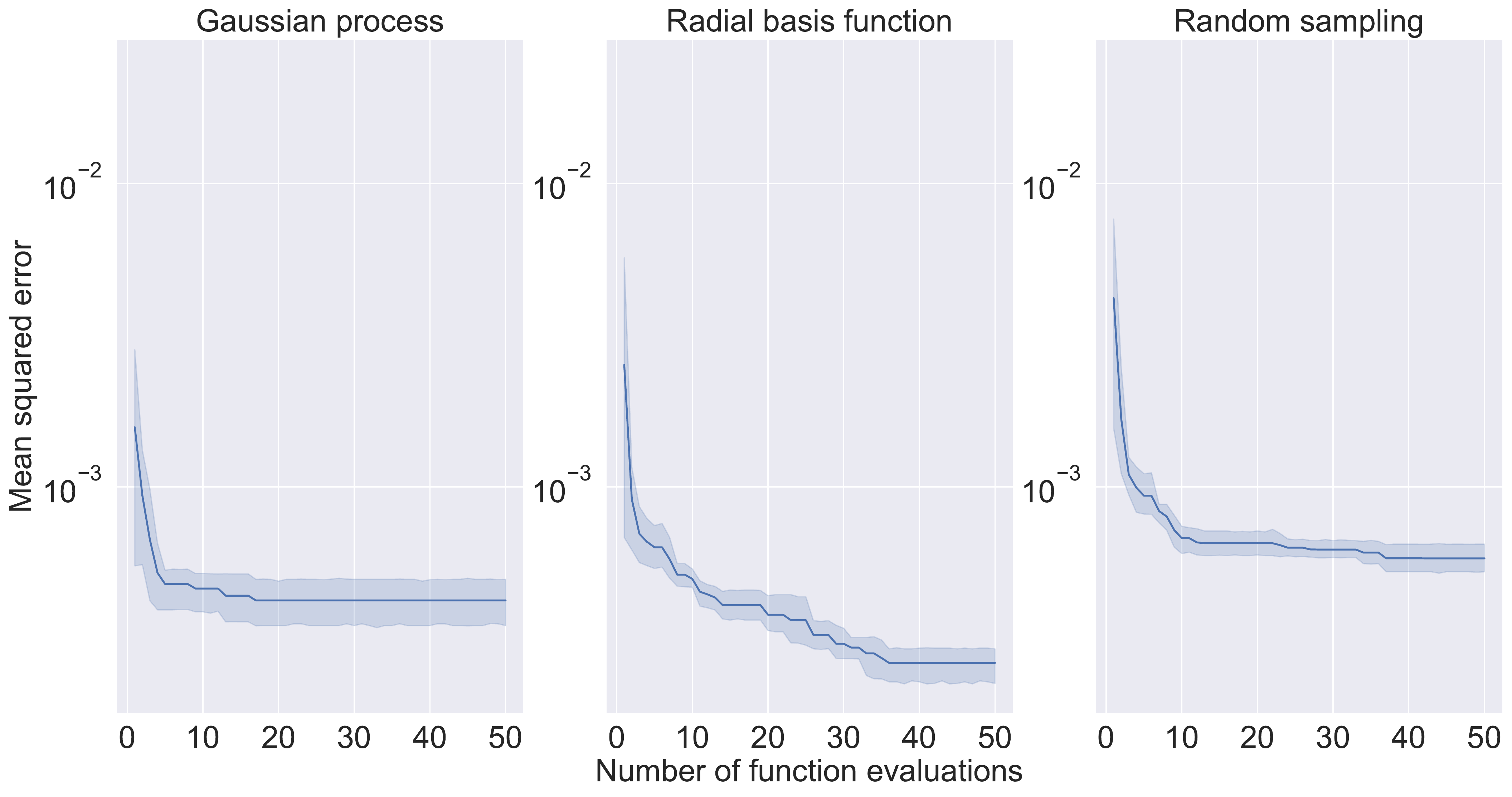}
        \caption{MLP}
        \label{fig:single_well_error_MLP}
    \end{subfigure}
    ~ 
    \begin{subfigure}[b]{0.48\textwidth}
        \includegraphics[width=\textwidth]{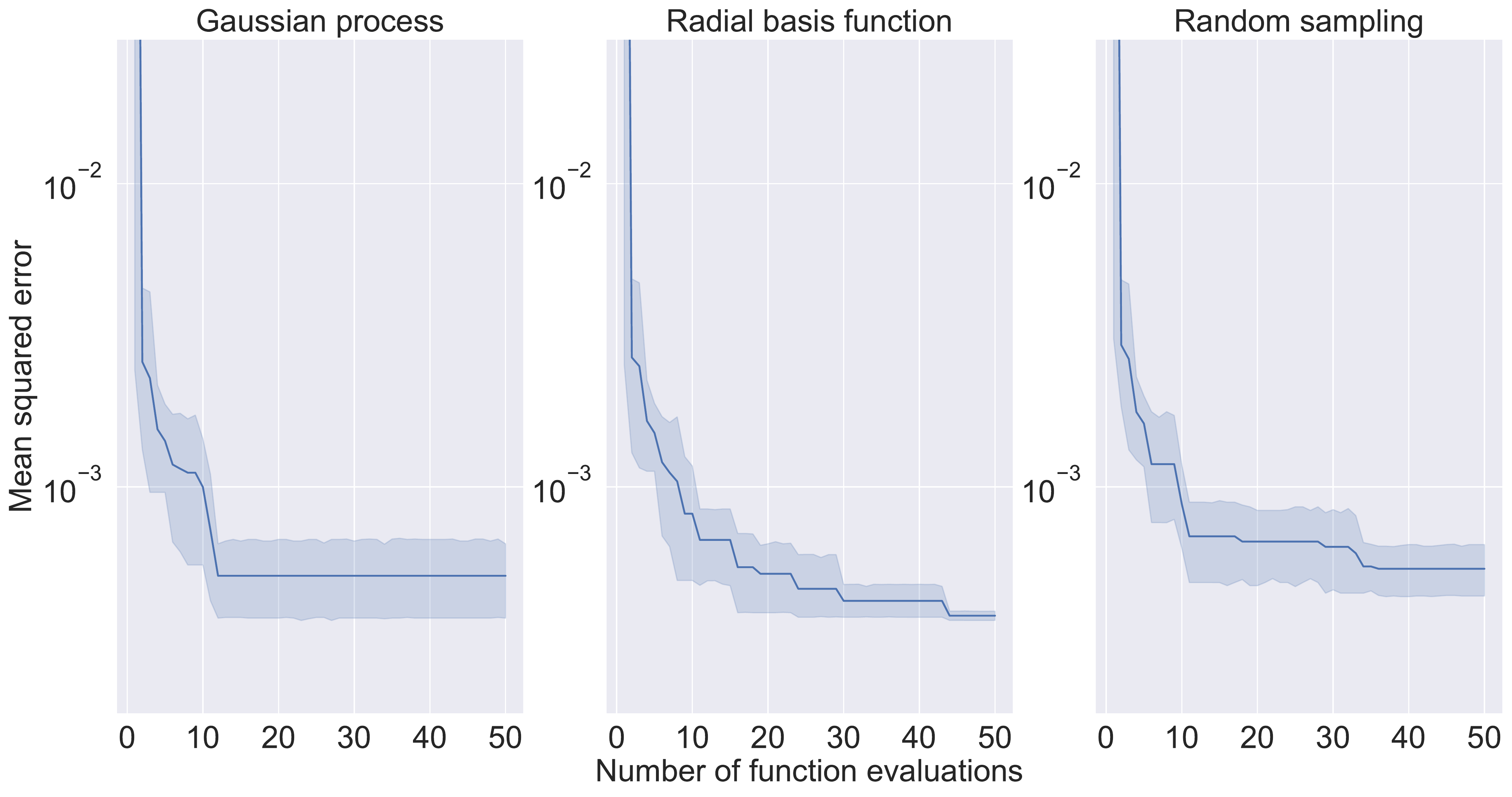}
        \caption{CNN}
        \label{fig:single_well_error_CNN}
    \end{subfigure}
  ~  
    \begin{subfigure}[b]{0.48\textwidth}
        \includegraphics[width=\textwidth]{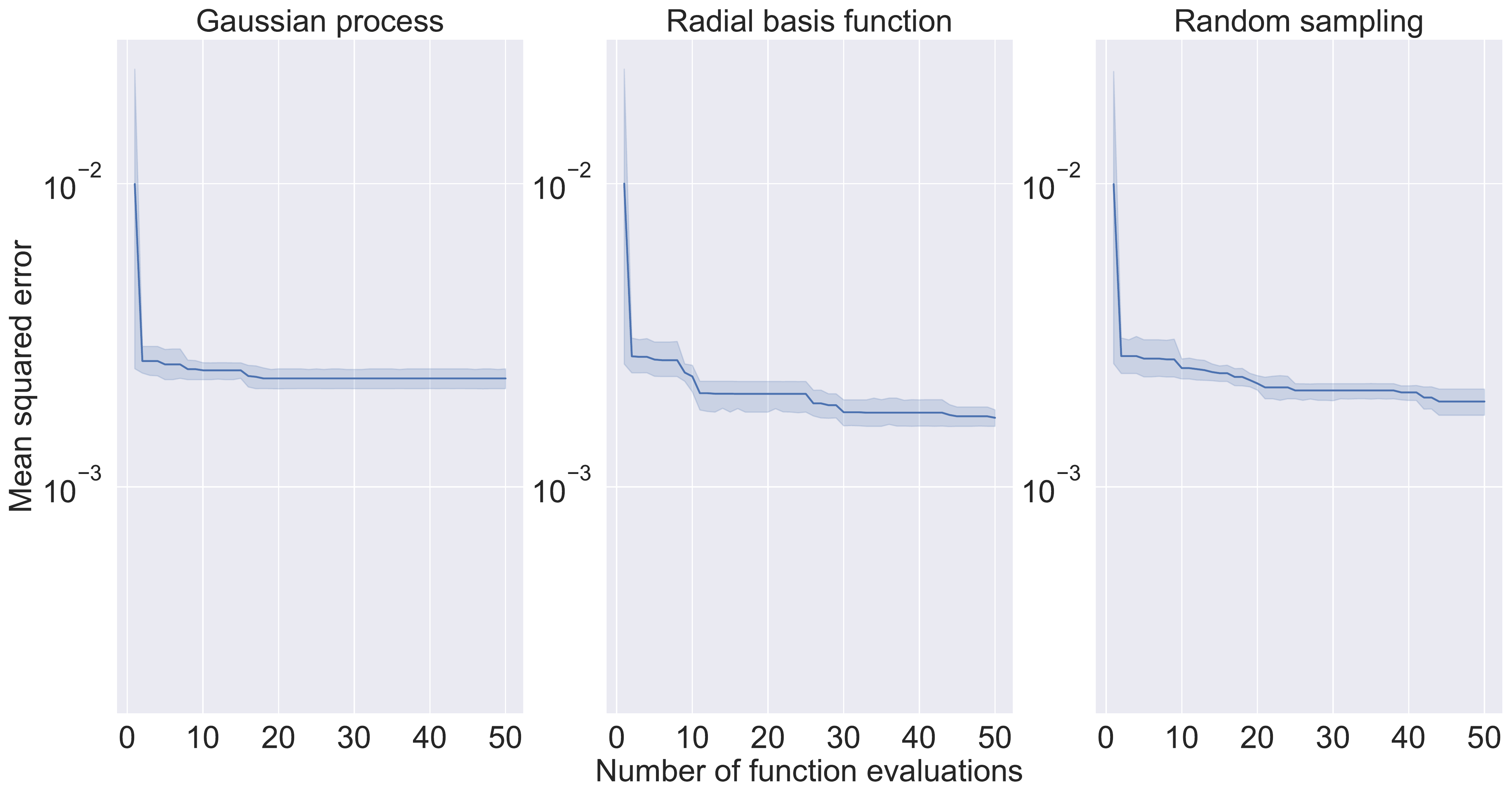}
        \caption{RNN}
        \label{fig:single_well_error_RNN}
    \end{subfigure}
    \caption{Convergence plots showing the average and the standard deviation of the  outer objective function values over five trials versus the number of function evaluations for all four  learning models for the single well case.}\label{fig:conv_singlewell}
\end{figure}

\paragraph{Analysis of HPO Convergence}
Besides the final  solutions obtained with the different HPO methods, we are also interested in how fast the methods converge, i.e., how many calls to the inner optimization loop are necessary. For this purpose, we use progress plots that illustrate how quickly the average and the standard deviation of the outer objective function values over five trials decrease as the number of function evaluations increases. Ideally, the graphs should drop off fast, which means that improved network architectures were  found within very few hyperparameter evaluations. The plot of the standard deviation  reflects how robust each method is. Small  standard deviation envelopes indicate that all five trials with the HPO methods converged to an approximately equally good solution. We show these convergence plots in Figure~\ref{fig:conv_singlewell}. 

For the LSTM (Figures~\ref{fig:single_well_error_LSTM}) we can see that  the GP and RBF approaches lead to more robust solutions (narrow standard deviation bands) than the RS approach. The standard deviation associated with the random sampling remains the largest throughout. This can be expected as the random sampling does not have a systematic  search approach that allows it to intensify the  search locally around good solutions.  

For MLP, on the other hand, the RBF performed best and the convergence plot (Figure~\ref{fig:single_well_error_MLP}) reflects that the RBF approach also converged faster  than the GP and the RS. For MLP,  GP has wider standard deviation bands than RBF throughout indicating less robustness. The standard deviations obtained with RS are smaller for MLP than for LSTM. Since the RS is lacking any systematic search for improvements, this may be an indicator that there are several solutions for MLP with  similar  performance.

For CNN (Figure~\ref{fig:single_well_error_CNN}), we can see significant differences in the convergence behavior between the three HPO methods. The RBF approach has smaller standard deviations  than GP and RS. We notice that the search with the GP gets stuck after about 12 function evaluations and it gets stuck in different solutions as is reflected by the width of the standard deviation envelopes.

Finally, for RNN (Figures~\ref{fig:single_well_error_RNN}), we observe the overall worst performance. All HPO models make only slow progress towards improved solutions and the five trials with the RBF method show the  average best performance.

\begin{figure}[htbp]
    \centering
    \begin{subfigure}[b]{0.48\textwidth}
        \includegraphics[width=\textwidth]{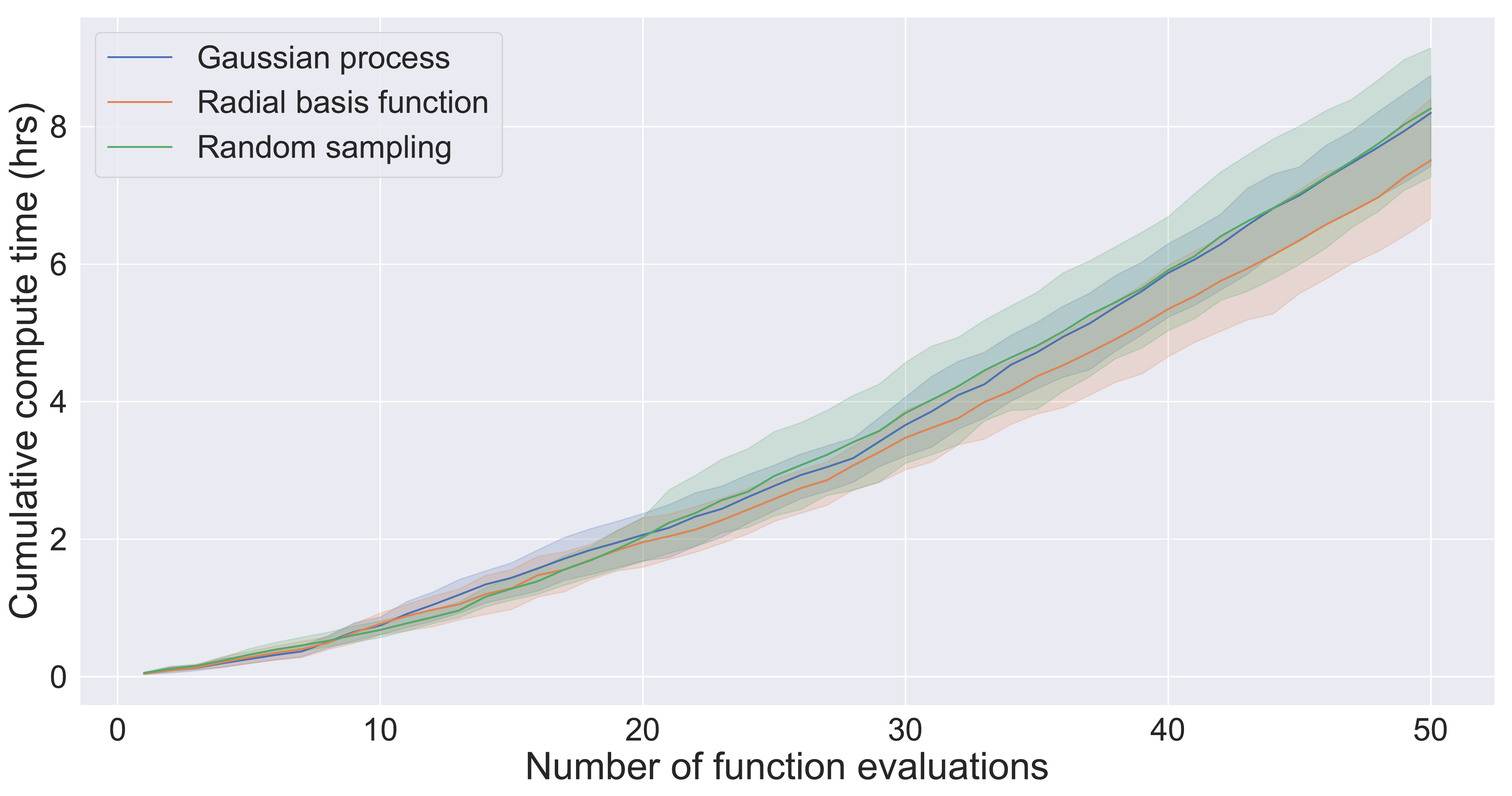}
        \caption{LSTM}
        \label{fig:single_well_compute_time_LSTM}
    \end{subfigure}
    ~ 
    \begin{subfigure}[b]{0.48\textwidth}
        \includegraphics[width=\textwidth]{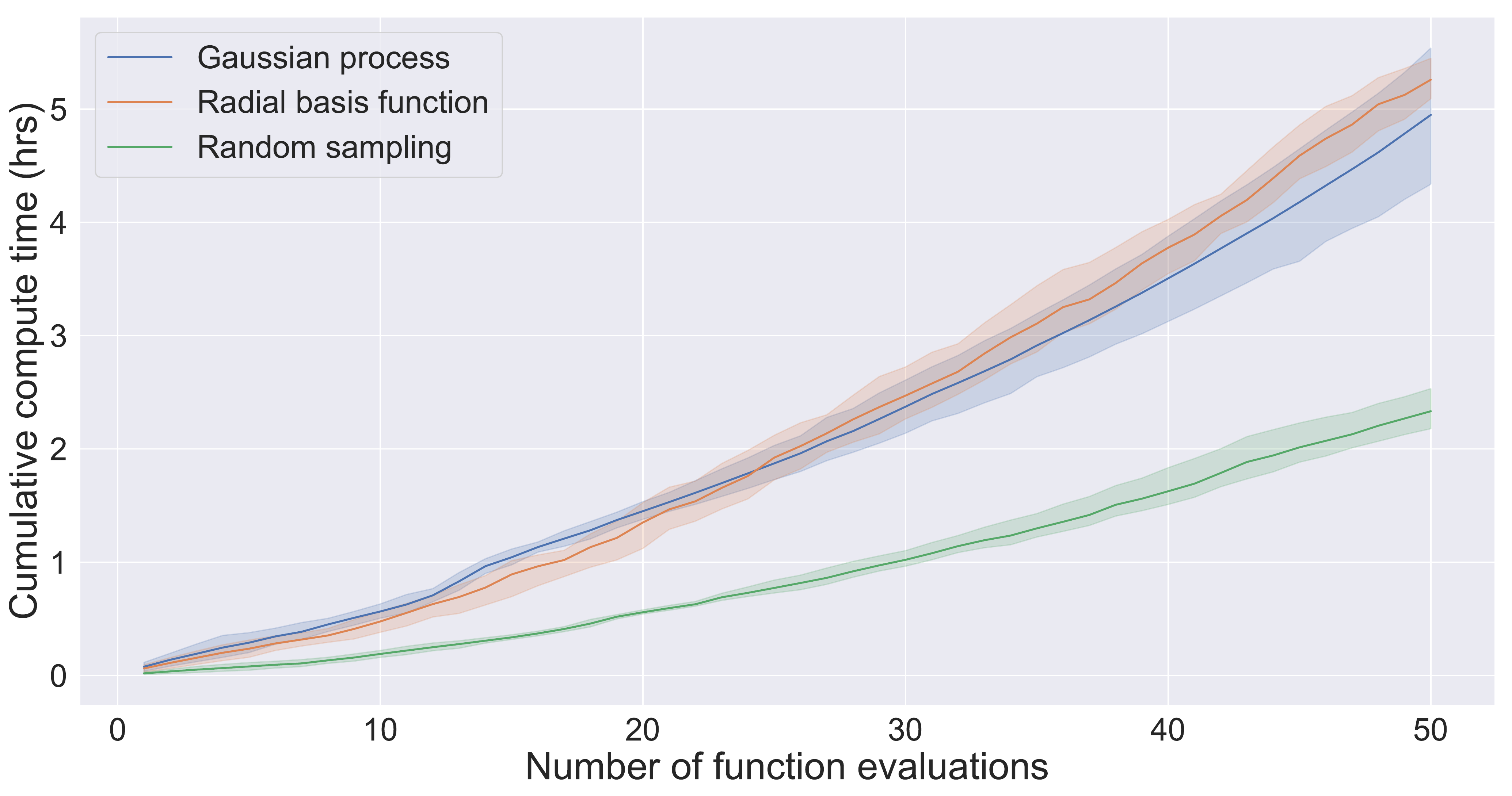}
        \caption{MLP}
        \label{fig:single_well_compute_time_MLP}
    \end{subfigure}
    ~ 
    \begin{subfigure}[b]{0.48\textwidth}
        \includegraphics[width=\textwidth]{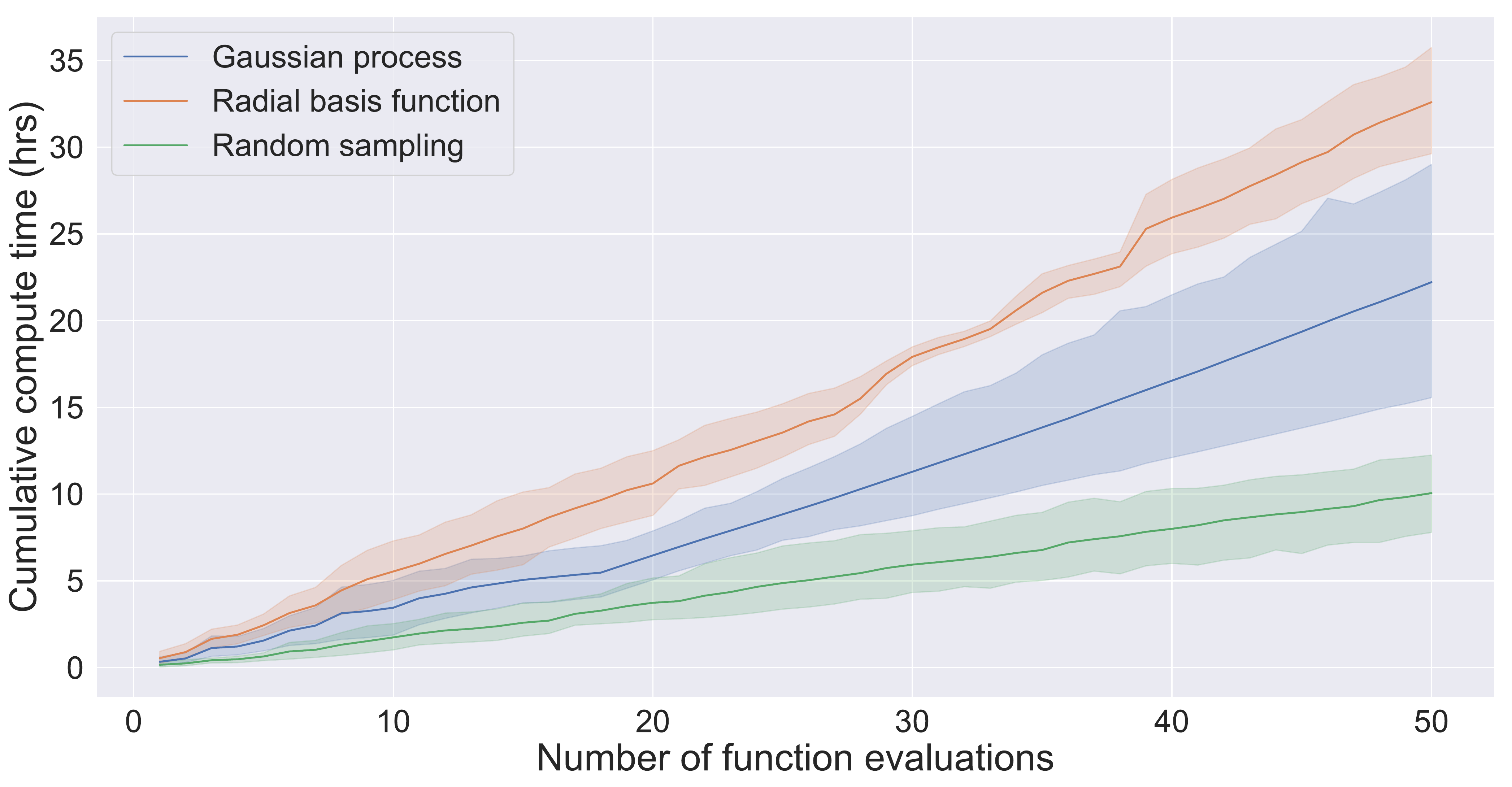}
        \caption{CNN}
        \label{fig:single_well_compute_time_CNN}
    \end{subfigure}
  ~  
    \begin{subfigure}[b]{0.48\textwidth}
        \includegraphics[width=\textwidth]{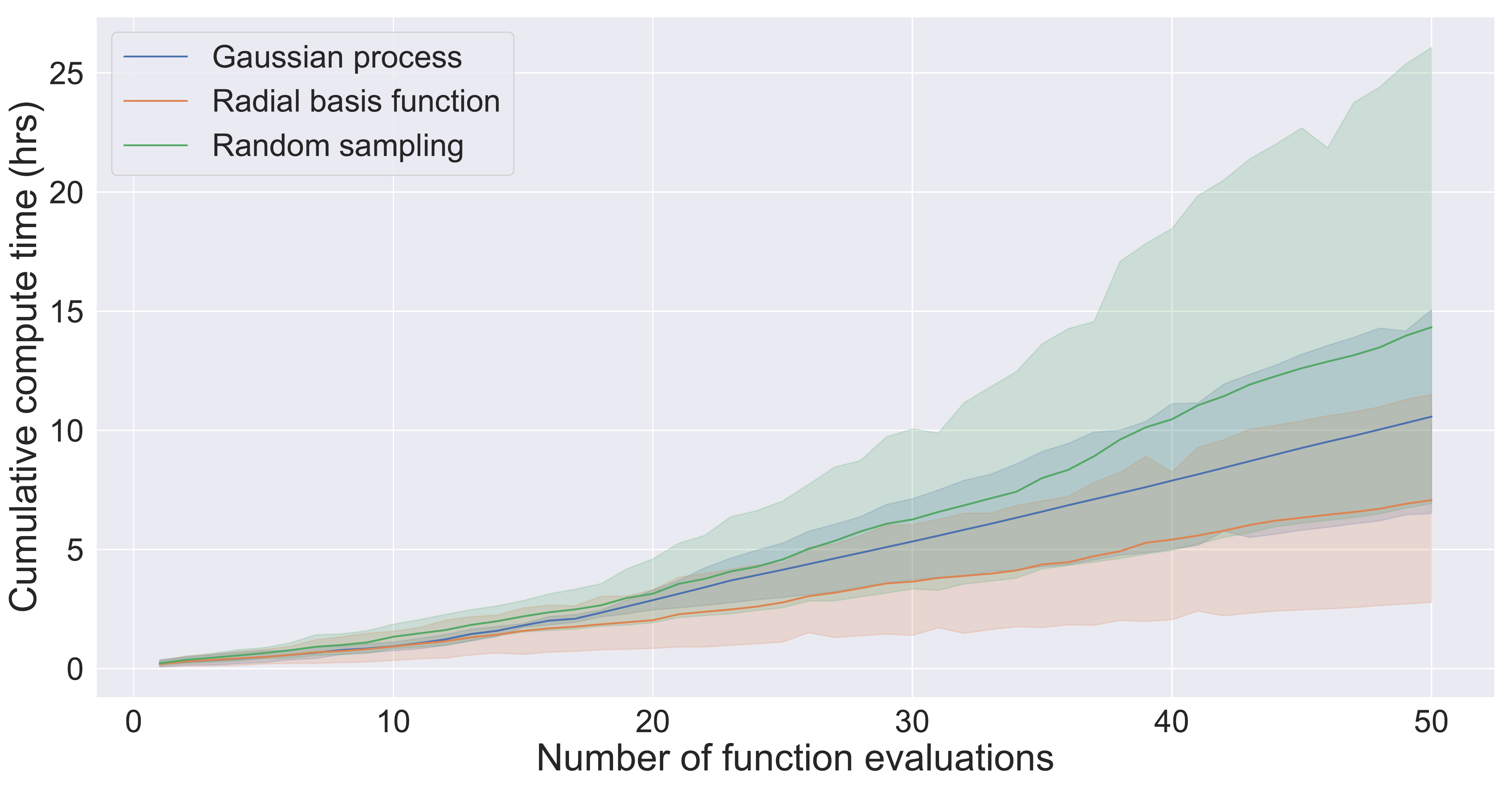}
        \caption{RNN}
        \label{fig:single_well_compute_time_RNN}
    \end{subfigure}
    \caption{Cumulative function evaluation time versus   number of function evaluations for  all four learning models  for the single well case.}
    \label{fig:comp_singlewell}
\end{figure}

\paragraph{Analysis of Computing Times}
As training larger, more complex networks requires more compute time, we also plot the cumulative function evaluation time for each network type and each HPO method in order to compare the wall clock time to solution. Here, we only compare the amount of time spent on  function evaluations and we assume that the HPO methods'  own computational overheads are negligible. We show the average and the standard deviations of the  computing times over all five trials  in Figure~\ref{fig:comp_singlewell}. Comparing the figures, we see that optimizing the hyperparameters of the MLP (Figure~\ref{fig:single_well_compute_time_MLP}) is  fastest for all HPO methods (within 5 hours) even though the optimal MLP architectures were fairly complex. 

For the LSTM (Figure~\ref{fig:single_well_compute_time_LSTM}), we can see that the computing times of all three HPO methods are approximately equal. Since the final optimized networks are smaller compared to MLP  for all HPO methods and the number of epochs is approximately the same for RS and GP (compare Figure~3a in Section C.1.1 in the online supplement), we conclude that on average all HPO methods tried networks of similar complexity. The computing times for the CNN (Figure~\ref{fig:single_well_compute_time_CNN}) are significantly larger than for all other network types (RBF needs on average over 30 hours for the  CNN  compared to less than 8 hours for all other network types). For RNN (Figure~\ref{fig:single_well_compute_time_RNN}), we observe large variations between the times for the individual optimization trials. RBF may take from 3 to 11 hours and RS  takes between 7 hours and 25 hours. Thus, if the goal is to train learning models  on a laptop or simple desktop computer to make predictions for future groundwater levels   in a timely manner, one may prefer using the MLP or the LSTM in terms of expected computing time.

\begin{figure}[htbp]
    \centering
    \begin{subfigure}[b]{0.49\textwidth}
        \includegraphics[width=\textwidth]{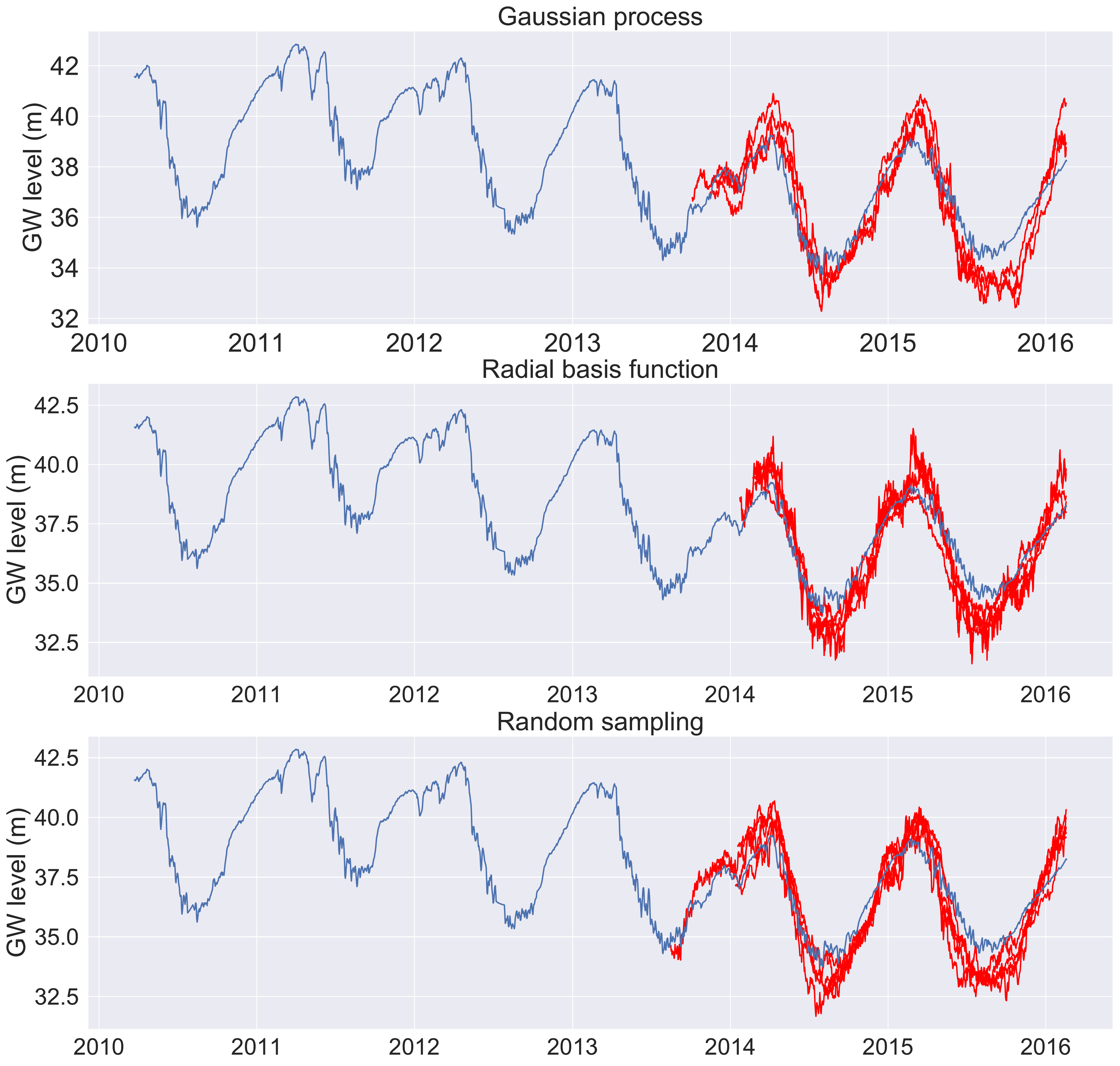}
        \caption{MLP}
        \label{fig:single_well_MLP_GW_pred}
    \end{subfigure}
  ~  
    \begin{subfigure}[b]{0.48\textwidth}
        \includegraphics[width=\textwidth]{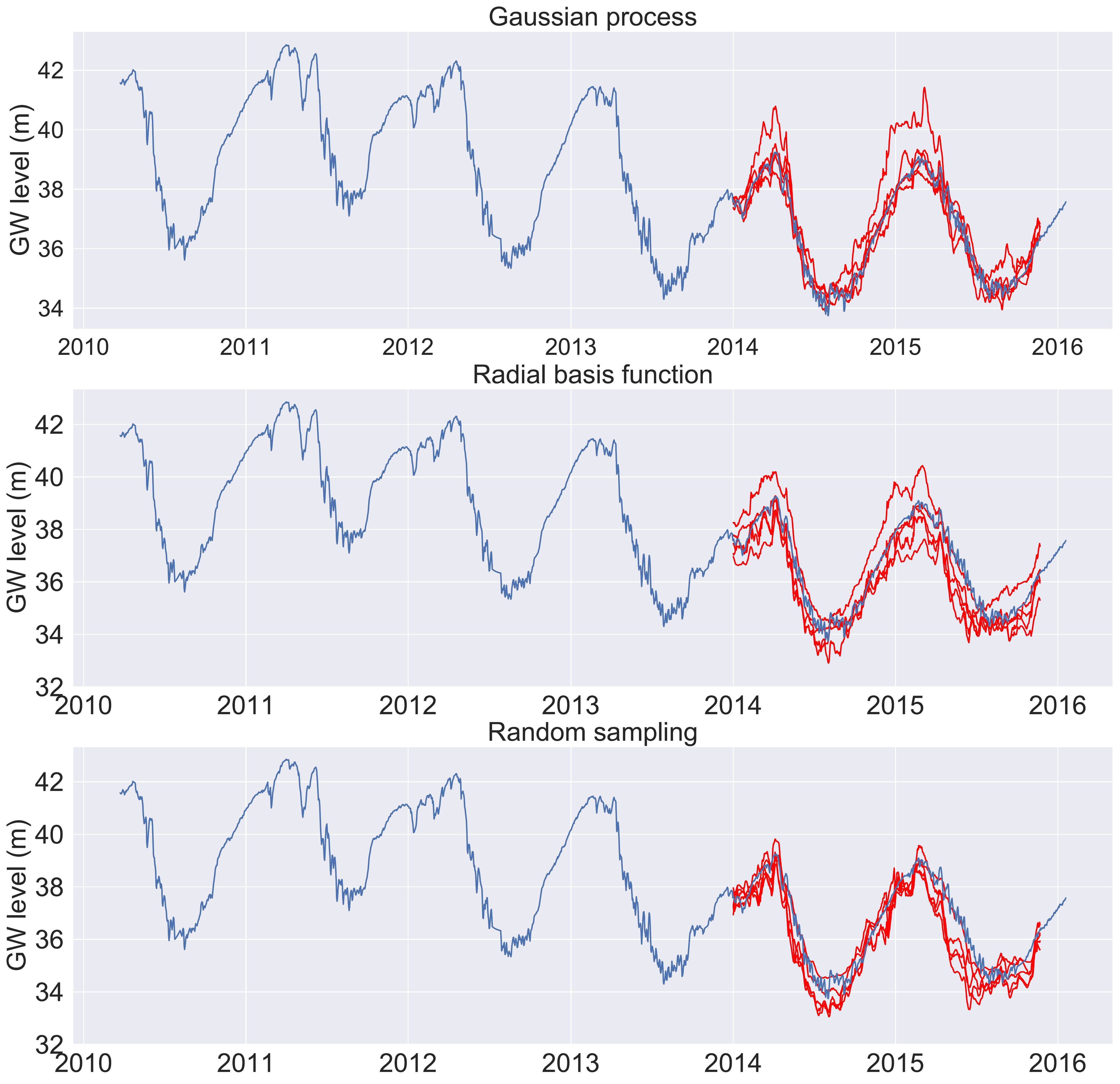}
        \caption{LSTM}
        \label{fig:single_well_LSTM_GW_pred}
    \end{subfigure}
    \caption{Groundwater level predictions obtained from five hyperparameter optimizations of MLP and LSTM, respectively,  for the single well case for each HPO method. GW=groundwater.}
    \label{fig:pred_singlewell}
\end{figure}

\paragraph{Analysis of Performance on Testing Data}
We illustrate in Figure~\ref{fig:pred_singlewell} how well the optimized MLP and LSTM models capture the testing time series data (for CNN and RNN results, see the online supplement, Section~C.1.2,  Figure~4). As described earlier, we train the models on a subset of the historical observations (approximately beginning 2010-end 2013) and we test their performance on another subset (approximately beginning 2014-end 2015) to compute the outer objective function values. In Figures~\ref{fig:single_well_MLP_GW_pred} and \ref{fig:single_well_LSTM_GW_pred}, we show the training data  and for each optimization trial,  we show the predictions  of the optimized models for the testing data. The closer the predictions (red graphs) are to the true data (blue), the better. We can see that both MLP and LSTM are able to capture the seasonality in the data well.   The predictions made with the MLP appear more erratic, i.e, we see many small jumps in daily groundwater predictions. This behavior is observed in the daily groundwater measurements and is due to, e.g., irrigation or other water withdrawal  activities. The LSTM predictions are somewhat smoother, which indicates that the daily drawdown patterns are not learned as well. 

\begin{figure}[htbp]
    \centering
    \begin{subfigure}[b]{0.48\textwidth}
        \includegraphics[width=\textwidth]{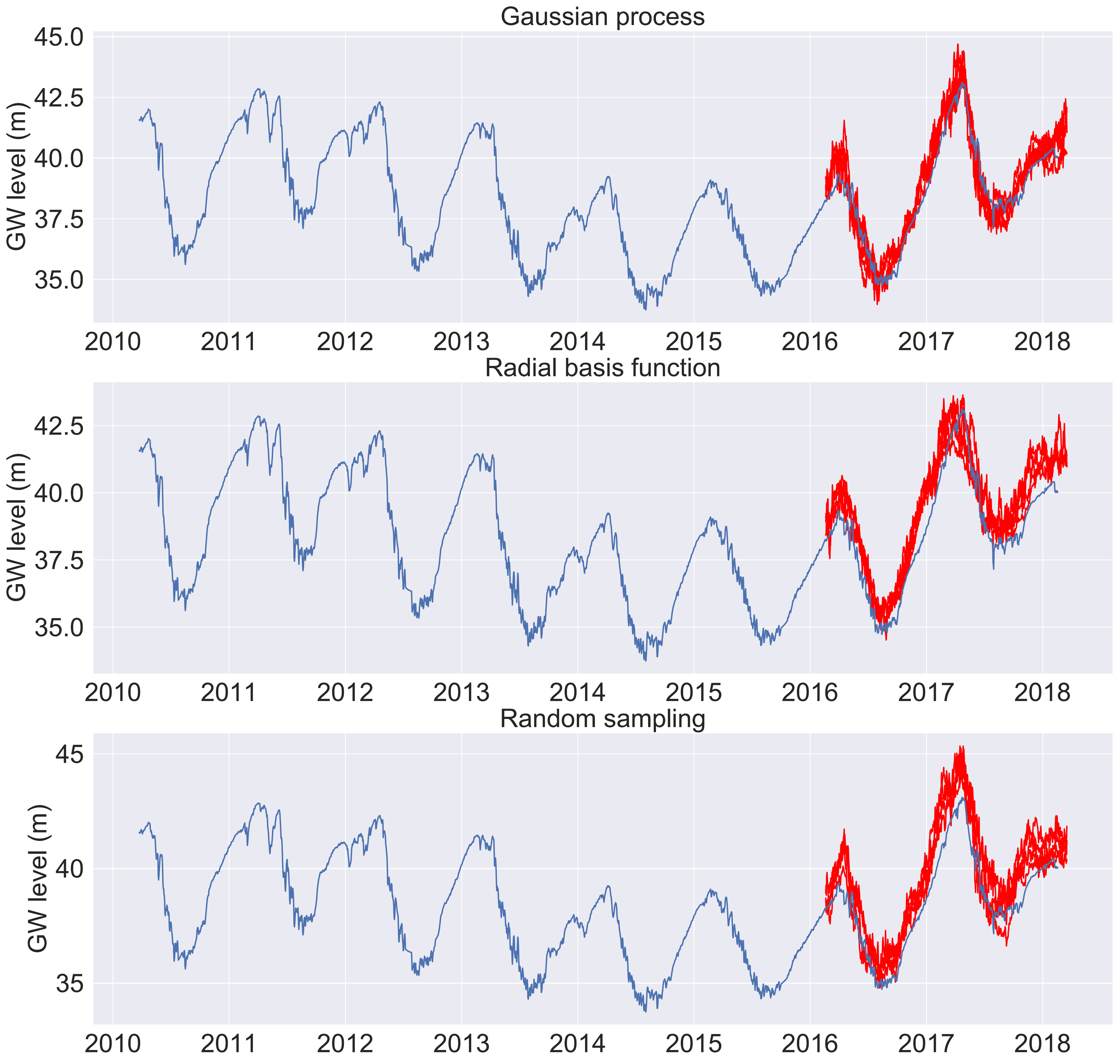}
        \caption{MLP}
        \label{fig:single_well_MLP_GW_pred2}
    \end{subfigure}
  ~  
    \begin{subfigure}[b]{0.48\textwidth}
        \includegraphics[width=\textwidth]{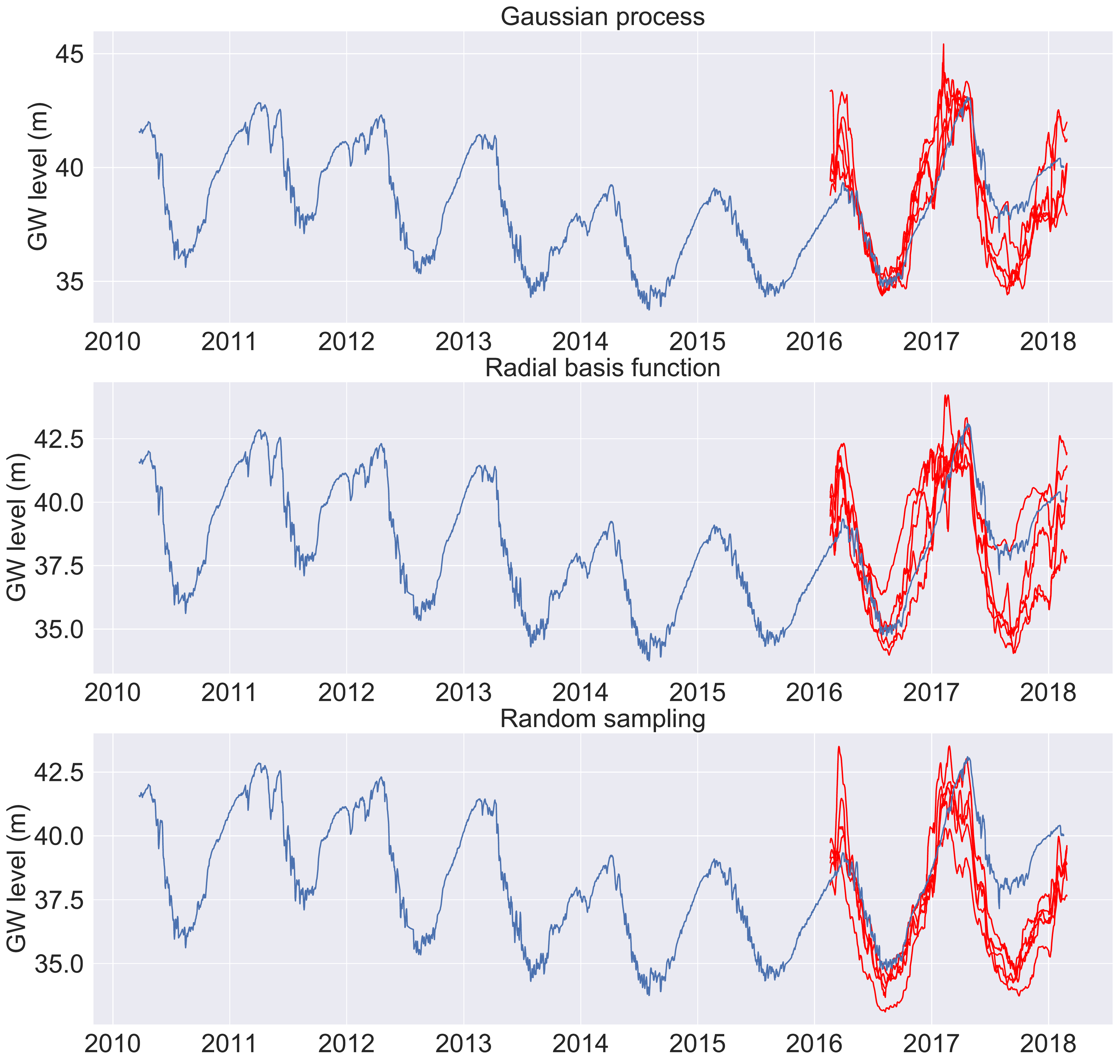}
        \caption{LSTM}
        \label{fig:single_well_LSTM_GW_pred2}
    \end{subfigure}
    \caption{Out-of-sample groundwater predictions with the five optimized MLP and LSTM models, respectively,   for the single well case. GW=groundwater.}
    \label{fig:outpred_singlewell}
\end{figure}

\paragraph{Out-of-Sample Future Groundwater Predictions}
Finally, we are interested in how well the optimized learning models will predict  groundwater data that was not used during the HPO. 
 We retrain the models with the optimized hyperparameters on the whole dataset $\mathcal{D}$ (2010-2016) and we  predict the groundwater levels from February~20, 2016 to February 28, 2018. 
We illustrate these ``out-of-sample'' predictions for each of the five trials for MLP and LSTM in Figure~\ref{fig:outpred_singlewell}.
We  see that both models  are able to keep the trend in the data (decreasing groundwater levels in the summer, increasing levels in the winter), but the  MLP predictions are much more accurate and  less variable  than the LSTM predictions.  The LSTM is not able to capture the higher groundwater levels in the wet years (2017, 2018).  The out-of-sample predictions for CNN and RNN are provided in the online supplement in Section C.1.3 in Figure~5. The RNN makes significantly  worse out-of-sample predictions than the MLP.  The distributions of the MSEs for the out-of-sample predictions are provided in the form of boxplots in the online supplement in Figure~6.                                                                                                                                                                                                                                                                                                                                                                                                                                                                                                                                                                

\subsection{Results for Predicting Groundwater at  Multiple Wells in  Butte County, CA}
Watersheds are  large and generally have more than a single well from which groundwater is pumped or with the help of which the amount of available groundwater is monitored. Using multiple wells  will allow us to better assess the average health of a watershed, and thus to avoid letting local fluctuations drive policies and decisions. Therefore, within Butte County, we selected three additional wells. Ideally, we do not want to train and optimize a separate learning model for each  well, but rather we want  one learning model that can be trained on and used for predicting the data at multiple wells simultaneously.    We train and optimize the hyperparameters as before with the only difference that we now have three groundwater input data streams  and three nodes in each output layer. We use these data to optimize the hyperparameters of MLP, LSTM, and CNN. 
 For reasons of space considerations, we present detailed results only for the MLP. The LSTM and CNN results are in the online supplement in Section C.2.

\begin{figure}[htbp]
    \centering
    \includegraphics[scale=.18]{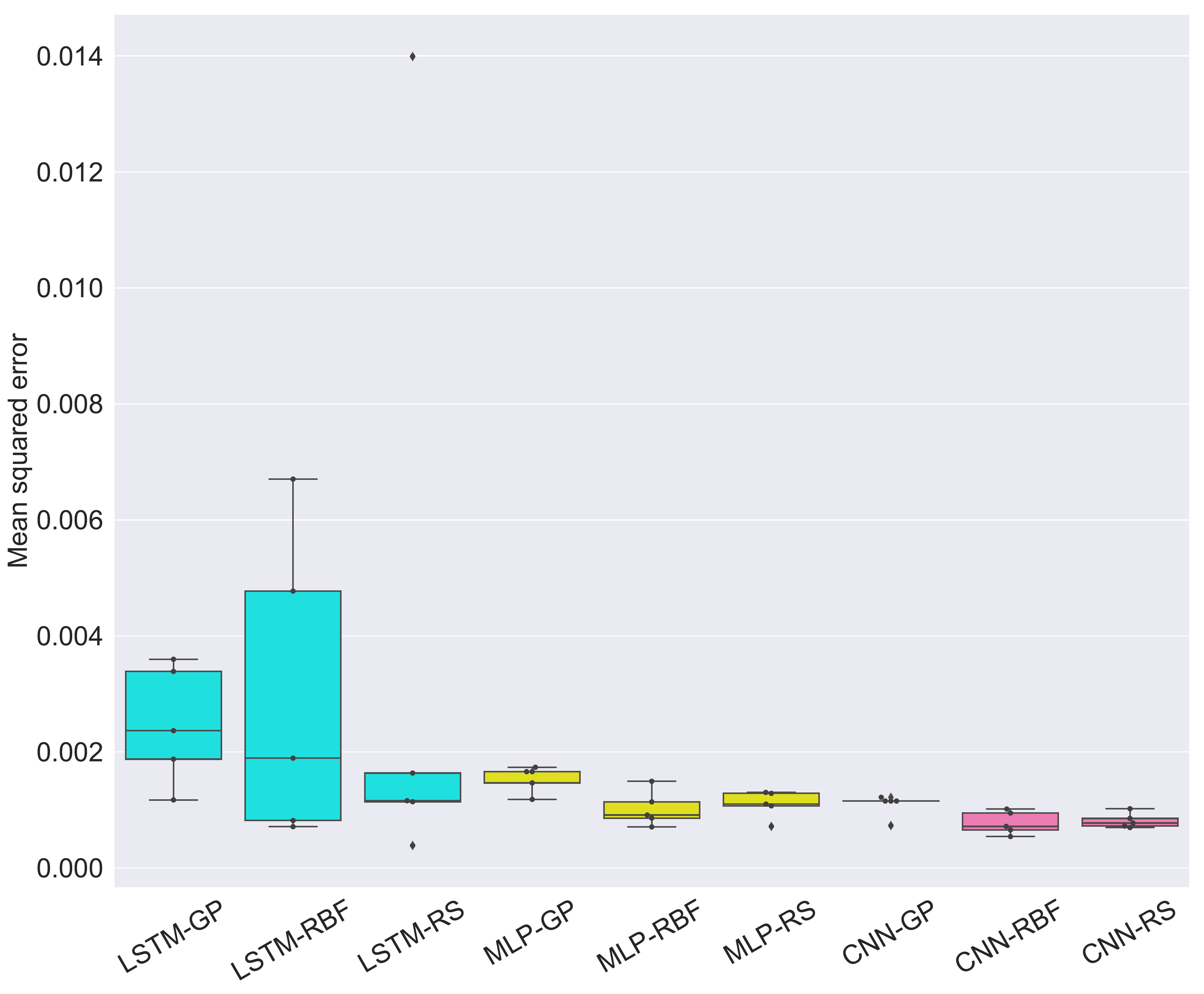}
    \caption{Boxplots of mean squared errors for the multi-well case. LSTM=long short-term memory network; MLP=multilayer perceptron; CNN=convolutional neural network; RNN=recurrent neural network; GP=Gaussian process; RBF=radial basis function; RS = random sampling.}
    \label{fig:multi_well_error_plot}
\end{figure}

\paragraph{Comparison of Mean Squared Errors}
We show a comparison of the distributions of the final mean squared errors for all three learning models and all HPO methods obtained from five optimization trials in Figure~\ref{fig:multi_well_error_plot}. The optimal mean squared errors are for almost all models and HPO methods  larger by an order of magnitude for the multi-well case than for the single well case (see also the numerical data provided in Table 3 in the online supplement). This is to be expected  as more time series must be fit and one set of hyperparameters may not be optimal for the data of all three wells. 
 We can see that the MLP and the CNN optimized with the RBF approach lead to  the lowest errors. Also RS performs reasonably well for all three models in the multi-well case, while it showed significantly worse performance in the single well case. Thus, in terms of performance consistency, one should opt for an HPO method that systematically explores the search space rather than the RS method. 

\begin{figure}[htbp]
    \centering
    \includegraphics[scale=.2]{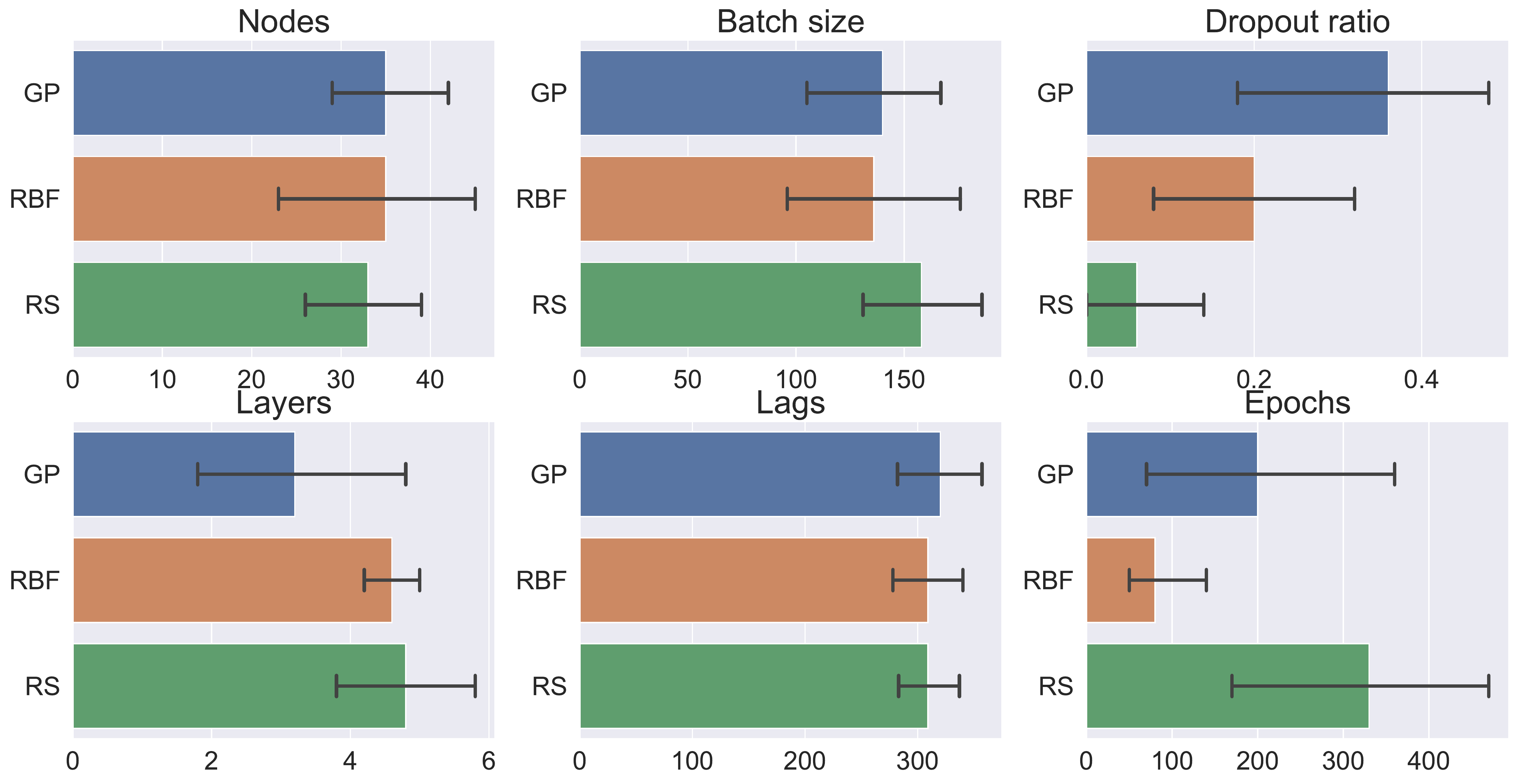}
    \caption{Distribution of the optimized hyperparameter values for MLP for the multi-well case. The colored bars show the mean of the values over five trials and the black lines indicate the standard deviation.  GP=Gaussian process; RBF=radial basis function; RS=random sampling.}
    \label{fig:Multi_well_MLP_hyperparameter}
\end{figure}

\paragraph{Analysis of Optimized Hyperparameters}
We show the distribution of the optimized hyperparameters only for the MLP (Figure~\ref{fig:Multi_well_MLP_hyperparameter}). The results for LSTM and CNN are in the online supplement in Section~C.2.1 in Figure~7. For the MLP, the RBF method chooses again a large complex network (4-5 layers and on average over 30 nodes per layer) and a large lag.   Also the optimal solutions for the CNN (see online supplement)  show that a complex network with many layers and filters is preferred and also the lag  is large. On the other hand, the optimal  LSTM network structure is significantly smaller with only 1-2 layers.

\begin{figure}[htbp]
    \centering
    \begin{subfigure}[b]{0.48\textwidth}
        \includegraphics[width=\textwidth]{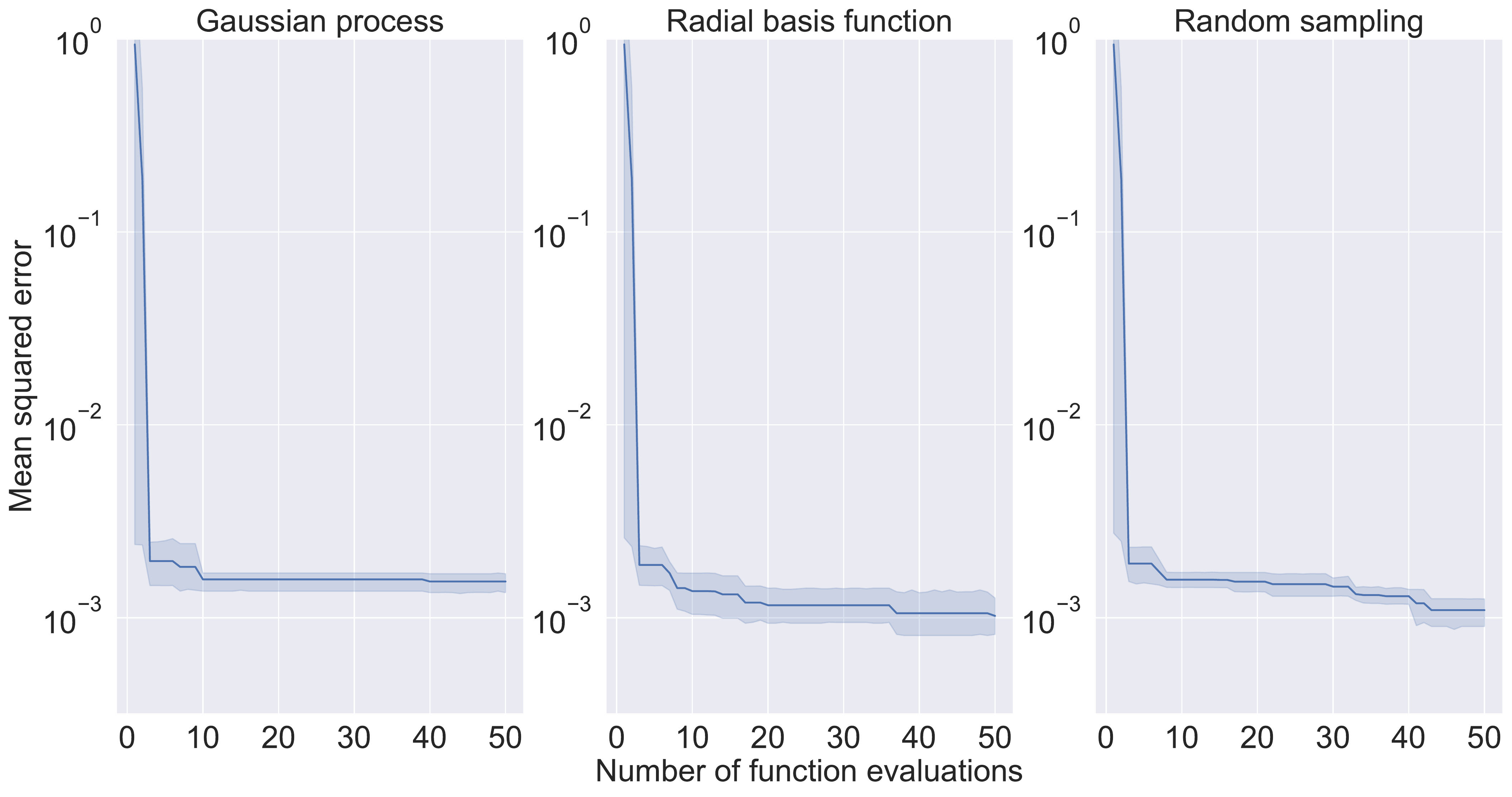}
        \caption{Convergence plot}
        \label{fig:Multi_well_error_MLP}
    \end{subfigure}
  ~  
    \begin{subfigure}[b]{0.48\textwidth}
        \includegraphics[width=\textwidth]{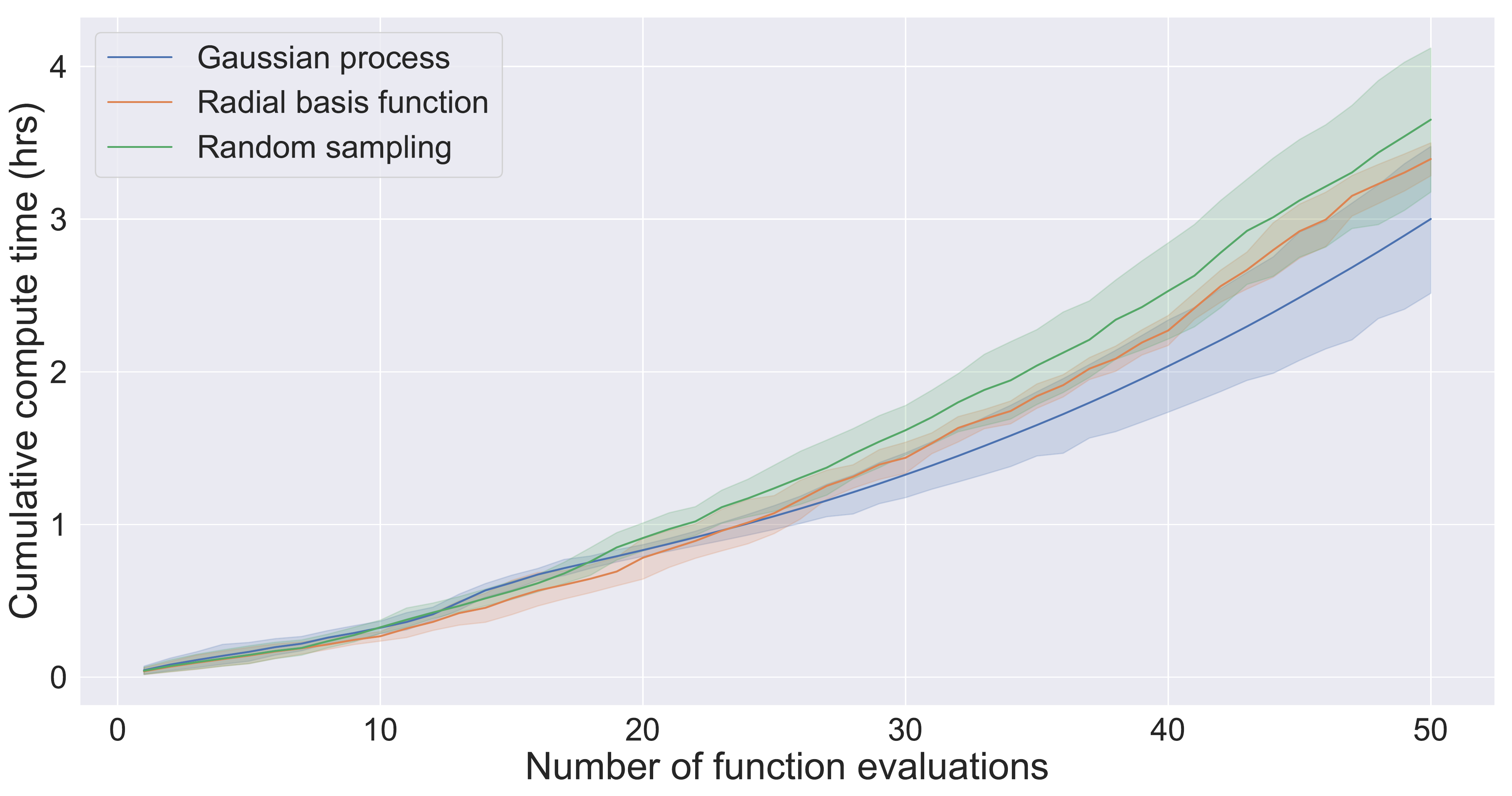}
        \caption{Cumulative evaluation time}
        \label{fig:Multi_well_compute_time_MLP}
    \end{subfigure}
    \caption{Convergence plot showing the average and the standard deviation of the outer objective function value  versus the number of function evaluations  (left) and  cumulative function evaluation time versus   number  of function evaluations (right)       for MLP  for the multiple well case.}
    \label{fig:MLP_multiwell}
\end{figure}

\paragraph{Analysis of HPO Convergence}
As for the single well case, we illustrate how fast each HPO method converges. We show the progress plots on the example of MLP in Figure~\ref{fig:Multi_well_error_MLP} (results for LSTM and CNN are in the online supplement in Section~C.2.2 in Figure~8). We observe that all HPO methods  converge fast and the RBF method achieves on average the best performance.

\paragraph{Analysis of Computing Times}
The cumulative computation time for the MLP over all function evaluations is shown in Figure~\ref{fig:Multi_well_compute_time_MLP}. All HPO methods require approximately the same time and the HPO is completed within 4 hours. The CNN  again requires the largest amount of time to solution (up to 30 hours) and the HPO for the LSTM completes within 10 hours  (see Figure~10 in Section~C.2.4 in the online supplement). Thus, we conclude that the MLP optimized with the RBF should be the learning model and HPO method of choice as it attains high quality  solutions within the lowest computation time.

\begin{figure}[htbp]
    \centering
    \includegraphics[scale=.25]{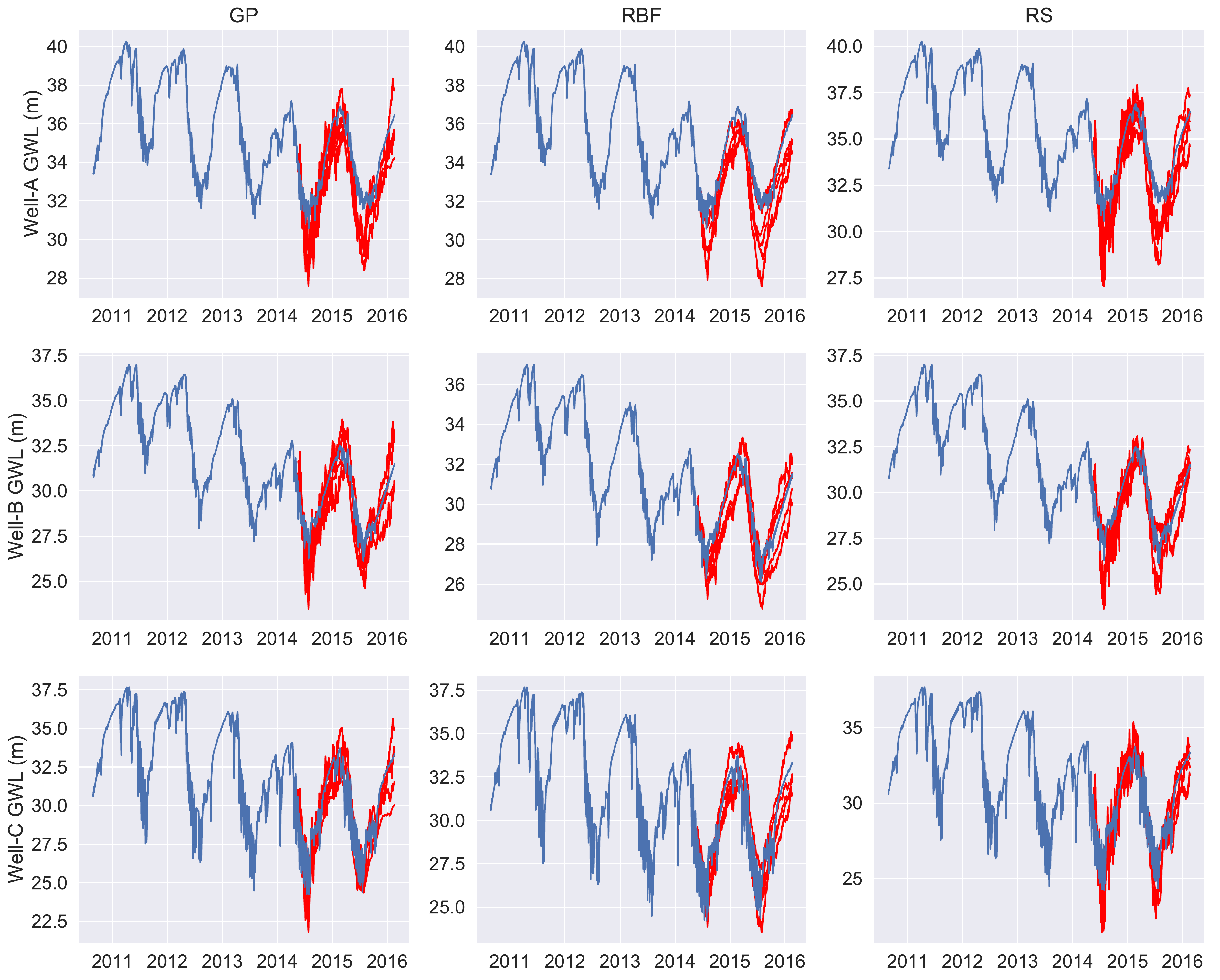}
    \caption{Groundwater level predictions obtained from five hyperparameter optimizations of MLP for the multiple well case for each HPO method. GWL=groundwater level.}
    \label{fig:Multi_well_MLP_GW_pred}
\end{figure}

\paragraph{Analysis of Performance on Testing Data}
In Figure~\ref{fig:Multi_well_MLP_GW_pred}, we show the predictions obtained with the optimized MLP models for all three wells for the testing data. All three HPO methods are able to capture the seasonality in the data and all three HPO methods appear to underpredict the true groundwater levels for wells A and B. Well C is generally better fitted by all three HPO methods.  This same observation holds true for the  CNN predictions (Figure~9b in Section~C.2.3 in the online supplement). One possible explanation is that  the errors made in the predictions for well C   dominate the HPO. This may be due to the more erratic behavior of the groundwater levels for well C, which are manifested by more severe drawdowns and thus more patterns must be learnt (the most complex time series dominated the optimization). Using  a weighted sum of the wells' errors as outer objective function could potentially help mitigating this problem. The LSTM predictions (Figure~9a in Section~C.2.3   in the online supplement) have generally a larger variability than CNN and MLP and we observe larger jumps in the daily groundwater predictions. Note that for the multi-well case, we do not have additional observations for all three wells for making ``out-of-sample'' predictions. 

\section{Summary and Directions of Future Research}\label{sec:concl}
Our work presented in this article is motivated by the need for computationally lightweight approximation models that can run on a laptop or simple desktop computer and that can reasonably accurately predict groundwater levels with the overarching goal to enable timely and reliable water resources management decisions.

In this article, we studied the applicability of several purely data-informed learning models whose hyperparameters were optimized with three different methods for predicting the groundwater level. 
We used four deep learning models, namely convolutional neural networks (CNNs), recurrent neural networks (RNNs), multilayer perceptron (MLP), and long short-term memory networks (LSTMs). We formulated a bilevel optimization problem for finding the optimal hyperparameters of the learning models. At the upper level, we have a pure-integer stochastic problem, and at the lower level, we have a large-scale global optimization problem. We solved the hyperparameter optimization (HPO) problem with three different methods. We used two surrogate model based optimization methods (one  based on radial basis functions (RBFs) and one based on Gaussian process (GP) models) and a random sampling method. In our numerical experiments, we trained the learning models on daily observation data of groundwater levels and meteorological variables at, respectively, one (``single well case'') and three (``multi-well case'') groundwater monitoring wells in Butte County, California, USA. 

Our study shows that with the right HPO method, different types of deep learning models, including CNN,  LSTM, and MLP (and to a limited extend also RNN), can be tuned to make accurate  forecasts of  daily groundwater levels (within two meters of the true values). The main difference between the performance of the methods is the total optimization time and the associated  optimal model complexity. We showed that the surrogate model based HPO methods lead  to better performing  network  architectures than  randomly sampling the  hyperparameter space. Our results show that \textit{the ``simplest'' learning model, the MLP, optimized with the RBF method,  generally performs best and its optimization time is lowest}. The RNN is the worst performing model and the most cumbersome to use as it often fails to fit a model for an architecture due to its problems during backpropagation. The CNN also performs well in terms of prediction accuracy, but it requires considerably more computation time than the other models. The results for the  LSTM are less robust and have a large variation. 

Directions of future research include applying the RBF-optimized MLP model for groundwater predictions at other watersheds in California and other parts of the United States to   assess how well our method generalizes to watersheds with different characteristics (e.g., geology, groundwater use, weather patterns). We will study the sensitivity of the model performance to different input variables. This analysis will inform us about  which data should be collected and thus where the necessary data acquisition infrastructure investments should be made. Similarly, we will study how sensitive the model predictions are to the amount of training data points (daily versus monthly observations) and  when fewer years of data are used in training.  
This future research direction also includes applying our  method for learning model optimization on other  time series data such as predicting water quality.

\paragraph{Acknowledgements}
This work was supported by Laboratory Directed Research and Development (LDRD) funding from Berkeley Lab, provided by the Director, Office of Science, of the U.S. Department of Energy under Contract No. DE-AC02-05CH11231. This research used resources of the National Energy Research Scientific Computing Center (NERSC), a U.S. Department of Energy Office of Science User Facility operated under Contract No. DE-AC02-05CH11231.

\bibliographystyle{unsrt}  
\bibliography{biblio1} 
\end{document}


\maketitle
\noindent
This online supplement contains additional details of the hyperparameter optimization method, data, and results. 

\section{Scoring in the Radial Basis Function  Method}
In Section 3.3  in the main document (\textit{Radial Basis Functions and Stochastic Sampling}) we described how to create a large number of candidate points for the next function evaluation.  We denote the set of candidate points by $\mathbf{x}_1, \ldots, \mathbf{x}_{2M}$ (here $M$ is a parameter which we set to 500). For each candidate point we use the radial basis function (RBF) to predict its objective function value (RBF score): $s_{\text{RBF}}(\mathbf{x}_i)$, $i=1,\ldots,2M$. We also compute the distance of each candidate to the set of already evaluated points (distance score):  $\Delta(\mathbf{x}_i, \boldsymbol\Theta)=\min_{j=1,\ldots,n}\{\| \mathbf{x}_i-\boldsymbol\theta_j \|_2 \}$ for each candidate $i=1,\ldots,2M$.
We scale these scores according to
\begin{equation}
V_\Delta(\mathbf{x}_i)= \frac{\Delta_{\text{max}}-\Delta(\mathbf{x}_i, \boldsymbol{\Theta})}{\Delta_{\text{max}} - \Delta_{\text{min}}} \quad \text{and }\quad 
V_s(\mathbf{x}_i)=\frac{s
_{\text{RBF}}(\mathbf{x}_i)-s_{\text{min}}}{s_{\text{max}}-s_{\text{min}}},
\end{equation}
where $\Delta_{\text{max}} = \max\{\Delta(\mathbf{x}_i,\boldsymbol\Theta), i =1, \ldots, 2M\}$, $\Delta_{\text{min}} = \min\{\Delta(\mathbf{x}_i,\boldsymbol\Theta), i =1, \ldots, 2M\}$, $s_{\text{max}} = \max\{s_{\text{RBF}}(\mathbf{x}_i), i = 1, \ldots, 2M\}$, and $s_{\text{min}} = \min\{s_{\text{RBF}}(\mathbf{x}_i), i = 1, \ldots, 2M\}$. If the maximum and minimum values of the distance score or the RBF score are equal, all candidate points get the same value for the respective score.  
We compute a weighted combination of both scores:
\begin{equation}
V(\mathbf{x}_i) = \omega V_\Delta(\mathbf{x}_i) + (1-\omega)V_s(\mathbf{x}_i),
\end{equation}
where $\omega$ cycles through the values $\{0, 0.25, 0.5, 0.75, 1\}$ as described in the paper. The candidate point with the lowest weighted score is selected  as the new evaluation point. Note that candidates that coincide with previously evaluated points are discarded.

\section{Time Series Plots and Sources of Observation Data}

In this section, we illustrate  the daily observation data time series of the variables we used in our numerical analysis, namely temperature, precipitation, streamflow discharge, and groundwater levels for both the single well and multi-well cases. We obtained the groundwater level measurements  for  particular well sites in Butte County from the California Natural Resources Agency  under their SGMA data viewer portal for 2010-2018. The wells are located within a $5km$ radial distance from each other. We retrieved the  daily averaged streamflow discharge at a nearby river site (Butte Creek)  from the California Data Exchange Center (CDEC) from the station code BCD: Butte Creek Durham. The daily averaged precipitation and temperature were also obtained from the CDEC website at the Chico  weather station. These stations are in close proximity to the groundwater wells and are located within the same Butte Creek basin. Table~\ref{tab:data_source} shows a summary of the data sources we use in our study.

\begin{figure}[htbp]
    \centering
    \includegraphics[scale=.3]{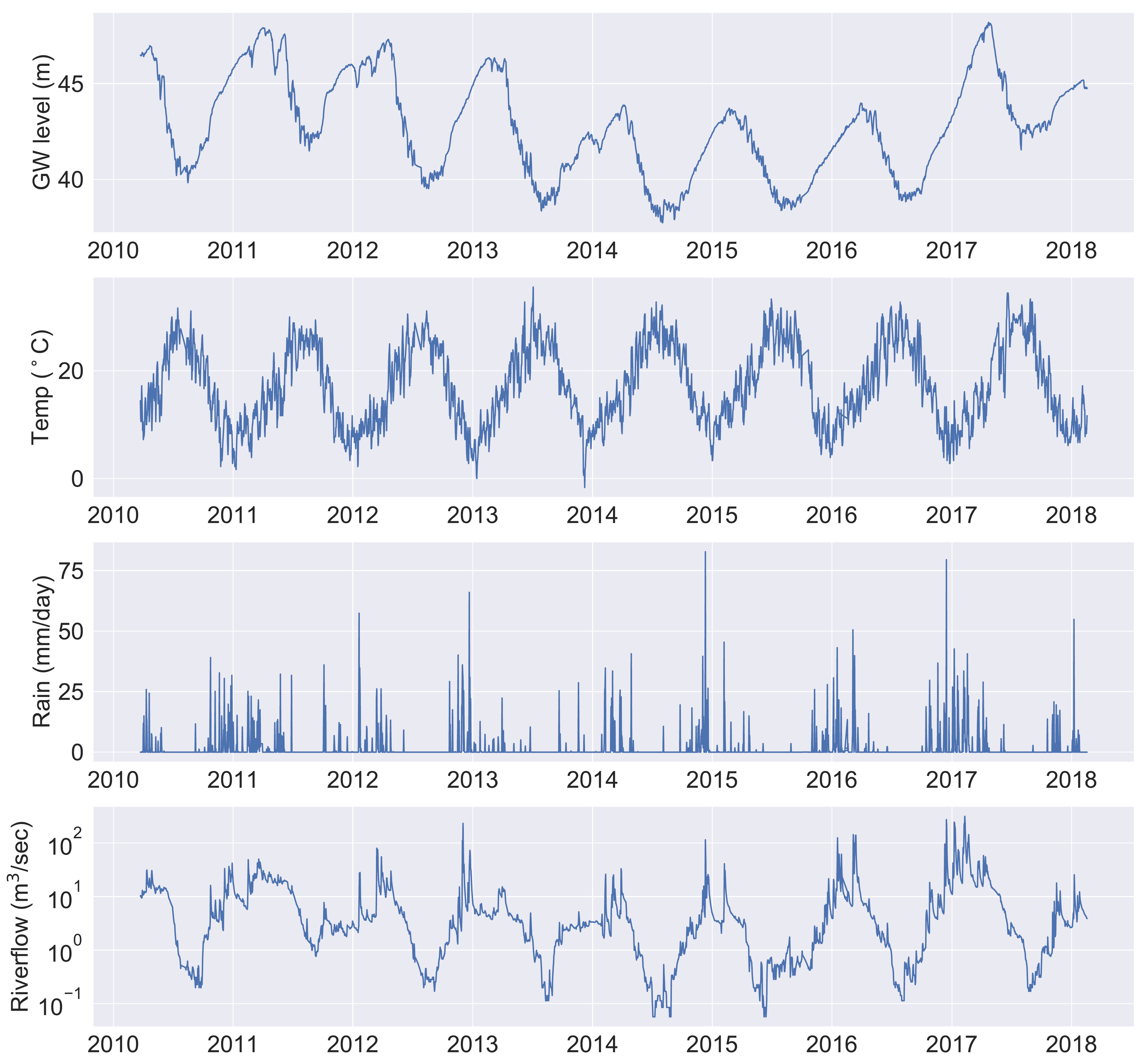}
    \caption{Time series of all observed variables used for training the learning models for  the single  well case. GW=groundwater; Temp = Temperature.}
    \label{single_well_series}
\end{figure}

\begin{figure}[htbp]
    \centering
    \includegraphics[scale=.3]{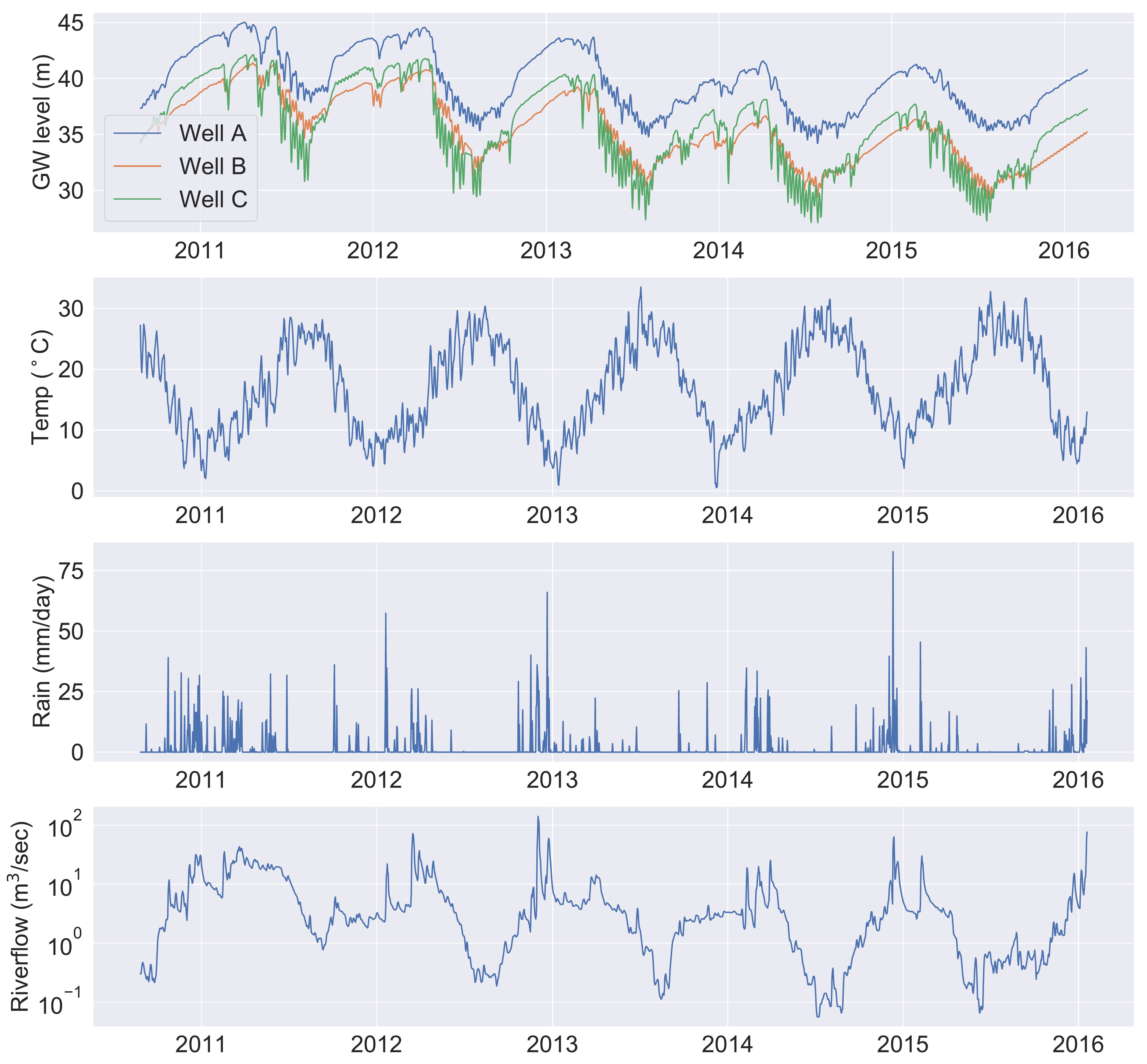}
    \caption{Time series of all observed variables used for training the learning models for the multi-well case. Top panel shows the groundwater (GW) levels of all  three wells.}
    \label{multi_well_series}
\end{figure}

Figure~\ref{single_well_series} shows the data for the single well case. The time series for the groundwater levels clearly reflects the drought years (2012-2016) during which we observe a steadily decreasing trend of groundwater levels. 2017 was a ``wet'' year, during which precipitation amounts were higher and thus also groundwater levels recovered more.  For the multi-well case, we only have daily observations between 2010 and 2016. Also in this case we observe the steadily decreasing trend  in groundwater levels in the drought years (2012 - 2016), see Figure~\ref{multi_well_series}.  

\begin{table}[htbp]
\caption{Data sources of historical observations of different measurement variables. CNRA=California Natural Resources Agency; CDEC=California Data Exchange Center.}
    \label{tab:data_source}
    \begin{adjustbox}{width=\textwidth}
    \begin{tabular}{llll}
    \hline\noalign{\smallskip}
         Measurement variable & Data source & Station / well ID or code & Temporal Frequency  \\
    \noalign{\smallskip}\hline\noalign{\smallskip}
         Groundwater level (Single well) & CNRA & 22N01E28J001M & daily average \\
         Groundwater level (Multi-well) & CNRA & 22N01E35E001M   & daily average \\
         && 21N02E30L001M&\\
         && 21N02E18C001M&\\
         Streamflow discharge & CDEC & Butte Creek Durham (BCD) & daily average \\
         Precipitation & CDEC & Chico (CHI)& daily cumulative \\
         Temperature & CDEC & Chico (CHI) & daily average\\
         \noalign{\smallskip}\hline
    \end{tabular}
    \end{adjustbox}
\end{table}

\section{Additional Results of Numerical Experiments}

\subsection{Single Well Case}
Table~\ref{tab:single-results} shows the  results for all five trials with each HPO method for each learning model. The table shows for each learning model and each hyperparameter optimization (HPO) method the optimal hyperparameter values, the mean squared error (MSE) for the testing data, and the corresponding root mean squared error in groundwater levels for the test data  (converted to meters). Note that the convolutional neural net (CNN) is the only model that uses filters and that the number of nodes in the CNN corresponds to the number of filters. Since  the multilayer perceptron (MLP), the recurrent neural network (RNN), and the long short-term memory network (LSTM) do not have a  filter size, the table entries are denoted by a dash.  Note that slightly different test MSE values can lead to the same groundwater RMSE when converted to meters due to rounding.

\begin{table}[htbp]
  \centering
  \caption{Detailed numerical results for all learning models and all HPO methods for each trial for the single well case. MSE = mean squared error; GW RMSE = groundwater root mean squared error; MLP = multilayer perceptron; CNN = convolutional neural network; RNN = recurrent neural network; LSTM = long short-term memory network; RS=random sampling; GP=Gaussian process; RBF=radial basis function.}
   \begin{adjustbox}{width=\textwidth}
     \begin{tabular}{cccccccccccc}
    \toprule
        Model & Method & Trial  & \begin{tabular}{@{}c@{}}Number of \\ epochs\end{tabular} & \begin{tabular}{@{}c@{}}Batch \\ size\end{tabular}  &\begin{tabular}{@{}c@{}} Number of \\ layers\end{tabular}  &\begin{tabular}{@{}c@{}} Number of nodes \\  (or filters) in a layer\end{tabular}  & \begin{tabular}{@{}c@{}} Size of \\ the filter\end{tabular}  & \begin{tabular}{@{}c@{}} Dropout\\ rate\end{tabular}  & Lag   & \begin{tabular}{@{}c@{}} Test\\  MSE\end{tabular} & \begin{tabular}{@{}c@{}} GW RMSE \\  in meter  \end{tabular} \\
    \midrule
  MLP & GP & 1   &  400 &   160   &   3    &   20    &   -    &  0.4  &  95  &  0.00054 & 1.0  \\
        &  & 2   & 150  &  140    &   4    &   35    &   -    &  0.5  &  355 &  0.00031     & 0.8     \\
        &  & 3   & 450  &  130    &   6    &   40    &   -    &  0.5  &  245 & 0.00043     & 0.9 \\
        &  & 4   & 300  &  55     &   6    &   10    &   -    &  0.2  &  235 & 0.00034      & 0.8 \\
        &  & 5   & 450  &  130    &   6    &   40    &   -    &  0.5  &  245 & 0.00048      & 0.9 \\ 
        \cmidrule{2-12}
 & RBF & 1      & 50      &  75     &   5    &   5    &   -    &  0.4     &  305     &  0.00031     & 0.8 \\
    &  & 2      & 50      &  195     &   5    &  5     &   -    &  0.4     &  310     &  0.00028     & 0.7      \\
    &  & 3      & 50      &  160     &   6    &   45    &  -     &  0.5     &  360     &  0.00027     & 0.7  \\
    &  & 4      & 50      &  155     &   5    &    45   &  -     &   0.1    &   335    &  0.00025     & 0.7  \\
    &  & 5      & 50      &  200     &   6   &   15    & -       &   0    &   340    &    0.00019   & 0.6  \\ 
    \cmidrule{2-12}
 & RS & 1   &  50       &  130  &  4     &   5    &  -     &  0.2     &  300     &   0.00050    & 1.0 \\
    &  & 2 &   250      &  105  &  2     &   15    &  -     & 0.5      &  145     &  0.00070     & 1.1       \\
    &  & 3 &   500      &  145  &  6     &   15    &   -    &  0.0     &  240     &  0.00061     & 1.1 \\
    &  & 4   &  50      & 200   &  4     &   35    &    -   &  0.0     &  305     & 0.00053      & 1.0 \\
    &  & 5   &  500     &  75   &  5     &   45    &    -   &  0.5     &  295     &  0.00056     & 1.0 \\ 
    \midrule
CNN & GP & 1     & 350   & 155   & 5     & 20    & 3     & 0.1   & 330   & 0.00039 & 0.9 \\
          &       & 2     & 450   & 130   & 3     & 45    & 3     & 0.3   & 295   & 0.00077 & 1.2  \\
          &       & 3     & 250   & 80    & 6     & 35    & 4     & 0.3   & 350   & 0.00065 & 1.1 \\
          &       & 4     & 350   & 155   & 5     & 20    & 3     & 0.1   & 330   & 0.00034 & 0.8 \\
          &       & 5     & 350   & 155   & 5     & 20    & 3     & 0.1   & 330   & 0.00038 & 0.8  \\ 
          \cmidrule{2-12}
          & RBF & 1     & 150   & 115   & 4     & 40    & 2     & 0.0     & 350   & 0.00036 & 0.8 \\
          &       & 2     & 450   & 50    & 5     & 45    & 2     & 0.1   & 345   & 0.00039 & 0.9 \\
          &       & 3     & 400   & 190   & 5     & 45    & 3     & 0.1   & 355   & 0.00036 & 0.8  \\
          &       & 4     & 350   & 155   & 5     & 20    & 3     & 0.1   & 330   & 0.00037 & 0.8  \\
          &       & 5     & 200   & 165   & 5     & 40    & 4     & 0.3   & 360   & 0.00040 & 0.9 \\ 
          \cmidrule{2-12}
          & RS & 1     & 450   & 105   & 6     & 30    & 2     & 0.3   & 355   & 0.00039 & 0.9      \\
          &       & 2     & 350   & 190   & 5     & 50    & 3     & 0.0     & 335   & 0.00047 & 0.9 \\
          &       & 3     & 250   & 115   & 5     & 25    & 5     & 0.2   & 295   & 0.00065 & 1.1   \\
          &       & 4     & 250   & 100   & 6     & 40    & 4     & 0.3   & 365   & 0.00046 & 0.9    \\
          &       & 5     & 100   & 95    & 6     & 35    & 3     & 0.4   & 360   & 0.00070 & 1.1   \\ 
          \midrule
SimpleRNN & GP &    1  & 300   & 95    & 2     & 35    &   -    & 0.4   & 9     & 0.00206 & 2.0   \\
          &       & 2     & 500   & 85    & 1     & 30    & -      & 0.4   & 54    & 0.00244 & 2.1 \\
          &       & 3     & 400   & 50    & 2     & 5     & -      & 0.1   & 12    & 0.00253 & 2.2 \\
          &       & 4     & 100   & 105   & 4     & 35    & -      & 0.1   & 48    & 0.00230 & 2.1 \\
          &       & 5     & 300   & 95    & 2     & 35    & -      & 0.4   & 9     & 0.00205 & 1.9 \\ 
          \cmidrule{2-12}
          & RBF & 1     & 450   & 195   & 3     & 45    &   -    & 0.4   & 39    & 0.00168 & 1.8     \\ 
          &       & 2     & 450   & 50    & 4     & 45    & -      & 0     & 42    & 0.00157 & 1.7  \\
          &       & 3     & 450   & 55    & 4     & 45    & -      & 0     & 42    & 0.00155 & 1.7  \\
          &       & 4     & 450   & 55    & 1     & 35    & -      & 0.4   & 57    & 0.00182 & 1.8  \\
          &       & 5     & 400   & 100   & 1     & 45    & -      & 0.4   & 42    & 0.00183 & 1.8   \\ 
          \cmidrule{2-12}
          & RS & 1     & 50    & 145   & 4     & 45    &    -   & 0.1   & 51    & 0.00183 & 1.8     \\
          &       & 2     & 500   & 75    & 3     & 45    & -      & 0.3   & 51    & 0.00170 & 1.8  \\
          &       & 3     & 250   & 120   & 5     & 30    & -      & 0.2   & 27    & 0.00212 & 2.0  \\
          &       & 4     & 300   & 140   & 2     & 40    & -      & 0.3   & 57    & 0.00223 & 2.0  \\
          &       & 5     & 450   & 80    & 5     & 30    &  -     & 0.2   & 48    & 0.00169 & 1.8   \\ 
          \midrule
LSTM      & GP & 1 &  200 & 450 & 4     &  30     &   -    &  0.4     &   3    &  0.00065     & 1.1 \\
          &    & 2 &  400 & 200 & 3     &  18     &   -    &  0.2    &    36   &  0.00073     & 1.2   \\
          &    & 3 &  500 & 250 & 3     &  24     &   -    &  0.3   &     75  &   0.00080    & 1.2 \\
          &    & 4 &  400 & 150 & 3      &  21     &  -     &  0.1     &  42     &  0.00080  & 1.2 \\
          &    & 5 &  500 & 100 & 4      &  30     &   -    &   0.2    &   72    &  0.00072  & 1.1 \\ 
          \cmidrule{2-12}
 & RBF  & 1 &  200 &  200     &   4    &   24    &   -    &   0.1    &   72    &  0.00068     & 1.1 \\
 &      & 2 &  200 &  250     &   3    &    6   &    -   &    0.4 &    48   &    0.00076   &   1.2     \\
 &      & 3 &  200 &  400     &   3    &    6  &     -  &    0.0    &   72    &  0.00067     & 1.1 \\
 &      & 4 &  400 &  450     &   3    &    27  &    -   &  0.1     &    3   &   0.00074    &  1.2 \\
 &      & 5 &  400 &  100     &   3    &    18   &   -    &  0.0     &   3    &  0.00077     & 1.2 \\ 
 \cmidrule{2-12}
 & RS & 1   &  200     &  450     &   2    &  30     &   -    &   0.4    &  9     &  0.00216     & 2.0 \\
   &  & 2   & 400      &  250     &   4    & 15      &   -    &   0.5      & 9      & 0.00254      & 2.2    \\
 &    & 3   &  200     &   300    &   4   &  21     &    -   &    0.5     &   3    &   0.00181    & 1.8 \\
 &    & 4   &  200     &   450    &   4    &  30     &   -    &   0.4      &  3     &  0.00217     & 2.0\\
 &    & 5   & 300      &   500    &   2    &  24       &  -     &  0.0     &  3     &  0.00240     & 2.1 \\
    \bottomrule
    \end{tabular}%
      \end{adjustbox}
  \label{tab:single-results}%
\end{table}%

\subsubsection{Distributions of Optimized Hyperparameters for RNN, CNN, and LSTM}
\begin{figure}
    \centering
    \begin{subfigure}[b]{0.48\textwidth}
        \includegraphics[width=\textwidth]{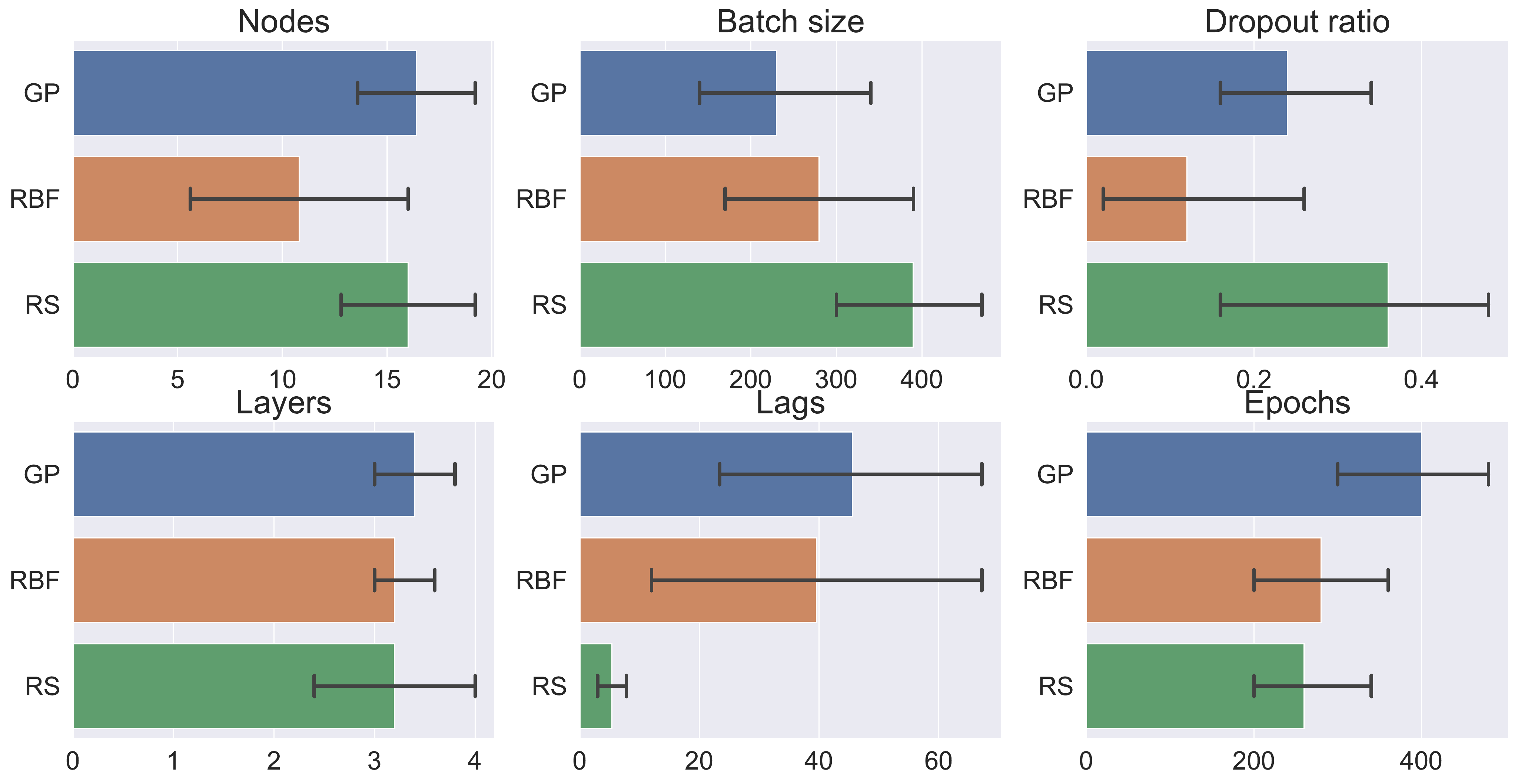}
        \caption{LSTM}
        \label{fig:single_well_LSTM_hyperparameter}
    \end{subfigure}
    ~ 
    \begin{subfigure}[b]{0.48\textwidth}
        \includegraphics[width=\textwidth]{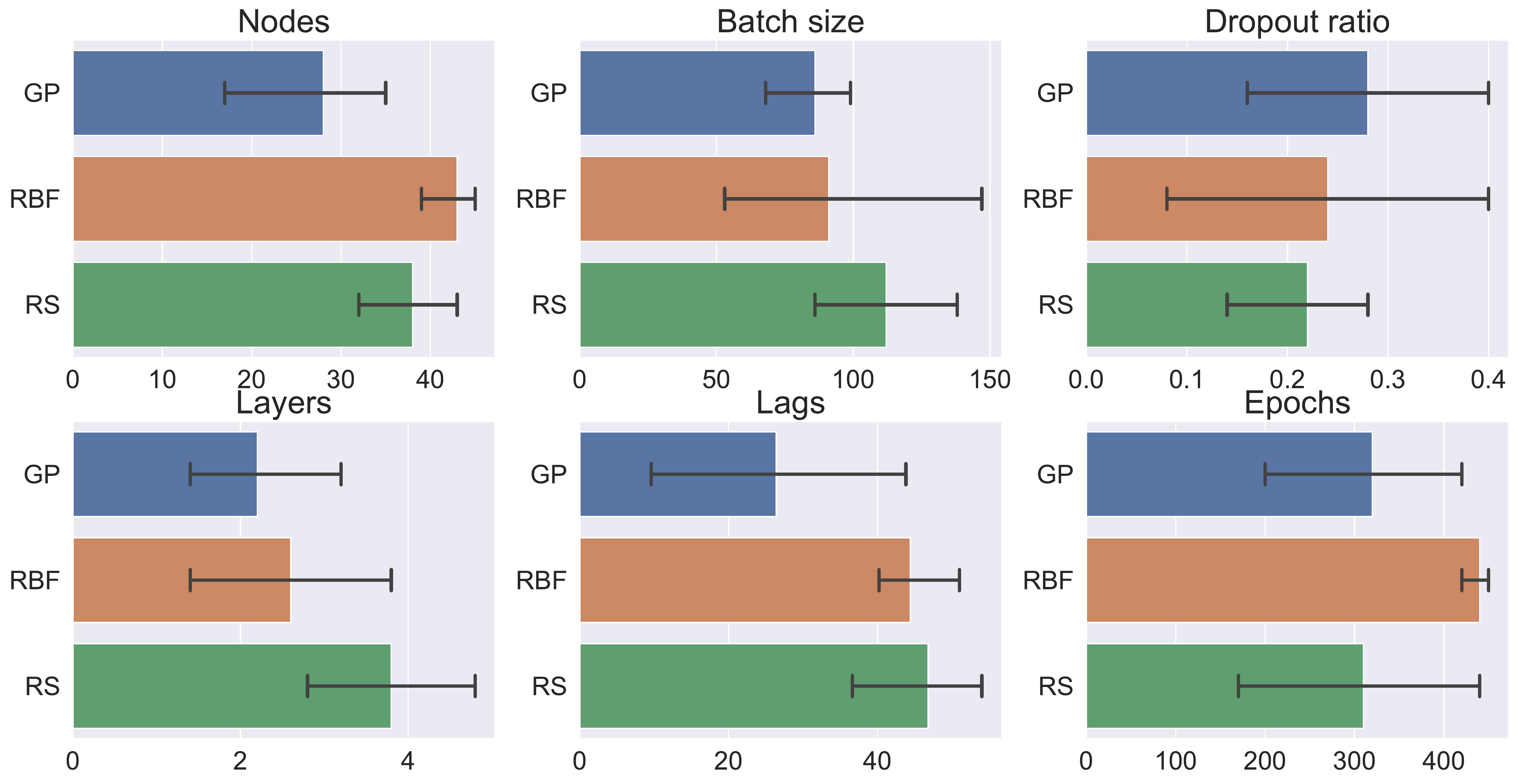}
        \caption{RNN}
        \label{fig:single_well_RNN_hyperparameter}
    \end{subfigure}
    ~ 
    \begin{subfigure}[b]{0.45\textwidth}
        \includegraphics[width=\textwidth]{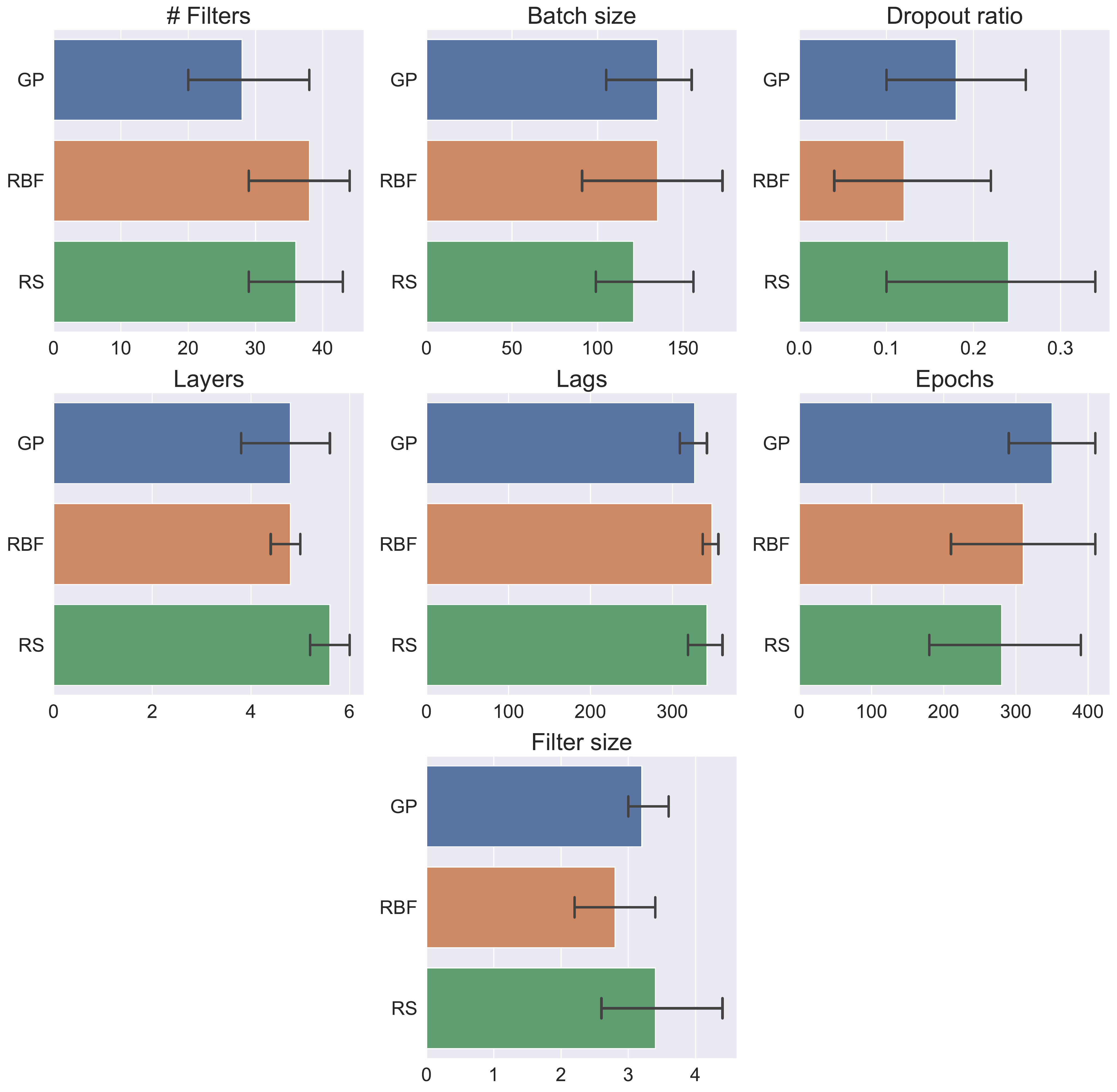}
        \caption{CNN}
        \label{fig:single_well_CNN_hyperparameter}
    \end{subfigure}
    \caption{Distribution of the optimized hyperparameter values for LSTM, RNN, and CNN for the single well case. Colored bars indicate mean values and black lines indicate standard deviation. RS=random sampling; GP=Gaussian process; RBF=radial basis function.}
    \label{fig:hyp_singlewell}
\end{figure}

Here, we show additional results of the distribution of the optimal hyperparameters for all models and all HPO methods over all five trials. The results for LSTM are shown in Figure~\ref{fig:single_well_LSTM_hyperparameter}. For the LSTM, the GP and the RBF approaches performed better than the random search. We note that the random search prefers very small lags, which may be the reason for its inferior performance.   

The distributions of the optimal hyperparameters for CNN over all five trials and for each HPO method are shown in Figure~\ref{fig:single_well_CNN_hyperparameter}. All models appear to favor networks with a large number of layers and large lags (over 300 days). The most significant  difference can be observed for the dropout ratio, but all three HPO methods have a large variability for this hyperparameter. 

The distributions of the optimal hyperparameters for the RNN model are shown in Figure~\ref{fig:single_well_RNN_hyperparameter}. Note that for RNN several optimization trials failed due to the error backpropagation problem, and thus the results shown are from five successful optimizations out of several more trials. No HPO method is able to improve the RNN performance to one comparable with the other learning models. We conclude that   given  the available hyperparameter options,  the RNN is not a suitable option for learning and predicting  groundwater data. 

\subsubsection{Performance of CNN and RNN on Testing Data}
\begin{figure}[htbp]
    \centering
    \begin{subfigure}[b]{0.48\textwidth}
        \includegraphics[width=\textwidth]{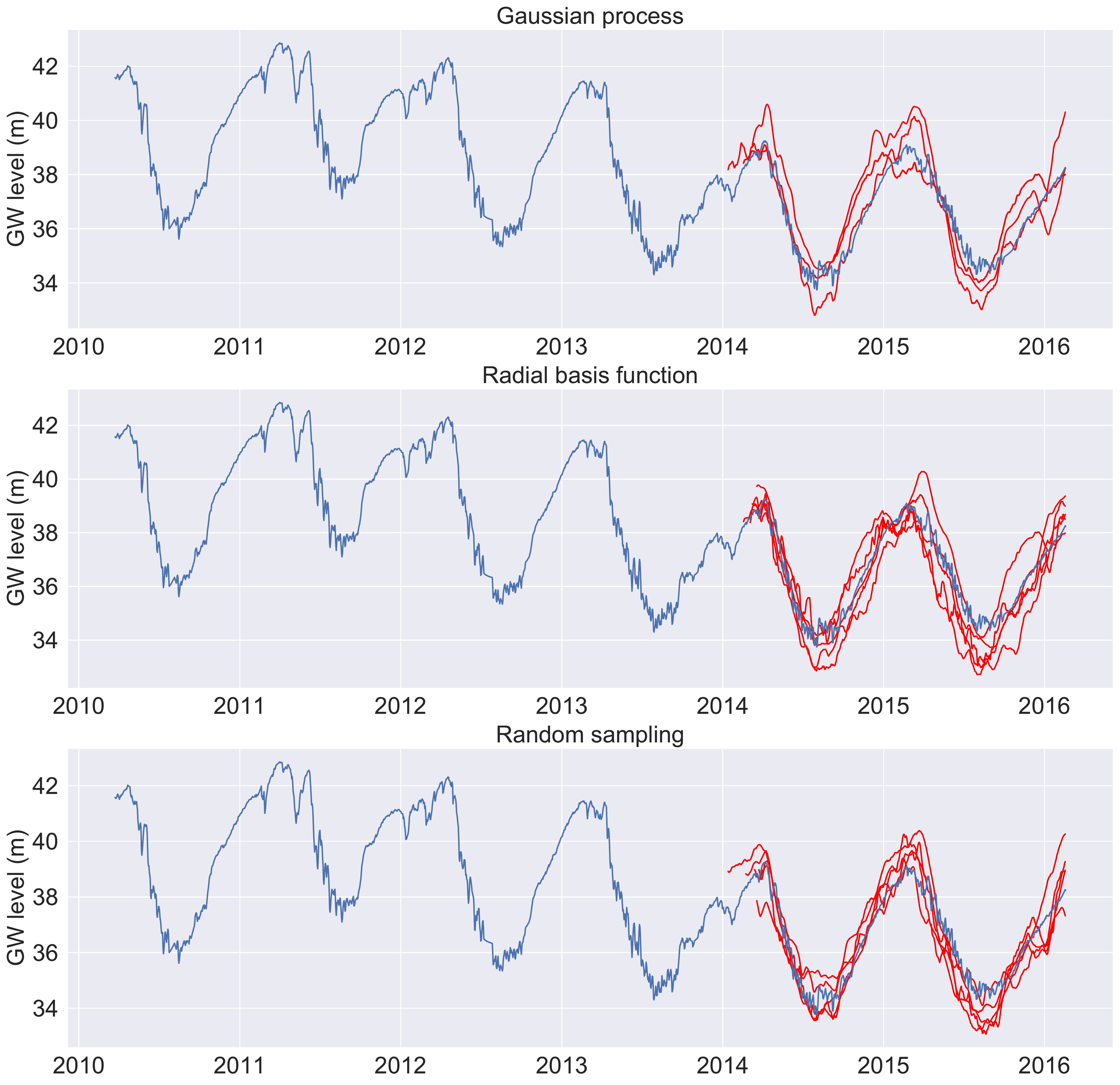}
        \caption{CNN}
        \label{fig:single_well_CNN_GW_pred}
    \end{subfigure}
    ~ 
    \begin{subfigure}[b]{0.48\textwidth}
        \includegraphics[width=\textwidth]{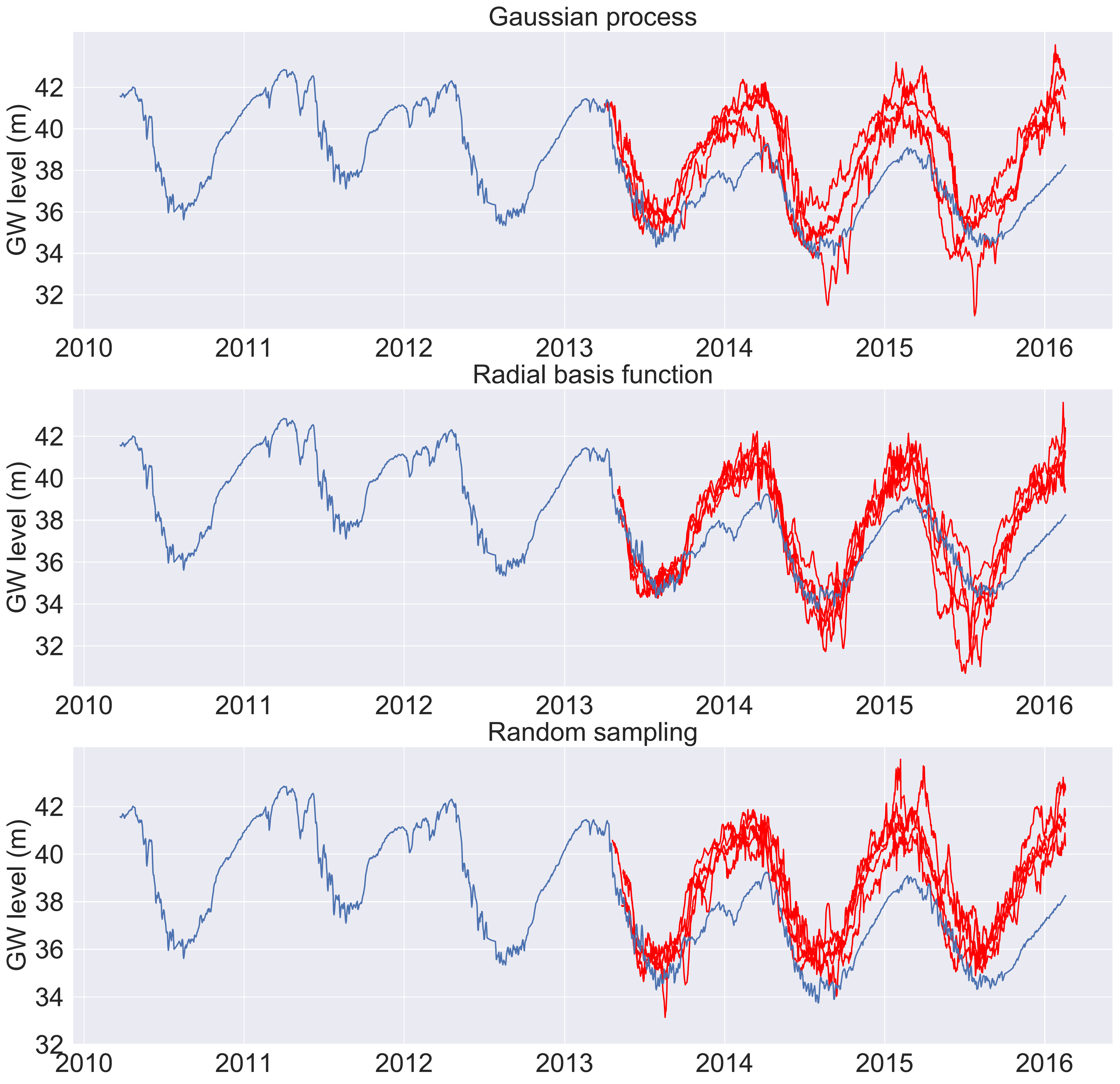}
        \caption{RNN}
        \label{fig:single_well_RNN_GW_pred}
    \end{subfigure}
    \caption{Groundwater (GW) level predictions obtained from five hyperparameter optimizations of CNN (left) and RNN (right) for the single well case for each HPO method.  Blue graphs are the true data and red graphs are the model predictions. Ideally, red and blue graphs agree. }
    \label{fig:oos_singlewell}
\end{figure}

We illustrate the groundwater time series data (training and testing data) and we illustrate the predictions of the learning models for the testing data. Figure~\ref{fig:single_well_CNN_GW_pred} shows our results for the CNN model and Figure~\ref{fig:single_well_RNN_GW_pred} shows the results for the RNN model. The five trials with the CNN lead to  good results for all three HPO methods and all trials lie close to the true data and follow the seasonality.   Note that for the CNN optimized with the GP approach, three trials led to the exact same hyperparameters (see also Table 2 above). The MSE values are slightly different (due to the stochastic optimizer that solves the lower level problem, but the  predictions for the testing data are very similar and thus, in the upper panel of Figure~\ref{fig:single_well_CNN_GW_pred}, it looks like only three red graphs are shown. On the other hand, the  RNN's  groundwater predictions show larger deviations from the truth, but the seasonality is captured.

\subsubsection{Out-of-sample Future Groundwater Predictions for  CNN and RNN}

\begin{figure}[htbp]
    \centering
    \begin{subfigure}[b]{0.48\textwidth}
        \includegraphics[width=\textwidth]{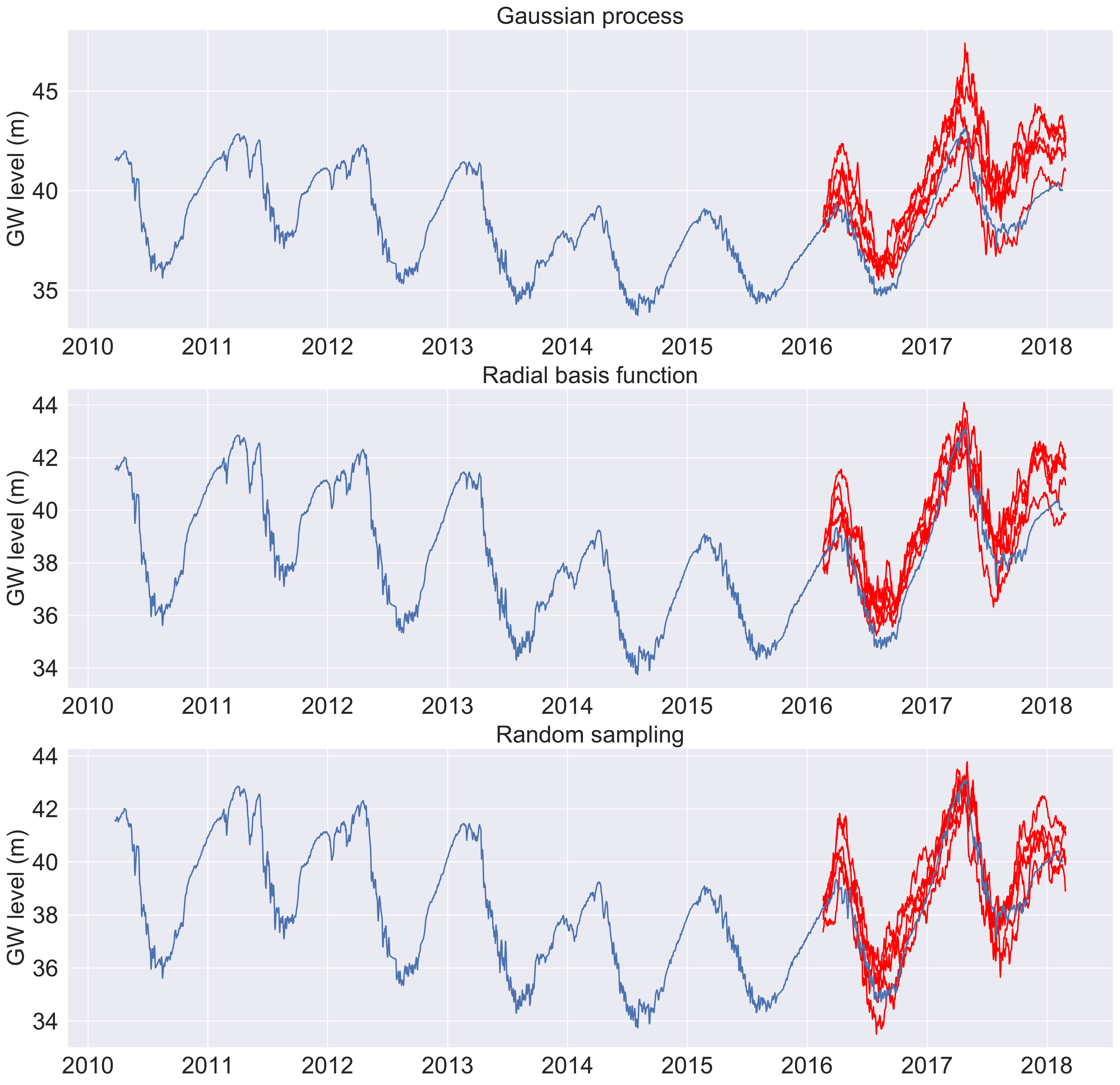}
        \caption{CNN}
        \label{fig:single_well_CNN_GW_pred2}
    \end{subfigure}
    ~ 
    \begin{subfigure}[b]{0.49\textwidth}
        \includegraphics[width=\textwidth]{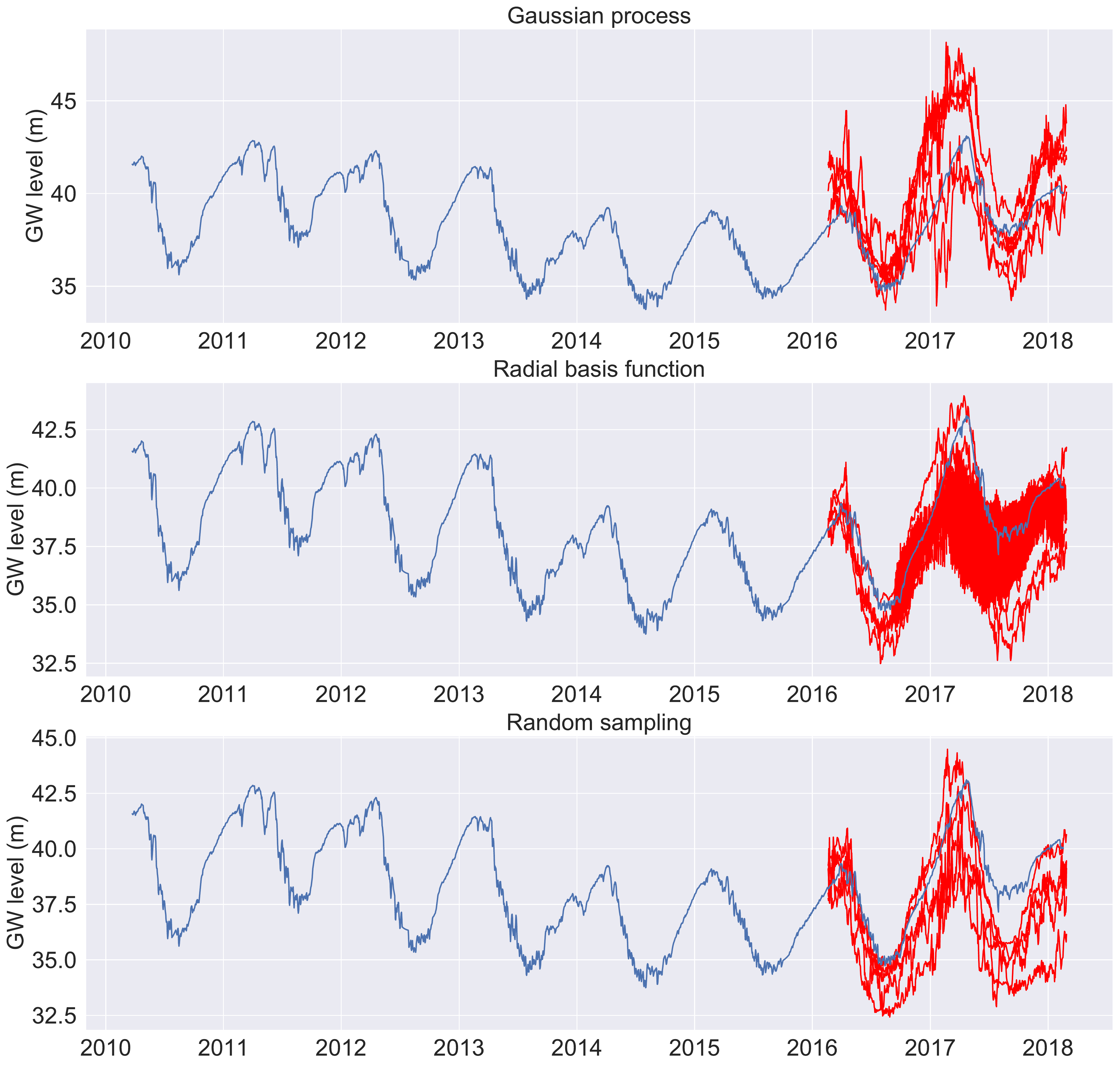}
        \caption{RNN}
        \label{fig:single_well_RNN_GW_pred2}
    \end{subfigure}
    \caption{Out-of-sample groundwater (GW)  predictions with the five optimized CNN (left) and RNN (right) models for all three HPO methods. Blue graphs are the true data and red graphs are the model predictions. Ideally, red and blue graphs agree. }
    \label{fig:oos_singlewell}
\end{figure}

Figures~\ref{fig:single_well_CNN_GW_pred2} and~\ref{fig:single_well_RNN_GW_pred2} show the future groundwater predictions obtained with the optimized CNN and RNN models, respectively. The observed groundwater level values for 2016-2018 were not used in training the models or for optimizing their hyperparameters. We use these ``out-of-sample" truth values and  compare them to the predictions made with CNN and RNN. We can see that  the  seasonality in the data (lower groundwater levels in summer, higher levels in winter) is captured -- better by the CNN than the RNN. The RNN shows a large variability in the predictions. For all learning models it holds that the prediction variability increases when we predict further  into the future. 

\begin{figure}[htbp]
    \centering
    \includegraphics[scale=.18]{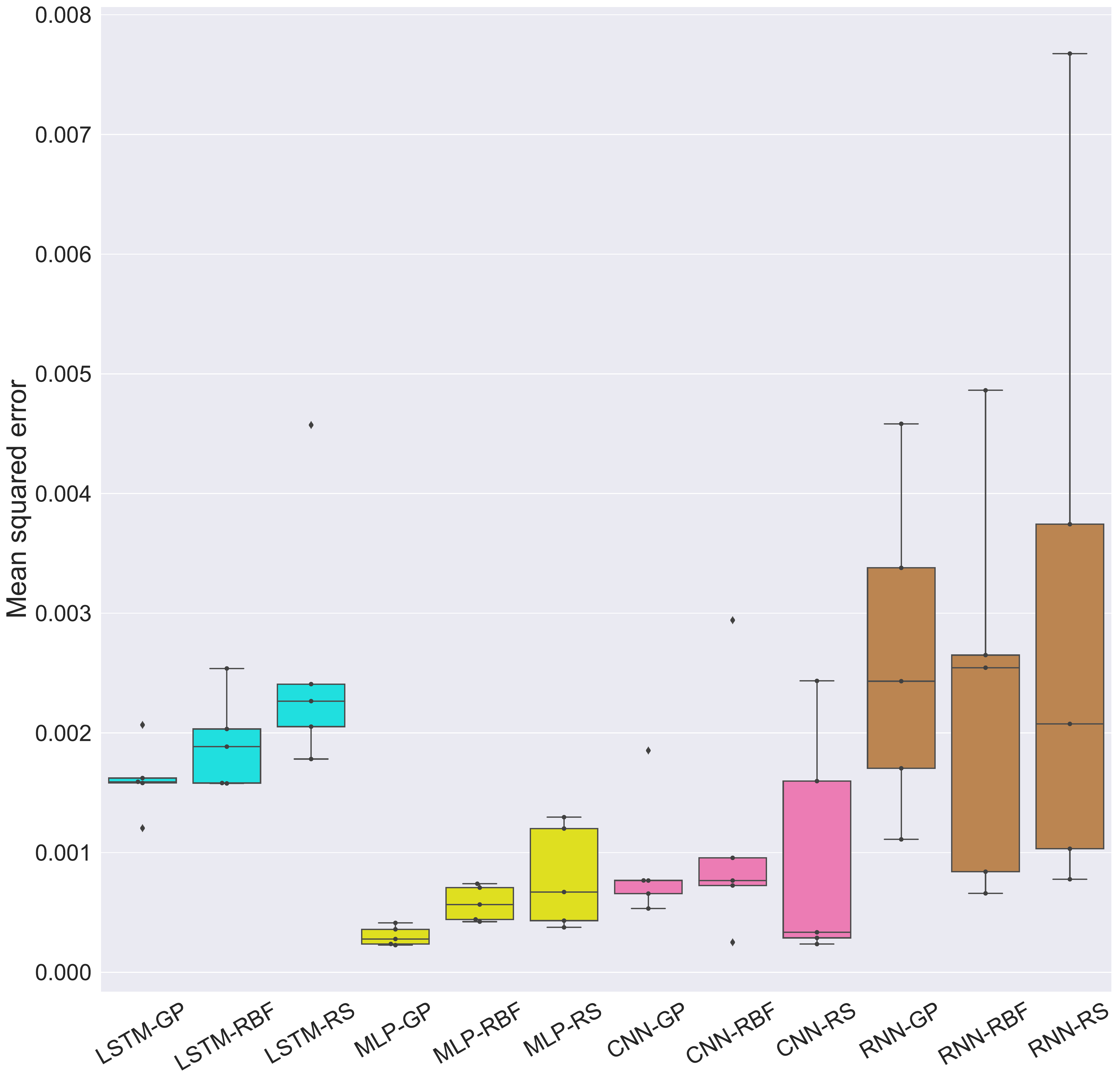}
    \caption{Boxplots of  mean squared errors of out-of-sample predictions for the single well case. Low values are better.  LSTM=long short-term memory network; MLP=multilayer perceptron; CNN=convolutional neural network; RNN=recurrent neural network; GP=Gaussian process; RBF=radial basis function; RS = random sampling.}
    \label{fig:MSE_Single_well_OutofSample_boxplot}
\end{figure}

Figure~\ref{fig:MSE_Single_well_OutofSample_boxplot} shows the distributions of the out-of-sample mean squared errors for all HPO methods. As can be expected, the errors are larger than for the testing data. However, MLP and CNN perform best, which agrees with the errors reported for the testing data. The performance of RNN is highly variable, which is also reflected in the graphs in Figure~\ref{fig:single_well_RNN_GW_pred2}.

\newpage
\subsection{Additional Results for the Multi-Well Case}
Table~\ref{tab:multi-results} shows the results of all numerical experiments for the multi-well case. For this case, we did not run the RNN model due to its numerical problems during backpropagation as outlined in the main document. Therefore, we show the results only for LSTM, CNN, and MLP. We find that both the CNN and MLP outperform the LSTM.  We note that the optimized LSTM has in most trials only a single layer. The optimized CNN and MLP models, on the other hand, are generally larger with multiple layers.

\begin{table}[htbp]
  \centering
  \caption{Detailed numerical results for CNN, LSTM, and MLP for all HPO methods for each trial for the multi-well case. GW RMSE = groundwater root mean squared error; MLP = multilayer perceptron; CNN = convolutional neural network;  LSTM = long short-term memory network; RS=random sampling; GP=Gaussian process; RBF=radial basis function.}
   \begin{adjustbox}{width=\textwidth}
     \begin{tabular}{ccccccccccc}
    \toprule
        Model & Method & Trial  & \begin{tabular}{@{}c@{}}Number of \\ epochs\end{tabular} & \begin{tabular}{@{}c@{}}Batch \\ size\end{tabular}  &\begin{tabular}{@{}c@{}} Number of \\ layers\end{tabular}  &\begin{tabular}{@{}c@{}} Number of nodes \\  (or filters) in a layer\end{tabular}  & \begin{tabular}{@{}c@{}} Size of \\ the filter\end{tabular}  & \begin{tabular}{@{}c@{}} Dropout\\ rate\end{tabular}  & Lag   & \begin{tabular}{@{}c@{}} Test\\  MSE\end{tabular} 
        \\
        
    \midrule
MLP & GP & 1     & 50    & 175   & 2     & 45    & -     & 0.4   & 260   & 0.00166 \\
          &       & 2     & 150   & 65    & 5     & 20    & -     & 0.1   & 265   & 0.00147 \\
          &       & 3     & 50    & 170   & 3     & 40    & -     & 0.4   & 275   & 0.00174 \\
          &       & 4     & 50    & 175   & 2     & 45    & -     & 0.4   & 260   & 0.00166 \\
          &       & 5     & 250   & 145   & 6     & 30    & -     & 0.5   & 365   & 0.00118 \\ \cmidrule{2-11}
 & RBF & 1     & 200   & 195   & 5     & 25    & -     & 0.4   & 355   & 0.00086 \\
          &       & 2     & 50    & 105   & 5     & 45    & -     & 0.3   & 275   & 0.00149 \\
          &       & 3     & 50    & 110   & 4     & 15    & -     & 0.0     & 280   & 0.00114 \\
          &       & 4     & 50    & 80    & 5     & 45    & -     & 0.1   & 355   & 0.00071 \\
          &       & 5     & 50    & 190   & 4     & 45    & -     & 0.2   & 280   & 0.00091 \\ \cmidrule{2-11}
 & Random &1     & 200   & 140   & 5     & 40    & -     & 0.1   & 355   & 0.00072 \\
          &       & 2     & 500   & 145   & 6     & 30    & -     & 0.2   & 300   & 0.00130 \\
          &       & 3     & 450   & 115   & 6     & 20    & -     & 0.0     & 320   & 0.00107 \\
          &       & 4     & 50    & 200   & 4     & 35    & -     & 0.0     & 305   & 0.00110 \\
          &       & 5     & 450   & 190   & 3     & 40    & -     & 0.0     & 265   & 0.00129 \\ \midrule
CNN & GP & 1     & 155   & 350   & 5     & 20    & 3     & 0.1   & 330   & 0.00122 \\
          &       & 2     & 70    & 200   & 1     & 35    & 3     & 0.5   & 310   & 0.00115 \\
          &       & 3     & 75    & 200   & 6     & 35    & 4     & 0.3   & 355   & 0.00073 \\
          &       & 4     & 70    & 200   & 1     & 35    & 3     & 0.5   & 310   & 0.00115 \\
          &       & 5     & 70    & 200   & 1     & 35    & 3     & 0.5   & 310   & 0.00115 \\ \cmidrule{2-11}
          & RBF & 1     & 130   & 50    & 4     & 35    & 2     & 0.0     & 345   & 0.00054 \\
          &       & 2     & 50    & 50    & 2     & 45    & 2     & 0.0     & 355   & 0.00066 \\
          &       & 3     & 145   & 350   & 6     & 35    & 5     & 0.5   & 310   & 0.00102 \\
          &       & 4     & 185   & 50    & 5     & 40    & 3     & 0.3   & 360   & 0.00095 \\
          &       & 5     & 60    & 300   & 6     & 50    & 2     & 0.1   & 360   & 0.00072 \\ \cmidrule{2-11}
          & RS & 1     & 150   & 50    & 4     & 50    & 4     & 0.5   & 300   & 0.00070 \\
          &       & 2     & 170   & 50    & 6     & 40    & 4     & 0.2   & 320   & 0.00085 \\
          &       & 3     & 115   & 250   & 5     & 25    & 5     & 0.2   & 295   & 0.00102 \\
          &       & 4     & 100   & 250   & 6     & 40    & 4     & 0.3   & 365   & 0.00072 \\
          &       & 5     & 125   & 250   & 5     & 45    & 4     & 0.0     & 255   & 0.00077 \\ \midrule
LSTM & GP   & 1 & 500 & 100 & 1 & 21 & - & 0.3 & 39 & 0.00228 \\
        &   & 2 & 500 & 50 & 1 & 21 & - & 0.2 & 39 & 0.00066 \\
        &   & 3 & 500 & 50 & 1 & 24 & - & 0.3 & 60 & 0.00056 \\
        &   & 4 & 400 & 150 & 1  & 24 & - & 0.1 & 9 & 0.00271 \\
        &   & 5 & 400 & 100 & 1 & 30 & - & 0.5 & 21 & 0.00229 \\ \cmidrule{2-11}
    & RBF   & 1 & 400 & 100 & 1 & 27 & - & 0.2 & 42 & 0.00261 \\
    &       & 2 & 500 & 100 & 1 & 24 & - & 0.4 & 48 & 0.00076 \\
    &       & 3 & 400 &  50 & 1 & 27 & - & 0.4 & 21 & 0.00069 \\
    &       & 4 & 200 &  50 & 1 & 27 & - & 0.0 &  6 & 0.00339 \\
    &       & 5 & 400 & 100 & 1 & 27 & - & 0.0 & 51 & 0.00243\\ \cmidrule{2-11}
    & Random &1 & 200 & 450 & 1 & 15 & - & 0.2 & 21 & 0.00198 \\
    &       & 2 & 200 & 150 & 1 & 9 & - & 0.0 & 3 & 0.00194 \\
    &       & 3 & 200 & 450 & 3 & 24 & - & 0.1 & 6 & 0.00169 \\
    &       & 4 & 200 & 500 & 4 & 21 & - & 0.3 & 3 & 0.00154 \\
    &       & 5 & 200 & 450 & 2 & 12 & - & 0.1 & 2 & 0.00149\\ 
    \bottomrule
    \end{tabular}%
      \end{adjustbox}
  \label{tab:multi-results}%
\end{table}%

\subsubsection{Analysis of Optimized Hyperparameters}
\begin{figure}[htbp]
    \centering
    \begin{subfigure}[b]{0.48\textwidth}
        \includegraphics[width=\textwidth]{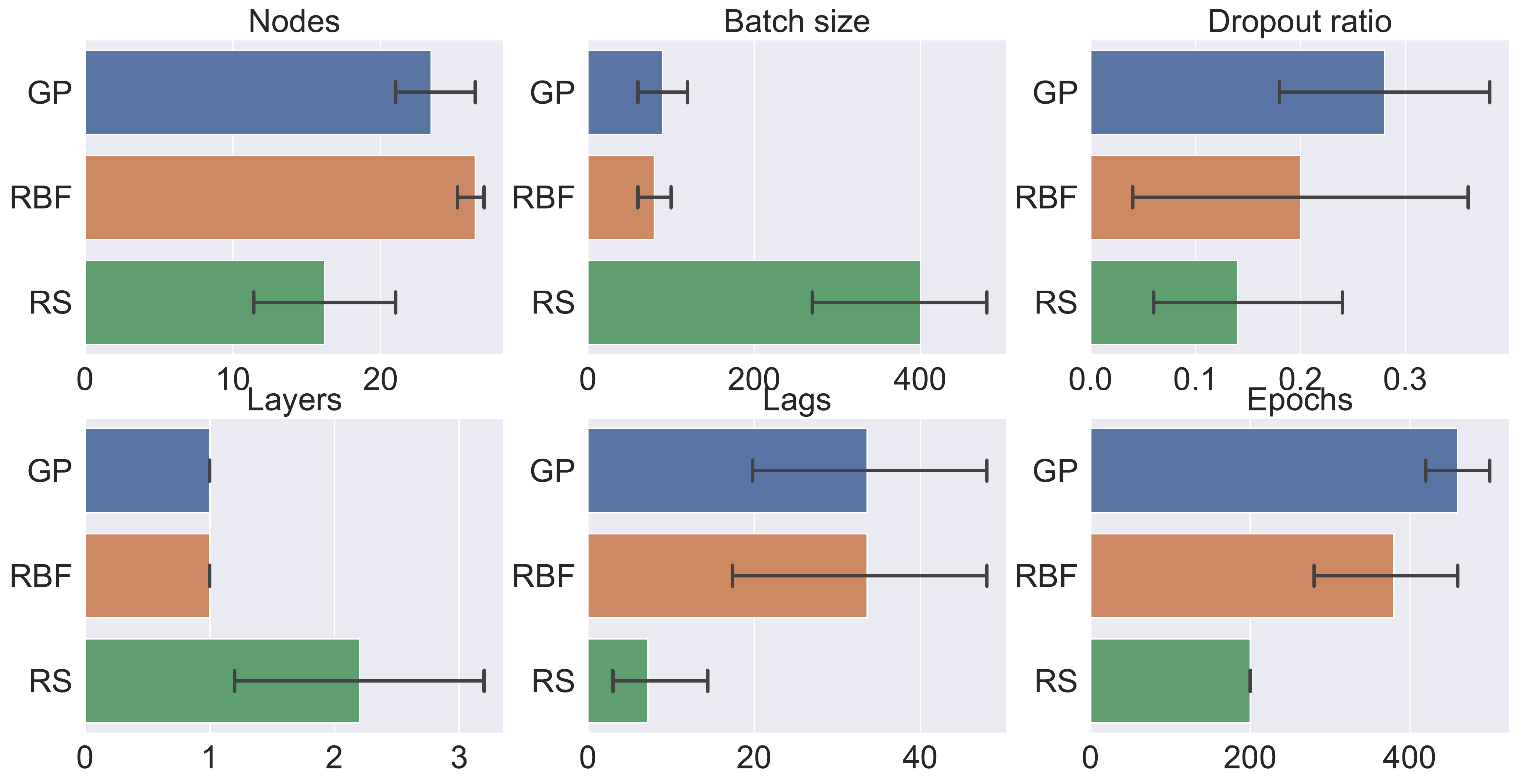}
        \caption{LSTM}
        \label{fig:Multi_well_LSTM_hyperparameter}
    \end{subfigure}
    ~ 
    \begin{subfigure}[b]{0.48\textwidth}
        \includegraphics[width=\textwidth]{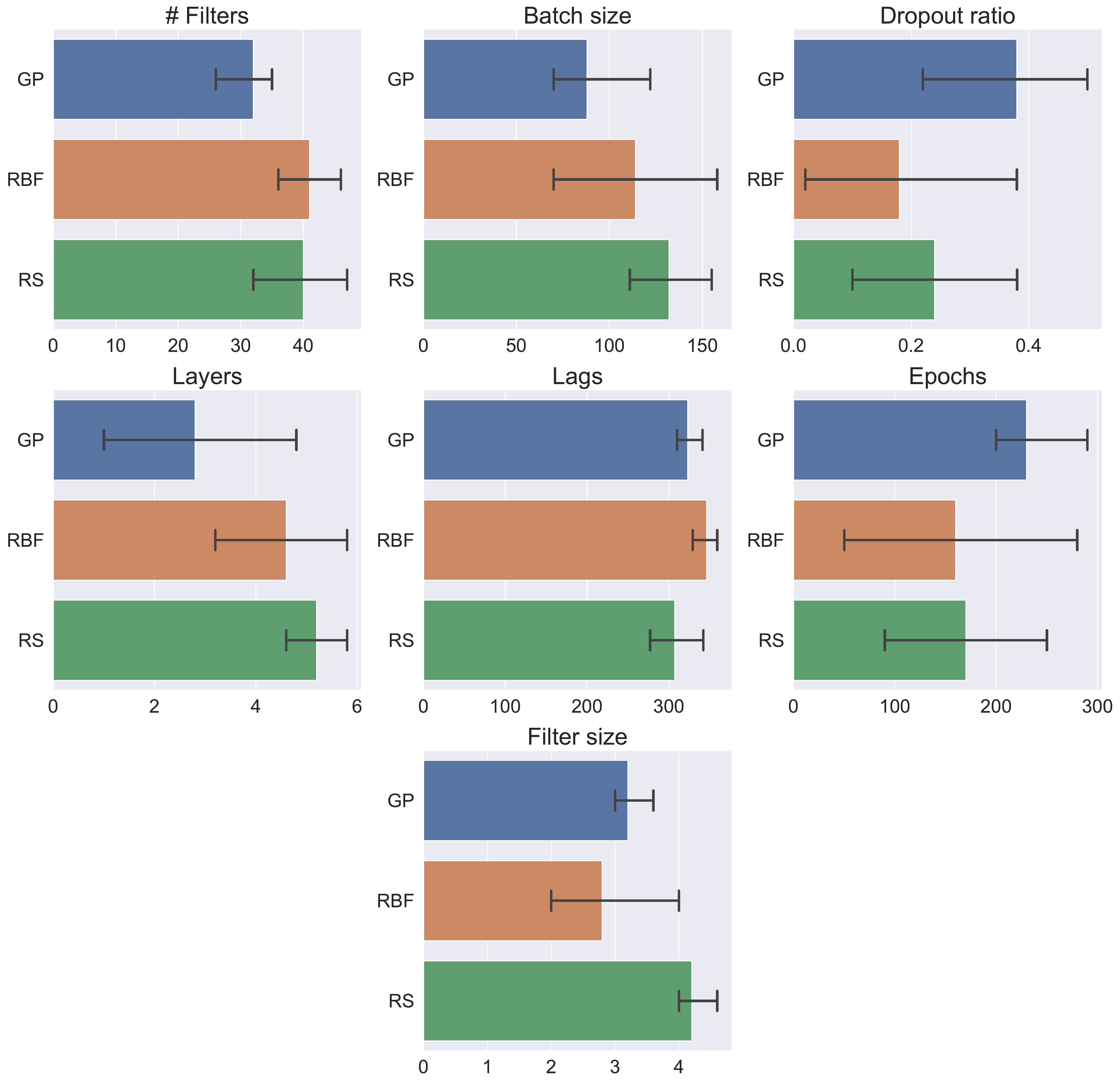}
        \caption{CNN}
        \label{fig:Multi_well_CNN_hyperparameter}
    \end{subfigure}
    \caption{Distribution of the optimized hyperparameter values for LSTM (left) and CNN (right) for the multi-well case. Colored bars indicate mean values and black lines indicate standard deviation. RS=random sampling; GP=Gaussian process; RBF=radial basis function.}
    \label{fig:hyp_multiwell}
\end{figure}

We show the distributions of the optimal hyperparameters for LSTM and CNN in Figures~\ref{fig:Multi_well_LSTM_hyperparameter} and~\ref{fig:Multi_well_CNN_hyperparameter}, respectively. For LSTM,  RBF and GP find that a large number of epochs and a small number of layers (1)  lead to the best  results. For the batch size, both GP and RBF agree that a small number is better. We  find in general that RS leads  to very different optimal hyperparameters. 

For CNN, we observe that the HPO methods find similar optimal values for the batch size, the  lag, the number of layers,  and the number of filters. For the remaining hyperparameters, there is more variability. The GP tends to choose smaller networks (smaller number of filters) than the RBF. 

\newpage
\subsubsection{Convergence  of LSTM and CNN}
\begin{figure}[htbp]
    \centering
    \begin{subfigure}[b]{0.48\textwidth}
        \includegraphics[width=\textwidth]{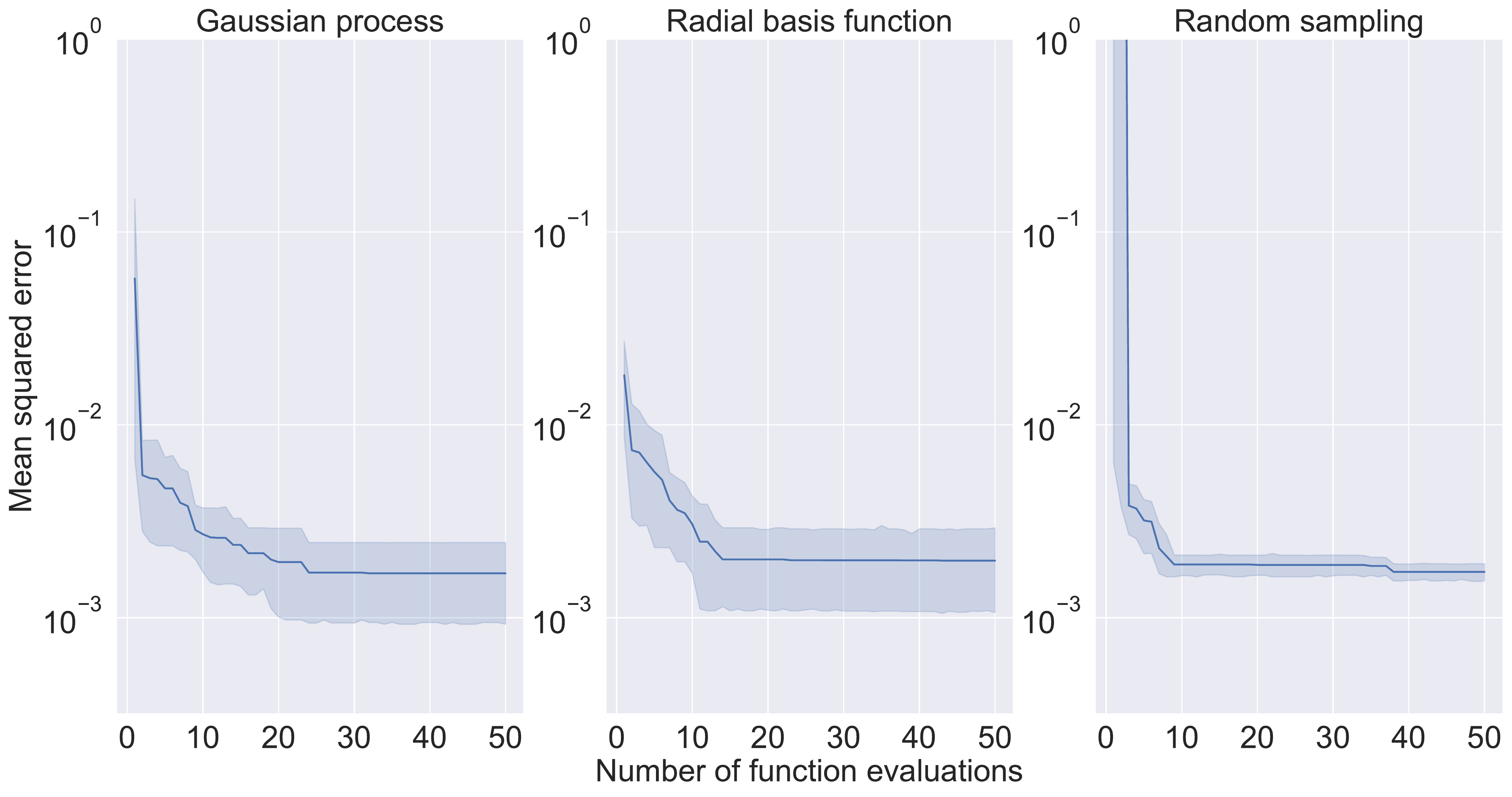}
        \caption{LSTM}
        \label{fig:Multi_well_error_LSTM}
    \end{subfigure}
    ~ 
    \begin{subfigure}[b]{0.48\textwidth}
        \includegraphics[width=\textwidth]{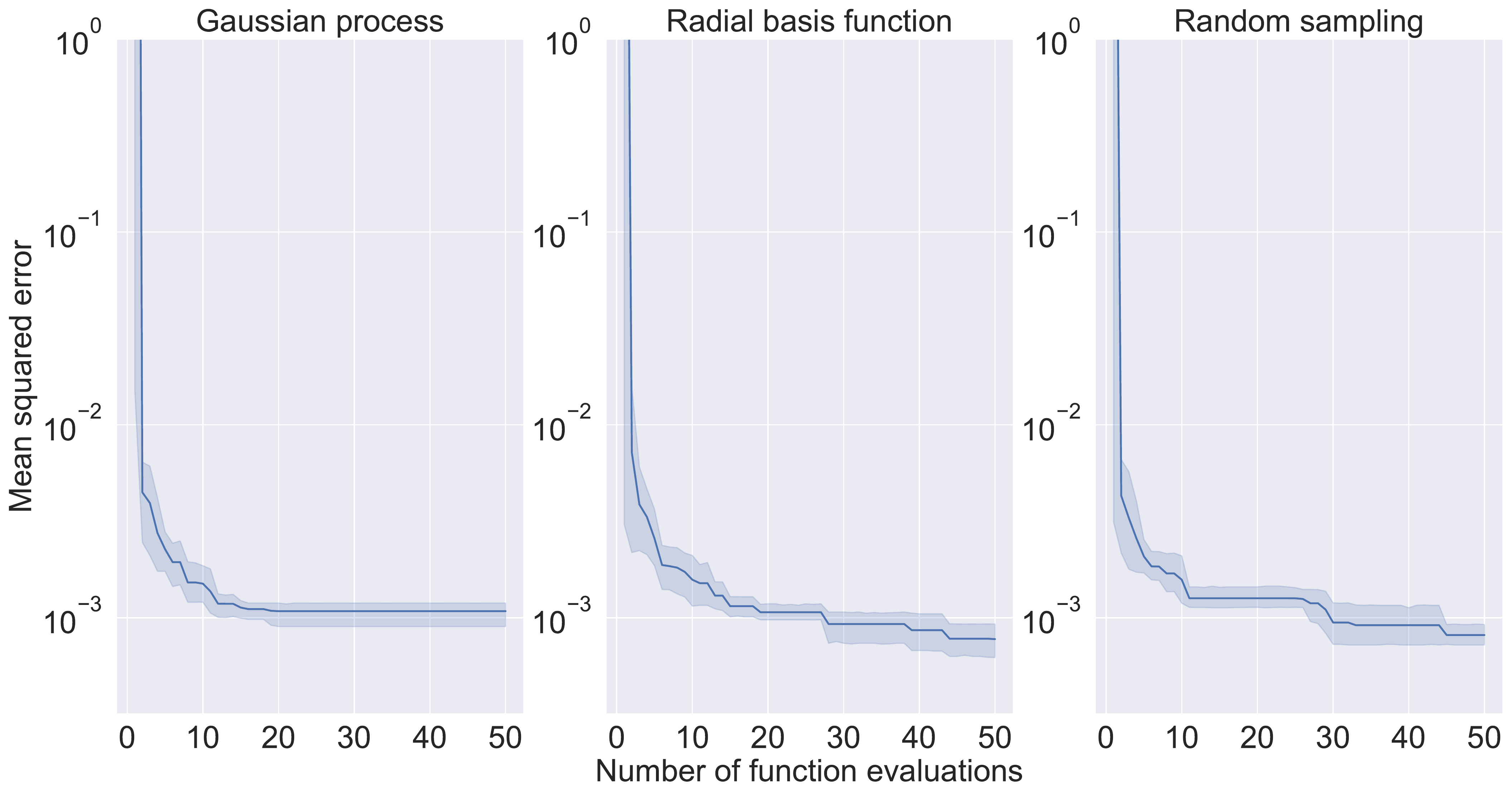}
        \caption{CNN}
        \label{fig:Multi_well_error_CNN}
    \end{subfigure}
    \caption{Convergence plot showing the average and the standard deviation of the  mean squared error over five trials versus the number of function evaluations for  LSTM (left) and CNN (right) for the multi-well case. }
    \label{fig:conv_multiwell}
\end{figure}

We are interested in how many hyperparameter sets each HPO method needs to converge. We illustrate this in the progress plots in Figures~\ref{fig:Multi_well_error_LSTM} and~\ref{fig:Multi_well_error_CNN} for LSTM and CNN. We show the average and the standard deviation of the best mean squared error found as they evolve with the number of evaluations. The faster the graphs drop off, the better because it means that improved solutions are found within fewer upper-level function evaluations. For the LSTM, we observe that RS makes  fast progress and outperforms RBF and GP. Also its standard deviation is smaller. However, since  RS has no systematic way for iteratively selecting new sample points, its  performance and progress are highly dependent on the goodness of the  randomly chosen parameter sets (compare the performance in the single well case reported in the main document in Figure~1).  For the CNN (Figure~\ref{fig:Multi_well_error_CNN}) all three HPO methods make approximately equally good progress.  

\subsubsection{Analysis of Performance on Testing Data for LSTM and CNN}
\begin{figure}[htbp]
    \centering
    \begin{subfigure}[b]{0.48\textwidth}
        \includegraphics[width=\textwidth]{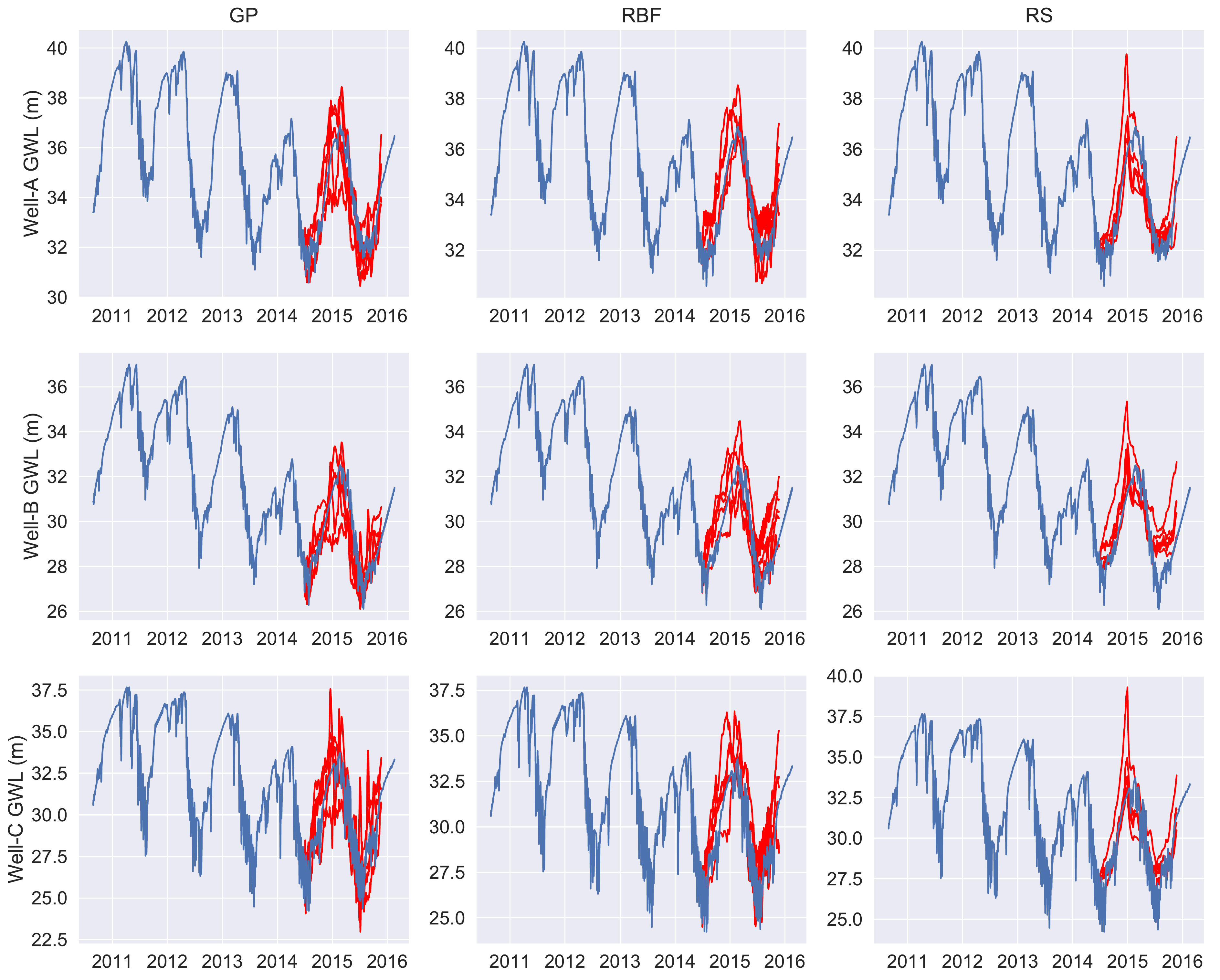}
        \caption{LSTM}
        \label{fig:Multi_well_LSTM_GW_pred}
    \end{subfigure}
    ~ 
    \begin{subfigure}[b]{0.48\textwidth}
        \includegraphics[width=\textwidth]{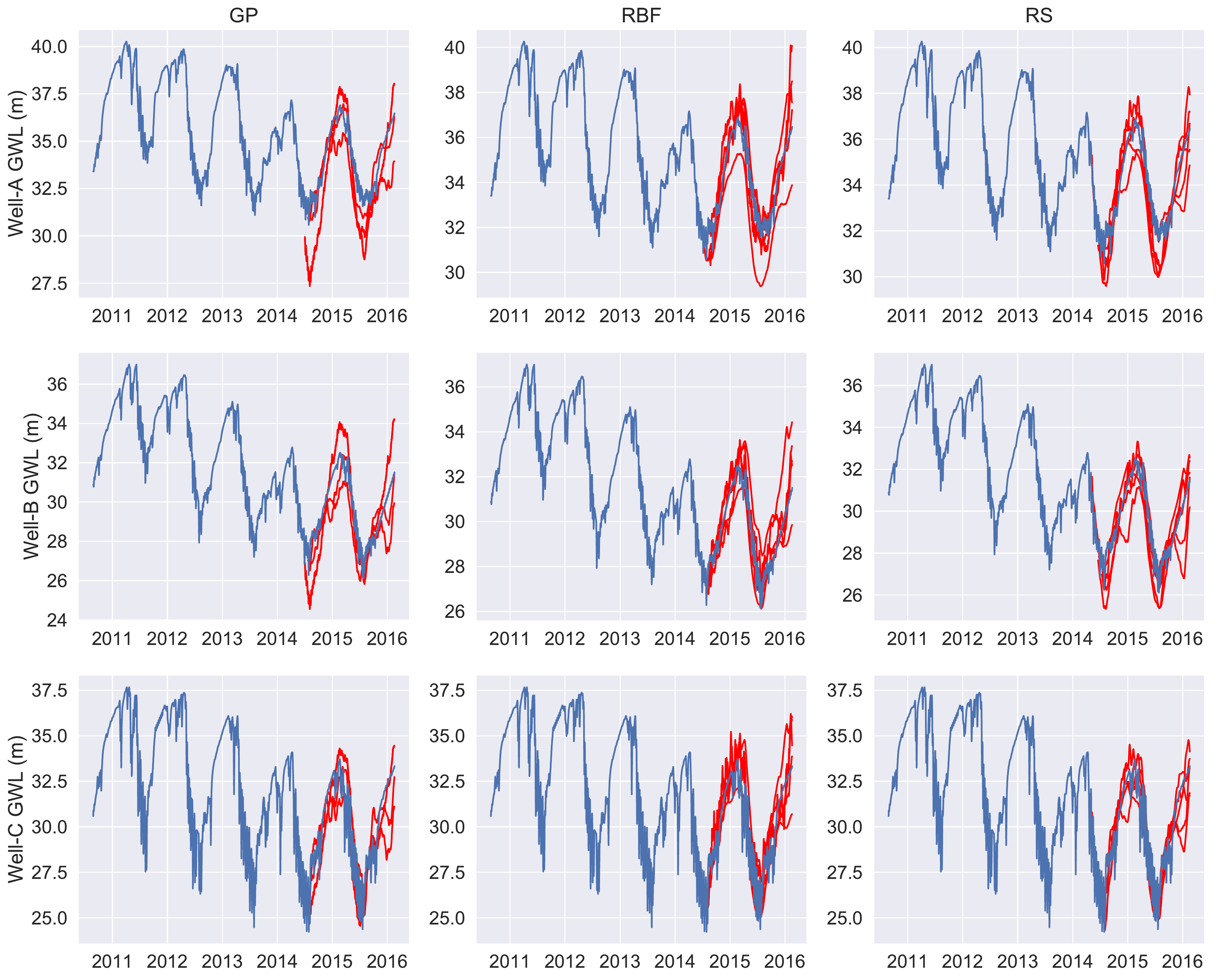}
        \caption{CNN}
        \label{fig:Multi_well_CNN_GW_pred}
    \end{subfigure}
    \caption{Groundwater level predictions obtained from five hyperparameter optimizations of LSTM (left) and CNN (right) for the multi-well case for each HPO method. }
    \label{fig:pred_multiwell}
\end{figure}

As for the single well case, we use  the optimized LSTM and CNN models to predict the groundwater levels for the testing data. The predictions for all three  wells, all HPO methods, and both models are shown in Figures~\ref{fig:Multi_well_CNN_GW_pred} (CNN) and~\ref{fig:Multi_well_LSTM_GW_pred} (LSTM). Similarly as for the MLP shown in the main document, the CNN predicts well C best as compared to the other two wells, which indicates that well C dominates the optimization (note that also for the multi-well case, three optimizations of CNN led to the same optimal solution). The LSTM predictions (Figure~\ref{fig:Multi_well_LSTM_GW_pred}) are generally further away from the true data for all three wells.  

\newpage
\subsubsection{Analysis of Computing Times}
\begin{figure}[htbp]
    \centering
    \begin{subfigure}[b]{0.48\textwidth}
        \includegraphics[width=\textwidth]{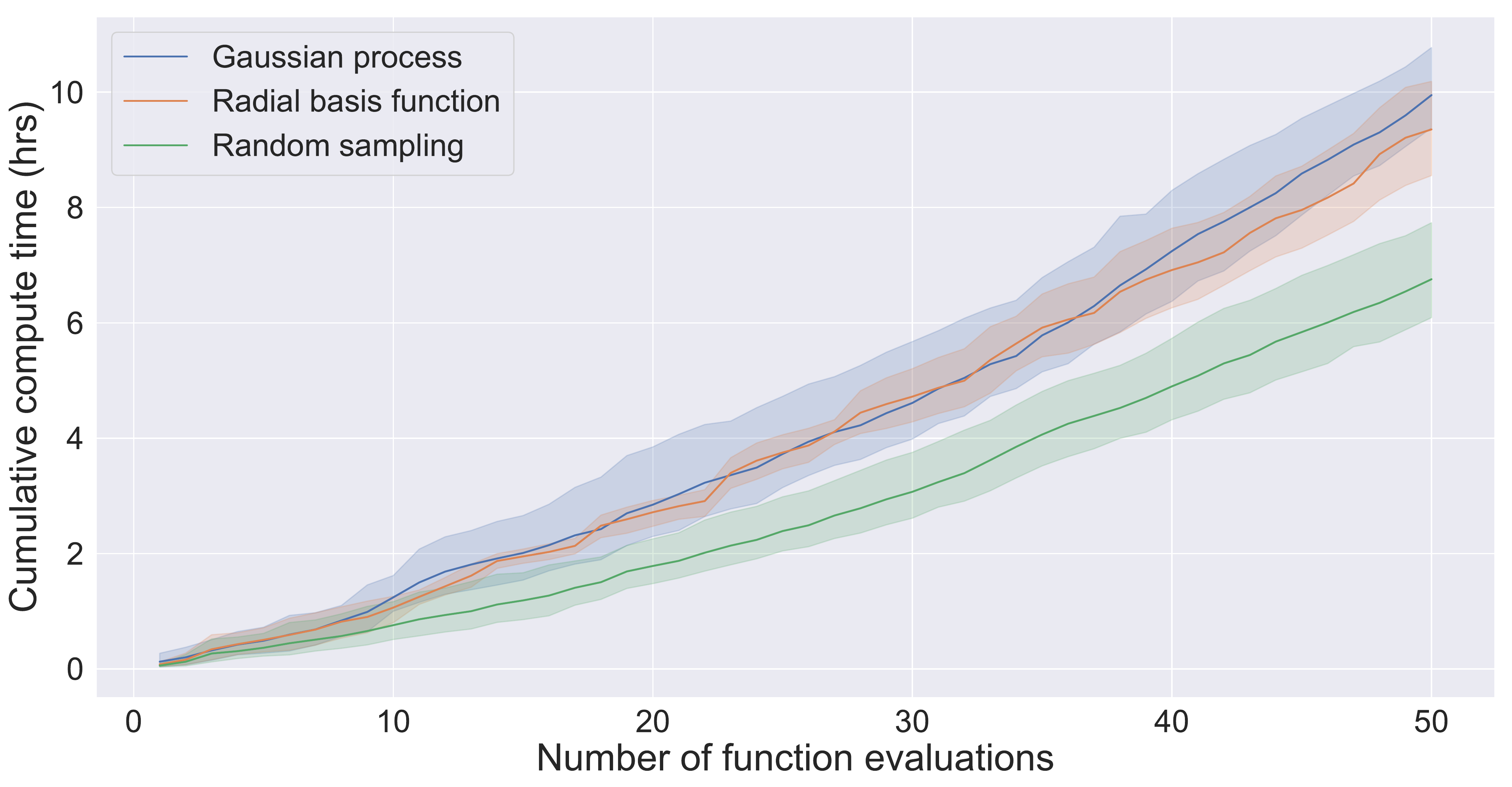}
        \caption{LSTM}
        \label{fig:Multi_well_compute_time_LSTM}
    \end{subfigure}
    ~ 
    \begin{subfigure}[b]{0.48\textwidth}
        \includegraphics[width=\textwidth]{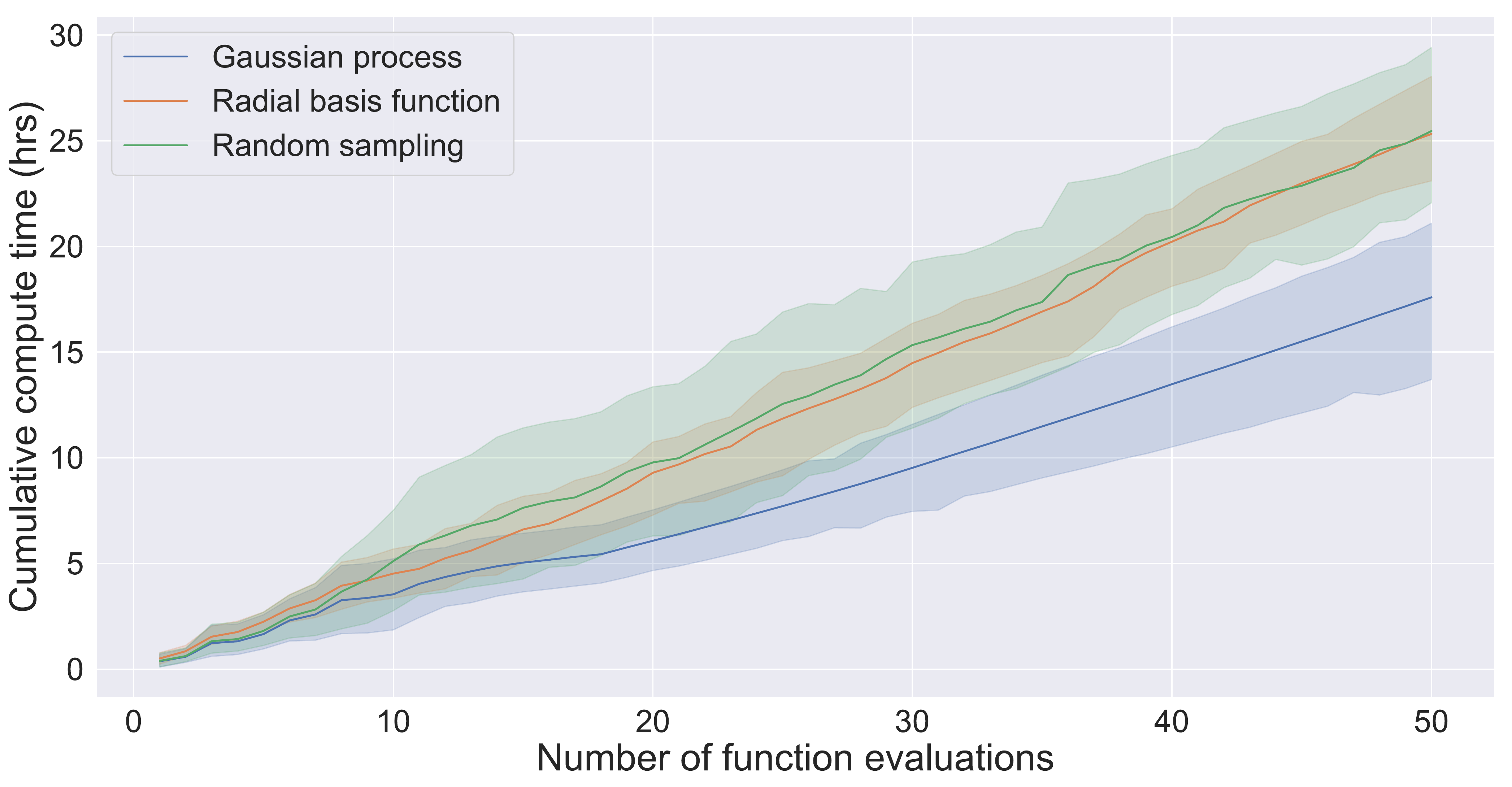}
        \caption{CNN}
        \label{fig:Multi_well_compute_time_CNN}
    \end{subfigure}
    \caption{Cumulative function evaluation time versus  number of function evaluations for  LSTM (left) and CNN (right)  for the multi-well case.}
    \label{fig:comp_multiwell}
\end{figure}

We illustrate in Figures~\ref{fig:Multi_well_compute_time_LSTM} and~\ref{fig:Multi_well_compute_time_CNN} the cumulative function evaluation times versus the number of function evaluations for LSTM and CNN, respectively. Note that these timings do not include the HPO methods' own computational overheads.  The goal of our study is to find a learning model and an HPO method that lead to the best performance on our groundwater data \textit{and} that computed within reasonable time on a laptop or simple desktop computer. Thus, in order to select the best method, we should take both metrics (accuracy and time) into account. Similarly as for the single well results, we find for the multi-well case that optimizing the hyperparameters of the CNN is considerably more time consuming than for MLP and LSTM. Optimizing the MLP for multiple wells takes less than 4 hours with all HPO methods, whereas optimizing the hyperparameters of the LSTM takes between 6 and 10 hours, depending on the HPO method. The HPO for the  CNN is computationally very expensive for all methods and may take up to 30 hours. Since MLP (optimized with the RBF) gives the best results for the multi-well case, and since it computes in less than 4 hours, we recommend this learning model and the RBF HPO. 
